%% file: arxiv_v1_main.tex
\pdfoutput=1
\RequirePackage{thmtools}
\documentclass[11pt, a4paper, copyright, gdm]{google}
\usepackage[authoryear, sort&compress, round]{natbib}

\usepackage{graphicx} %
\usepackage{amsfonts} 
\usepackage{amsmath}
\usepackage[dvipsnames]{xcolor}
\usepackage{multirow}
\usepackage{booktabs}
\usepackage{colortbl}
\usepackage{mathtools}
\usepackage{caption}
\usepackage{makecell}

\usepackage[utf8]{inputenc}
\usepackage[most]{tcolorbox}
\usepackage{listings}

\newtcolorbox{promptbox}[1][]{
  enhanced,
  breakable,
  colback=gray!4!white,          
  colframe=gray!70!black,        
  colbacktitle=gray!15!white,    
  coltitle=black,                
  fonttitle=\bfseries\small,
  arc=1.5mm,
  boxrule=0.6pt,
  left=6pt, right=6pt, top=6pt, bottom=6pt,
  toptitle=3pt, bottomtitle=3pt,
  #1
}

\newtcolorbox{jsonbox}[1][]{
  enhanced,
  colback=black!3!white,
  colframe=black!15!white,
  arc=1mm,
  boxrule=0.4pt,
  left=4pt, right=4pt, top=4pt, bottom=4pt,
  #1
}

\input{math_commands.tex}

\usepackage{subcaption}
\usepackage{wrapfig}
\usepackage[table]{xcolor} 

\usepackage{amsthm}
\usepackage{thmtools}

\usepackage{thm-restate}
\usepackage[colorlinks=true, allcolors=blue]{hyperref}
\usepackage[capitalize]{cleveref}

\usepackage{xcolor} 
\usepackage{pifont} 

\newtheorem{theorem}{Theorem}

\newtheorem{proposition}[theorem]{Proposition}

\newtheorem*{assumption*}{Assumption}

\usepackage{algorithm}
\usepackage{algpseudocode}
\usepackage{float}

\floatstyle{ruled}
\restylefloat{algorithm}
\algrenewcommand\algorithmicreturn{\textbf{Return}}

\floatname{algorithm}{Algorithm}
\algrenewcommand\algorithmicrequire{\textbf{Require:}}
\algrenewcommand\algorithmicensure{\textbf{Return:}}
\algnewcommand\algorithmicinitialize{\textbf{Initialize:}}
\algnewcommand\Initialize{\item[\algorithmicinitialize]}

\newcommand{\cmark}{\textcolor{black}{\ding{51}}}%
\newcommand{\xmark}{\textcolor{black}{\ding{55}}}%

\newcommand{\dd}{\mathop{}\!\mathrm{d}}

\definecolor{colorAR}{HTML}{555555}    %
\definecolor{colorFMLM}{HTML}{D62728}  %
\definecolor{colorGD}{HTML}{009E73}    %
\definecolor{colorGSF}{HTML}{1F77B4}   %

\colorlet{bgAR}{colorAR!12}    
\colorlet{bgFMLM}{colorFMLM!12}
\colorlet{bgGD}{colorGD!14}  
\colorlet{bgGSF}{colorGSF!20} 

\def\rvx{{\mathbf{x}}}
\def\rvy{{\mathbf{y}}}

\def\rvv{{\mathbf{v}}}
\def\rvf{{\mathbf{f}}}
\def\rvu{{\mathbf{u}}}
\def\rvl{{\boldsymbol{\ell}}}
\def\rvxi{{\boldsymbol{\xi}}}
\def\vocab{{\mathcal{V}}}

\newcommand{\scite}[1]{{\scriptsize\citep{#1}}}
\newcommand{\tcite}[1]{{\tiny\citep{#1}}}

\title{Gumbel Straight Flow: Distilling Autoregressive Models into One-step Flow Maps}
\renewcommand{\today}{September 2026}

\author[1,2]{Yeongmin~Kim}
\author[1]{Arnaud~Doucet}
\author[1]{Andrew~Campbell}
\author[1]{Valentin~De~Bortoli}
\author[1]{\protect\\Thomas~Mensink}
\author[1]{David~Ruhe}

\affil[1]{\thepa{}{}}
\affil[2]{Work done as a Student Researcher at Google DeepMind Amsterdam}

\begin{abstract}
We present Gumbel Straight Flow (GSF), a continuous flow map language model that leverages the noise-data coupling of a pretrained autoregressive language (AR) model.
We theoretically demonstrate that the coupling between Gumbel noise and one-hot token sequences induced by an autoregressive model yields non-intersecting linear paths connecting the noise to the sequence representations.
To further enhance high-quality few-step path sampling, we use a flow map semigroup objective where the tangent (velocity) condition is guided directly by the AR teacher.
Across various benchmarks, including pretraining and downstream tasks, GSF can outperform current few-step language generation baselines. 
\end{abstract}

\begin{document}

\maketitle

\section{Introduction}
\begin{wrapfigure}{r}{0.48\textwidth}
    \centering
    \vspace{-3mm}
    \includegraphics[width=\linewidth]{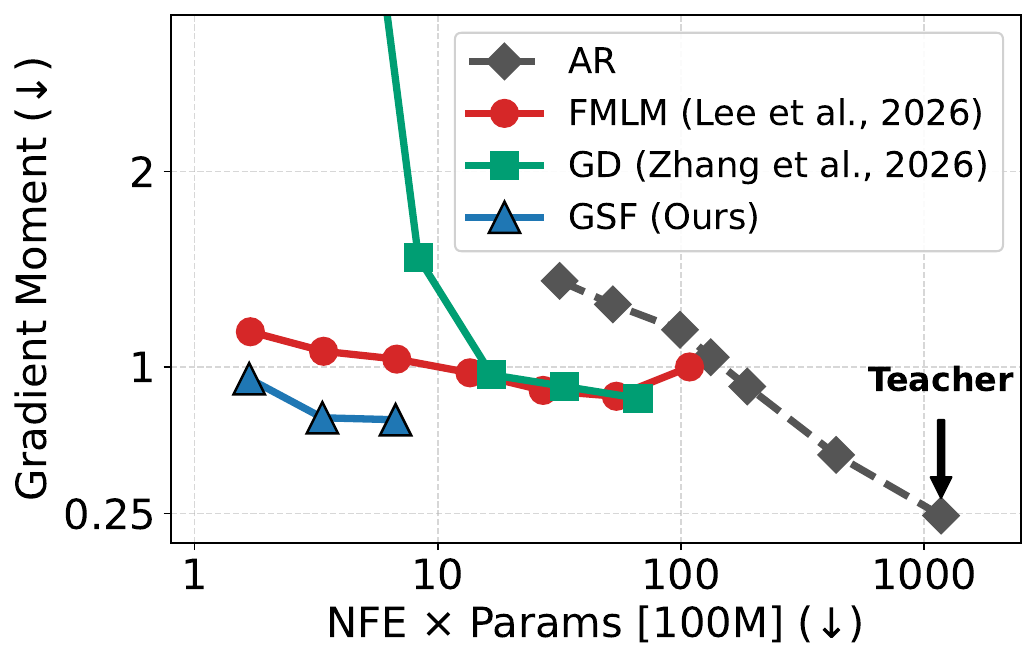}
    \vspace{-6mm}
    \caption{Efficiency-performance trade-offs for GSF, primary distillation baselines (varying step
    sizes), and AR models (varying model sizes).}
    \label{fig:model_perform}
    \vspace{-3mm}
\end{wrapfigure}

\begin{table}[b]
    \centering
    \vspace{-2mm}
    \caption{Comparison of GSF against existing language generation paradigms. This paper shows that
    the AR model's coupling forms a non-intersecting straight path and distills the path into GSF.}
    \label{tab:model_comparison}
    \vspace{-2mm}
    \small
    \begin{tabular}{lccc}
    \toprule
    \textbf{Model} & \textbf{Parallel Generation} & \textbf{Deterministic Pairing} & \textbf{Straight Path} \\
    \midrule
    Autoregressive Model              & \xmark  & \cmark &  \cmark \\ 
    Discrete Diffusion Model          & \cmark  & \xmark & \xmark  \\  
    Continuous Flow (Diffusion)       & \cmark  & \cmark & \xmark  \\
    \rowcolor{bgGSF}
    \textbf{Gumbel Straight Flow (GSF)} & \cmark  & \cmark & \cmark \\
    \bottomrule
    \end{tabular}
\end{table}

Autoregressive models (AR) are widely used for language generation due to their strong generative
performance.
However, their sequential generation process is computationally expensive.
To address this limitation, parallel-generation language models based on discrete diffusion or
continuous flow models have been
introduced~\citep{austin2021structured,dieleman2022continuous,li2022diffusion,sahoo2024simple,shi2024simplified,hu2026elf,lee2026flow}.
Nevertheless, discrete diffusion and continuous-flow models often require many sampling steps to
control factorization or discretization errors.

To accelerate inference in parallel generative models, a natural direction is distillation, i.e.,
compressing multi-step iterative generation into a few sampling steps or a single step
~\citep{pmlr-v202-song23a,deschenaux2025beyond,pmlr-v267-hayakawa25a,hoogeboom2026beyond}, often
using a discrete diffusion model as a teacher.
Given their better empirical performance on text modeling, we explore whether autoregressive (AR)
models can be used as more capable teachers.
Recent work harnesses AR teachers to supervise parallel student models by constructing deterministic
sequence-level noise–data pairs~\citep{draxler2026parallel,zhang2026gumbel}.
However, learning sequence-level noise-to-data mappings is an exceptionally hard supervised problem;
for instance, Gumbel-Distill (GD)~\citep{zhang2026gumbel} must still fall back to multi-step
refinement over a discrete diffusion backbone.

\begin{figure*}[t]
  \centering

  \begin{minipage}[c]{0.58\textwidth}
    \centering

    \hspace{0.065\linewidth} \makebox[0.275\linewidth][c]{\small\textbf{Independent}}
    \hspace{0.012\linewidth} \makebox[0.275\linewidth][c]{\small\textbf{Deterministic}}
    \hspace{0.012\linewidth}
    \makebox[0.275\linewidth][c]{%
  \raisebox{-1.5mm}{%
    \shortstack{%
      \footnotesize\textbf{Deterministic \&}\\
      \footnotesize\textbf{Non-intersecting}%
    }%
  }%
}

    \makebox[0.075\linewidth][c]{%
      \raisebox{0.07\height}{%
        \rotatebox{90}{\small\bfseries Data coupling}%
      }%
    }%
    \includegraphics[width=0.275\linewidth]{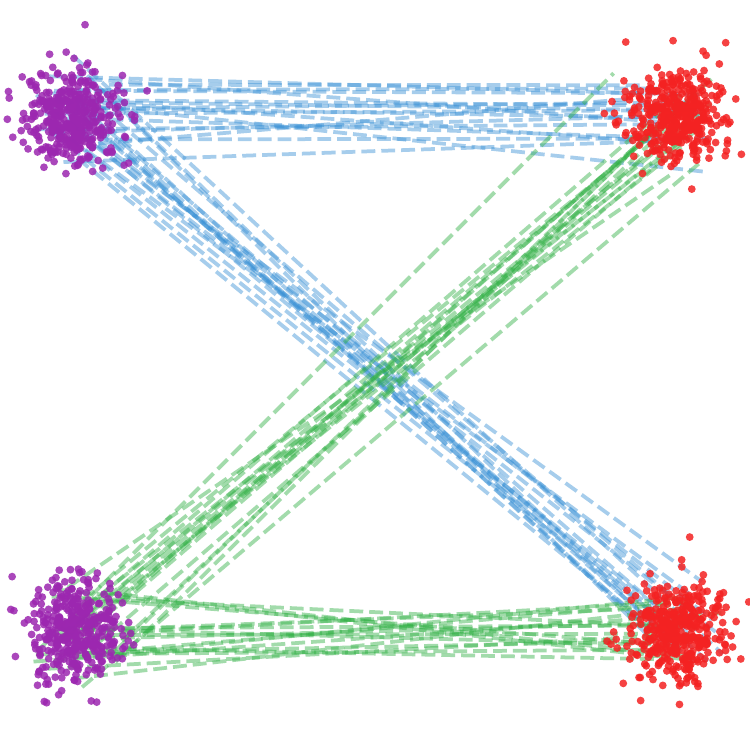}
    \hspace{0.012\linewidth}
    \includegraphics[width=0.275\linewidth]{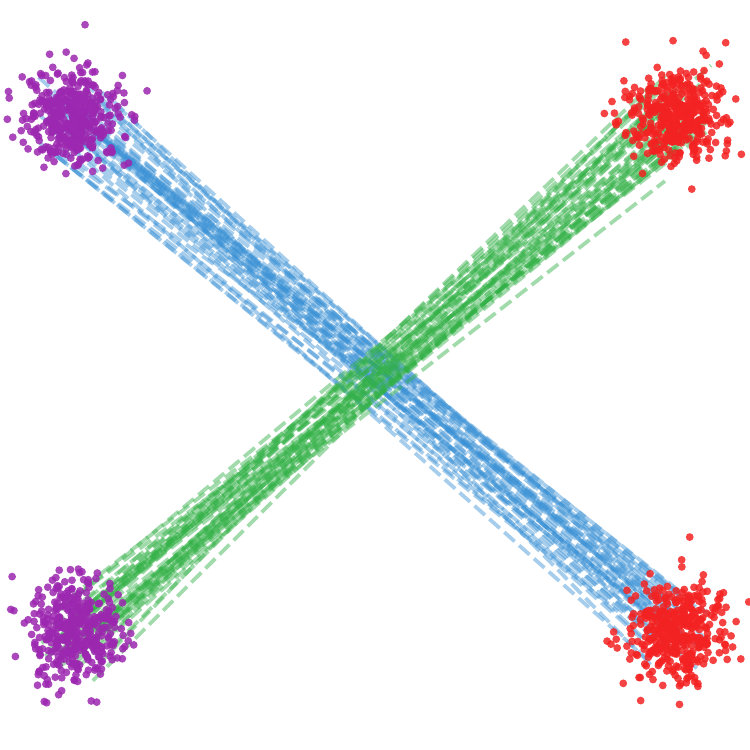}
    \hspace{0.012\linewidth}
    \includegraphics[width=0.275\linewidth]{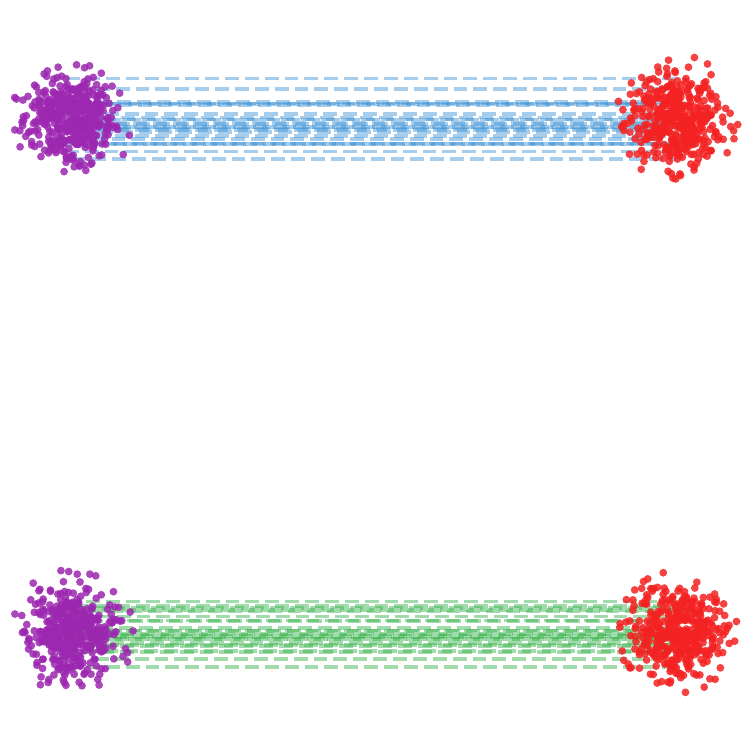}

    \vspace{1.8mm}

    \makebox[0.075\linewidth][c]{%
      \raisebox{0.0\height}{%
        \rotatebox{90}{\small\bfseries Underlying path}%
      }%
    }%
    \includegraphics[width=0.275\linewidth]{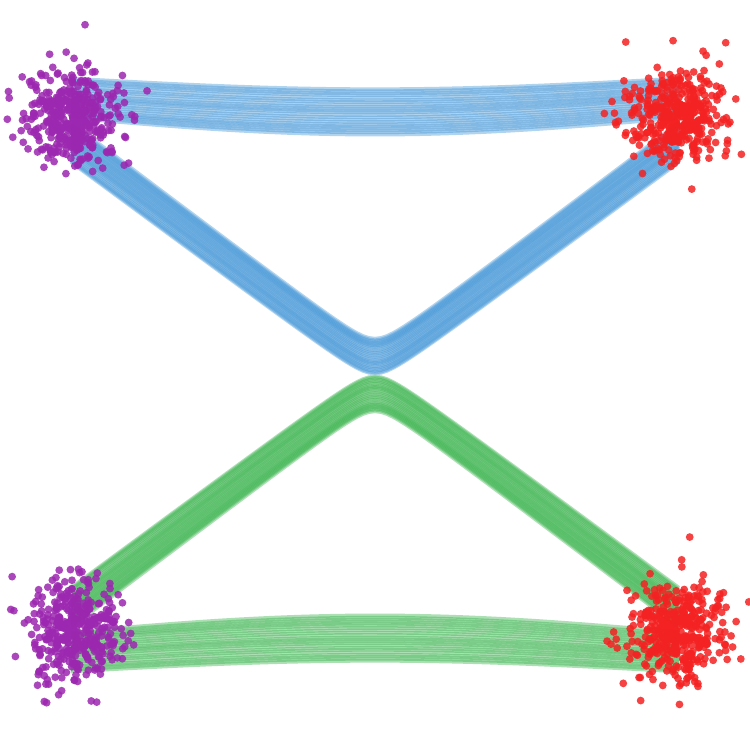}
    \hspace{0.012\linewidth}
    \includegraphics[width=0.275\linewidth]{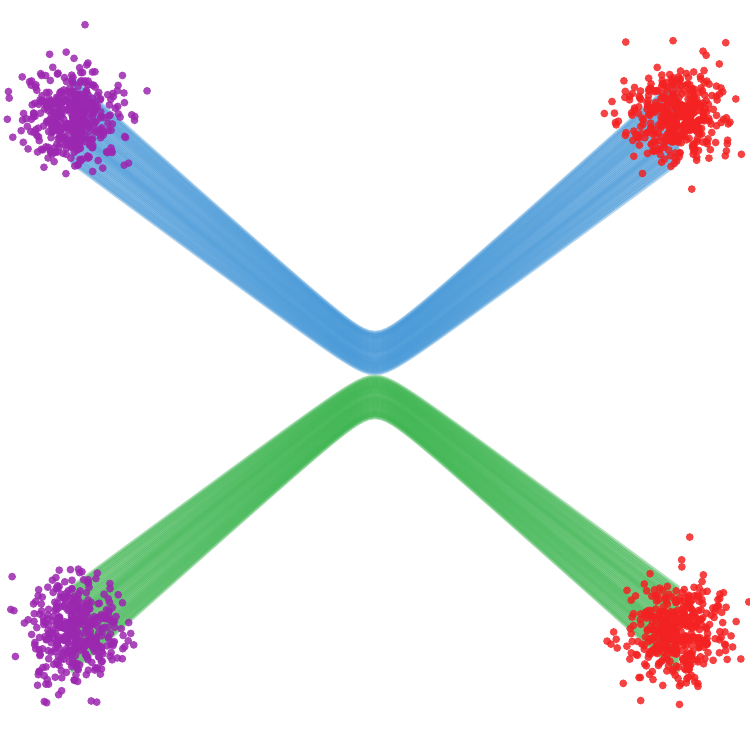}
    \hspace{0.012\linewidth}
    \includegraphics[width=0.275\linewidth]{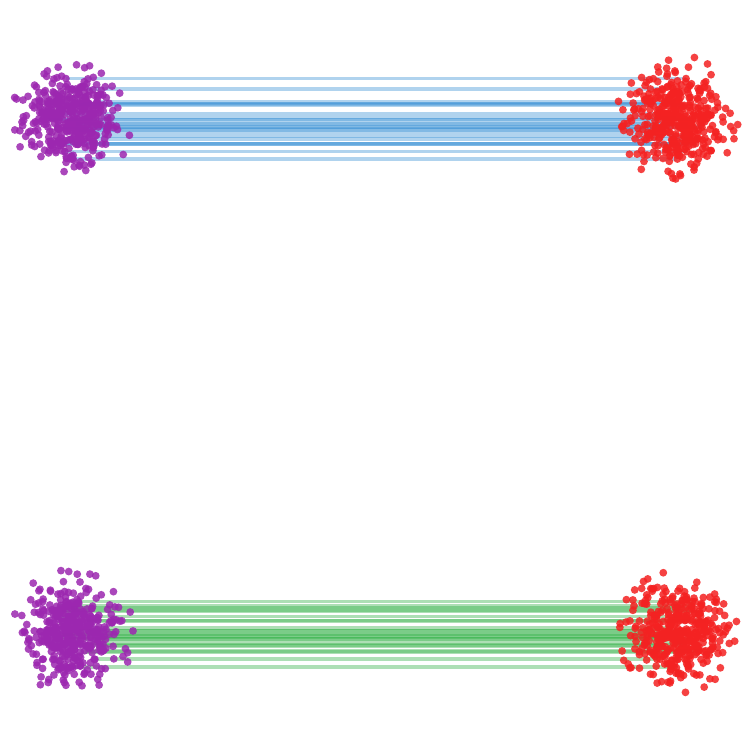}

  \end{minipage}
  \hfill
  \begin{minipage}[c]{0.38\textwidth}
    \centering
    \includegraphics[width=\linewidth]{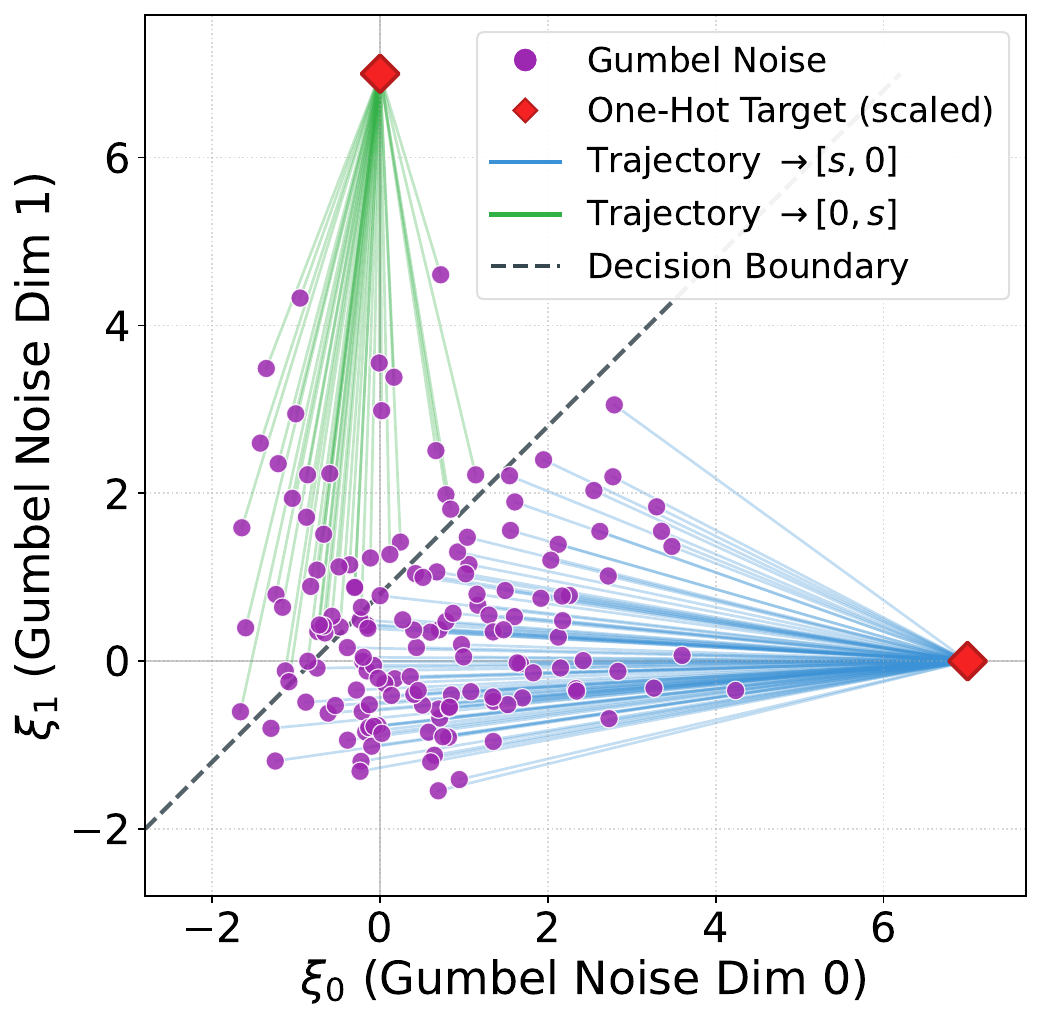}
  \end{minipage}

  \caption{ (Left) Three types of noise--data coupling and their corresponding optimally learned
  flow paths.
  (Right) The Gumbel coupling induced by an AR model belongs to the non-intersecting case and yields
  straight underlying paths.
  }
  \label{fig:seven_panels}
  \par\vspace{-3mm}
\end{figure*}

In this paper, we instead define a flow path in one-hot continuous space directly over the AR
model's Gumbel-max sampling.
This defines a coupling between the distribution of Gumbel sequences $\rvx_0$ and corresponding
one-hot token vectors $\rvx_1$ such that, for endpoints distributed according to this coupling, the
linear interpolants $\rvx_t = (1-t)\rvx_0 + t\rvx_1$ do not intersect (\Cref{fig:seven_panels}).
Gumbel Straight Flow (GSF) then learns these straight paths provided by an AR teacher while enabling
parallel generation.
Straight paths are desirable: minimizing trajectory curvature allows numerical ODE solvers to
generate samples in a few steps without discretization issues~\citep{liu2023flow}.
We further enhance few-step performance using a flow map parameterization (FMLM)~\citep{lee2026flow}
where the straight tangent (velocity) supervision is given by the AR teacher.
As summarized in \Cref{tab:model_comparison}, GSF is (to our knowledge) the first
parallel-generation model to combine deterministic pairing with the straight, non-intersecting
trajectory structure of autoregressive sampling.

Empirical results show that one-step GSF achieves a Gradient Moment (GM)\footnote{GM tends to be
more robust against mode collapse than Gen.
PPL~\citep{franca2026hacking}.}~\citep{hoogeboom2026beyond} score of 0.94, establishing a new
efficiency-performance frontier compared to GD and FMLM as shown in \Cref{fig:model_perform}.
Furthermore, GSF sits on a new, better Pareto frontier than that of even AR models in terms of the
efficiency-compute trade-off.
The four-step GSF achieves Gen.
PPL = 99.7 and Entropy = 5.32, placing it in the frontier region (\Cref{fig:ppl_entropy_tradeoff})
for fidelity-diversity trade-offs under the identical sampling budget.

\section{Preliminaries}

\subsection{Autoregressive Models}
\label{sec:2.1}
Let $\mathcal V$ denote a finite vocabulary and $L$ the sequence length.
At each position $i \in \{1, \dots, L\}$, conditioned on the preceding tokens
$\rvy_{<i} = (y_1, \dots, y_{i-1})$, an autoregressive (AR) language model produces a logit vector
$\rvl_i(\rvy_{<i}) = (\ell_{i, 1}(\rvy_{<i}), \dots, \ell_{i, |\mathcal V|}(\rvy_{<i})) \in \mathbb
R^{|\mathcal V|}$ and computes the categorical probability of token $j \in \mathcal V$ as:
\begin{equation}
p_{\text{AR}}(y_i = j \mid \rvy_{<i}) = \frac{\exp(\ell_{i, j}(\rvy_{<i}))}{\sum_{k \in \mathcal V} \exp(\ell_{i, k}(\rvy_{<i}))}. \label{eq:categorical}
\end{equation}
The joint probability of generating a sequence of $L$ tokens $\rvy_{1:L} = (y_1, \dots, y_L)$ is
given by the chain rule: $p_{\text{AR}}(\rvy_{1:L}) = \prod_{i=1}^{L} p_{\text{AR}}(y_i \mid
\rvy_{<i})$.

\subsection{Gumbel-Max Categorical Sampling}
\label{sec:2.2}
Sampling a token sequence from the autoregressive distribution in \cref{eq:categorical} can be
performed ancestrally via the Gumbel-max trick~\citep{gumbel1954statistical}.
Let $\rvxi_i = (\xi_{i, 1}, \dots, \xi_{i, |\mathcal V|})\in \mathbb R^{|\mathcal V|}$ denote
independent standard Gumbel noise vectors with $\xi_{i, j} \overset{\text{i.i.d.}}{\sim} \mathcal
G(0, 1)$ across all positions $i \in \{1, \dots, L\}$ and vocabulary coordinates $j \in \mathcal V$.
At each position $i$, the next token is generated deterministically given the noise and the previous
history:
\begin{equation}
\label{eq:gumbel-max}
y_i = \arg\max_{j \in \{1, \dots, |\mathcal V|\}} \big(\ell_{i, j}(\rvy_{<i}) + \xi_{i, j}\big).
\end{equation}
By the Gumbel-max trick, the resulting token $y_i$ is distributed according to
$p_{\text{AR}}(\cdot \mid \rvy_{<i})$ from \cref{eq:categorical}, ensuring that the full sampled
sequence satisfies $\rvy_{1:L} \sim p_{\text{AR}}(\rvy_{1:L})$.
Writing $\rvxi_{1:L} \in \mathbb R^{L \times |\mathcal V|}$ and $\rvy_{1:L} \in \mathcal V^L$,
ancestral Gumbel sampling defines a deterministic mapping $T_{\text{AR}}: \rvxi_{1:L}
\xrightarrow{\mathrm{det.}} \rvy_{1:L}$ that couples the continuous Gumbel noise sequence with the
discrete (teacher) generated sequence.

\subsection{Continuous-state Language Flow}
\label{sec:2.3}
Following \citet{lee2026flow}, we embed text sequences into continuous space via one-hot encoding:
\begin{align}
& \text{Encoding: } \rvx=\operatorname{onehot}(\rvy_{1:L})
\in\mathbb R^{L\times|\mathcal V|}, \quad\quad \text{Decoding: } \rvy_{1:L}=\arg\max(\rvx). \nonumber
\end{align}
\noindent\textbf{Flow Language Model (FLM).}
FLM~\citep{lee2026flow} is based on flow matching (FM)~\citep{lipman2023flow,liu2023flow}.
It defines two independent marginal distributions: a Gaussian source distribution $\rvx_0 \sim p_0
\coloneqq \mathcal{N}(\mathbf{0},\mathbf{I})$ and a one-hot-encoded data distribution
$\rvx_1 \sim p_1 \coloneqq p_{\text{data}}$.
FLM defines the generative process as $\dd \rvx^{{\theta}}_t=\rvv_{\theta}(\rvx^{\theta}_t,t) \dd t$
and can train the velocity model $\rvv_{\theta}$ using the mean squared error (MSE) loss:
\[
\min_{\theta}
\int_0^1
\mathbb E_{\rvx_0 \sim p_0, \rvx_1 \sim p_1}
\Big[
\|
(\rvx_1-\rvx_0)-\rvv_{\theta}(\rvx_t,t)
\|_2^2
\Big]
\dd t, \qquad \rvx_t=(1-t)\rvx_0 + t \rvx_1,
\]
where the optimal velocity is $\rvv(\rvx_t,t) \coloneqq \mathbb{E}_{p(\rvx_0,\rvx_1 \mid
\rvx_t)}[\rvx_1-\rvx_0 \mid \rvx_t]$.
However, this velocity parameterization does not exploit the fact that the data lie on the one-hot
simplex.
Recent studies~\citep{lee2026flow,potaptchik2026discrete} instead adopt a denoising formulation, in
which the denoising network uses a softmax output layer and is trained with the cross-entropy loss:
\begin{equation}
\min_{\theta}
\int_0^1 \mathbb E_{\rvx_0 \sim p_0, \rvx_1 \sim p_1} \Big[\text{CE}(\rvx_1||D_{\theta}(\rvx_t,t))\Big]\dd t, \quad \quad D_{\theta}(\rvx_t,t):= \rvx_t + (1-t) \rvv_{\theta}(\rvx_t,t). \label{eq:FLM_train}
\end{equation}
This formulation has the same optimum as the MSE objective.
Specifically, the optimal denoiser is given by $D(\rvx_t,t) :=\mathbb{E}[\rvx_1 \mid \rvx_t] =
\rvx_t+(1-t)\rvv(\rvx_t,t)$.

\noindent\textbf{Flow Map Language Model (FMLM).}
FMLM~\citep{lee2026flow,roos2026categorical} is a flow map variant of FLM that defines $X_{s,t}:
\mathbb{R}^{L\times|\vocab|}\rightarrow\mathbb{R}^{L\times|\vocab|}$ by integrating the velocity
field:
\[
X_{s,t}(\rvx_s) := \rvx_s + \int_{s}^{t}\rvv(\rvx_\tau,\tau) \dd \tau = \rvx_s + (t-s) \mathbf{u}_{s,t}(\rvx_s),
\]
where $\mathbf{u}_{s,t}$ denotes the average velocity from time $s$ to $t$.
A flow map satisfying this property characterizes the underlying flow path:
\begin{equation}
\rvu_{t,t}(\rvx_t) = \lim_{s\rightarrow t} \rvu_{s,t}(\rvx_t) = \rvv(\rvx_t,t), \quad \quad X_{u,t}(X_{s,u}(\rvx_s)) = X_{s,t}(\rvx_s). \label{eq:semigroup_cond}
\end{equation}
Define the corresponding two-time denoiser formulation as:
\[
\delta_{s,t}(\rvx_s):= \rvx_s + (1-s) \rvu_{s,t}(\rvx_s), 
\]
and the objective for training the parameterized two-time denoiser $\delta^{\theta}$ to satisfy the
flow map condition in \cref{eq:semigroup_cond} is given by:
\begin{align}
\mathcal{L}_{\text{FMLM}}(\theta)
&= \underbrace{\mathbb{E}_{t, \rvx_0, \rvx_1}\Big[ \mathrm{CE}\!\left( D(\rvx_t,t) \,\middle\|\, \delta^{\theta}_{t,t}(\rvx_t) \right) \Big]}_{\textstyle :=\mathcal{L}_{\text{diag.}}(\theta; D)} \nonumber \\
&\quad + \mathbb{E}_{s,u,t, \rvx_0, \rvx_1}\Big[\underbrace{ \mathrm{CE}\!\left( \operatorname{sg}\!\left( \gamma \delta^{\theta}_{s,u}(\rvx_s) + (1-\gamma)\delta^{\theta}_{u,t}\!\left(X_{s,u}(\rvx_s)\right) \right) \,\middle\|\, \delta_{s,t}^{\theta}(\rvx_s) \right)}_{\textstyle :=\ell_{\text{off-diag.}}(\theta)} \Big], \label{eq:flowmap}
\end{align}
where $\gamma \coloneqq \frac{(1-t)(u-s)}{(1-u)(t-s)}$, and $\text{sg}(\cdot)$ is a stop gradient
operation.
Here, $\rvx_0 \sim p_0$ and $\rvx_1 \sim p_1$ are sampled independently.
In practice, $D$ is replaced with a pretrained FLM model from \cref{eq:FLM_train}.

\begin{figure*}[t]
    \centering
        \includegraphics[width=\linewidth]{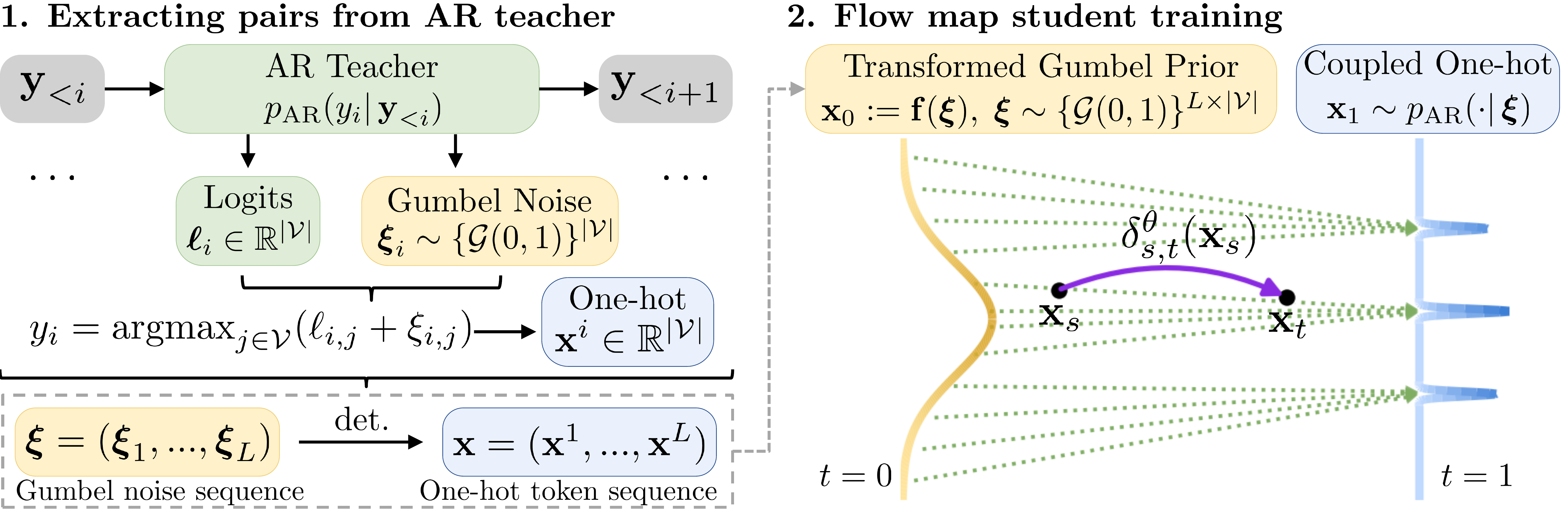}
    \caption{Overview of GSF: distilling AR-induced straight paths into a flow map student.}
    \label{fig:overview}
    \par\vspace{-2mm}
\end{figure*}

\section{Method}
\label{sec:3}
This section introduces Gumbel Straight Flow (GSF), a method for distilling an autoregressive (AR)
teacher into a flow map.
\Cref{sec:3.1} analyzes the properties of AR-induced paths (green dotted lines in
\Cref{fig:overview}), and \Cref{sec:3.5} explains how to distill these paths into GSF (purple arrow
in \Cref{fig:overview}).

\subsection{Analysis of AR-induced Gumbel Coupling Paths}
\label{sec:3.1}
\Cref{sec:2.2} shows that AR models provide deterministic noise-to-data supervision. However, this deterministic relationship alone does not guarantee underlying straight paths (second column of \Cref{fig:seven_panels}). This section uses a flow matching parametrization to show that the %
Gumbel coupling induced by AR models yields straight paths, corresponding to the third column of
\Cref{fig:seven_panels}.

\textbf{Gumbel Straight Flow with Flow Matching (GSF-FM)}.
We define the generative process as $\dd \rvx^{\theta}_t=\rvv_{\theta}(\rvx^{\theta}_t,t) \dd t$ and
train it using the AR-induced coupling, where $\rvx_0:=\rvxi \sim {\mathcal{G}(0,1)}^{L\times
|\vocab|}$ and $\rvx_1 \sim p_{\text{AR}}(\rvx_1|\rvxi)$.
The training objective is:
\begin{align}
\mathcal{L}_{\text{GSF-FM}}(\theta,p_{\text{AR}}):=
\int_0^1
\mathbb E_{\rvx_0 \sim \mathcal{G},\; \rvx_1 \sim p_{\text{AR}}(\cdot\mid\rvx_0)}
\Big[
\|
(\rvx_1-\rvx_0)-\rvv_{\theta}(\rvx_t,t)
\|_2^2
\Big]
\dd t,\label{eq:GSF-FM}
\end{align}
with the usual interpolant $\rvx_t=(1-t)\rvx_0 + t \rvx_1$.

\subsubsection{Straightness}
\label{sec:3.2}
The straightness measure~\citep{liu2023flow} for a path
$\{\rvx^{\theta}_t\}_{t\in[0,1]}$~\footnote{The path is a random variable, over which the
expectation is taken for straightness calculation in \cref{eq:straightness}.
We omit the time index range $t\in[0,1]$ hereafter for brevity.} with velocity $\rvv_{\theta}$ is
defined as
\begin{equation}
S(\{\rvx_t^{\theta}\}):= \int_0^1 \mathbb{E}[||(\rvx^{\theta}_1 - \rvx^{\theta}_0) - \rvv_{\theta}(\rvx^{\theta}_t,t)||_2^2] \dd t, \label{eq:straightness}
\end{equation}
and the path $\{\rvx^{\theta}_t\}$ is straight if $S(\{\rvx^{\theta}_t\})=0$.
The following theorem states that the optimal path learned using $\mathcal{L}_{\text{GSF-FM}}$ is
straight; i.e., the underlying paths provided by the AR model are straight.
\begin{restatable}{theorem}{thma} \label{thma}
\textbf{(Straightness)} Assume sufficient model capacity.
Let the optimal velocity be $\rvv_{\theta^*}$ where $\theta^*:=\arg\min_{\theta}
\mathcal{L}_{\text{GSF-FM}}(\theta,p_{\text{AR}})$, and let $\{\rvx^{\theta^*}_t\}$ denote the
corresponding path.
Then the path is straight, i.e.,
\[
S(\{\rvx^{\theta^*}_t\}) = 0.
\]
\end{restatable}
The proof (\Cref{sec:A.2}) relies on the fact that, under our Gumbel coupling, both $\rvx_0$ and
$\rvx_1$ are uniquely identifiable given $\rvx_t$, because the paired paths do not intersect.
By non-intersection, we mean that two distinct trajectories cannot be equal to the same state at the
same time $t$ for $t<1$.
Assuming the linear pairings intersect, we obtain a contradiction: the argmax ordering from
\cref{eq:gumbel-max} no longer holds.
The right panel of \Cref{fig:seven_panels} visualizes the straight paths in the single token case.
As detailed in \Cref{sec:app_ot}, this non-intersection property admits a semi-discrete optimal
transport interpretation~\citep{gabriel2019computational}: for a single token, Gumbel-max sampling
is the exact optimal transport map where teacher logits act as closed-form dual weights.
For token sequences, the transport is obtained by applying a sequence of single-token transport maps
successively; i.e., it is a causal transport map \citep{backhoff2017causal}.

\subsubsection{Straightness Holds under Prior Gaussianization}
\label{sec:3.3}
\begin{figure*}[t]
    \centering
    \begin{subfigure}[t]{0.485\textwidth}
        \centering
        \includegraphics[width=\linewidth]{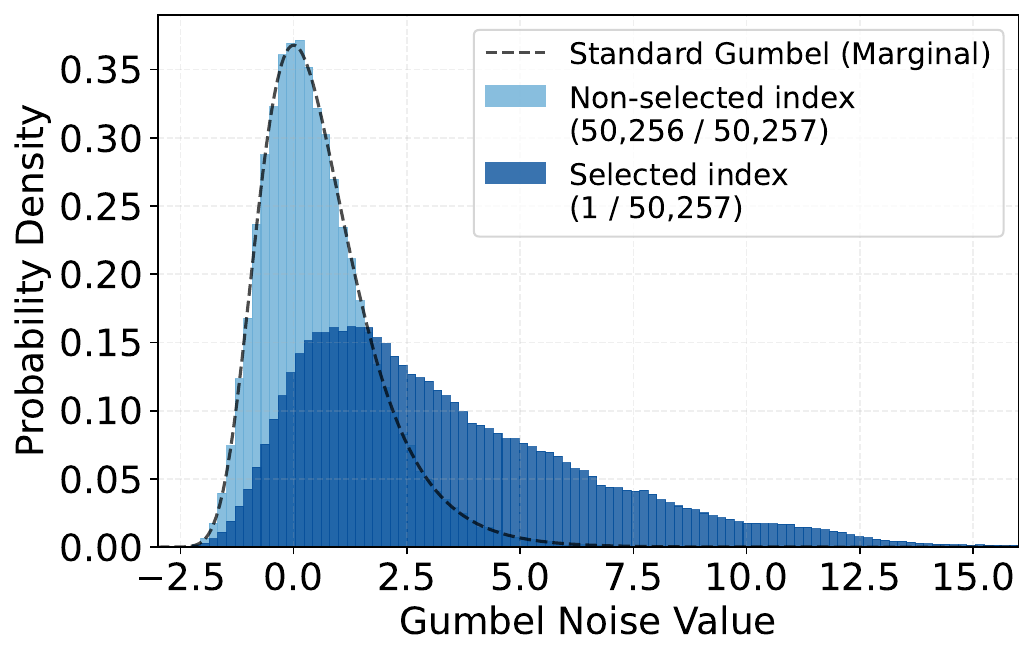}
        \caption{Gumbel noise statistics from 8 sentences extracted from an AR model.}
        \label{fig:gumbel_stat}
    \end{subfigure}
    \hfill
    \begin{subfigure}[t]{0.485\textwidth}
        \centering
    \includegraphics[width=\linewidth]{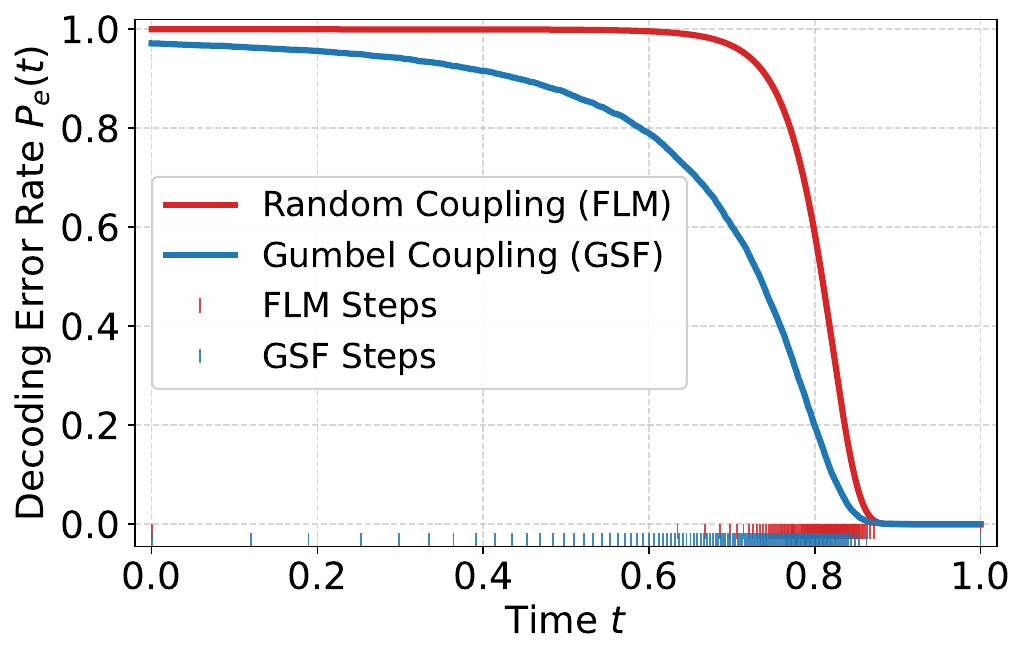}
        \caption{Decoding error rate and the reparameterized step $\tau(t)$ for uniformly
        distributing denoising information.}
        \label{fig:coupled_der}
    \end{subfigure}
    \hfill
    \caption{(a) Under the Gumbel-max trick, the selected index tends to have a larger value than
    non-selected indices.
    Since only one of the $\vocab$ indices is selected, the marginal distribution follows a standard
    Gumbel distribution.
    (b) GSF injects informative noise; consequently, decoding errors occur as the noise-injection
    signal increases and are distributed more uniformly over time $t$.}
    \label{fig:gumbel_stat_schedule}
    \par\vspace{-5mm}
\end{figure*}

Despite its theoretical straightness, \cref{eq:GSF-FM} uses a Gumbel prior distribution, which is
strongly right-skewed, as shown by the dotted line in \Cref{fig:gumbel_stat}.
Since existing flow model architectures are typically designed for Gaussian priors, they may not
effectively capture information from the Gumbel distribution.
We extend our method by applying a scalar transformation $f: \mathbb{R} \rightarrow \mathbb{R}$
element-wise to the Gumbel prior $\rvxi$.
Let $\rvf$ denote its element-wise extension, i.e., $[\mathbf{f}(\rvxi)]_{i,j}=f(\xi_{i,j})$, and
choose $f$ such that $\mathbf{f}(\rvxi)$ follows a Gaussian distribution.
The corresponding extended objective function becomes
$\mathcal{L}_{\text{GSF-FM}}(\theta,p_{\text{AR}}; \rvf):= \int_0^1 \mathbb E_{\rvxi\sim
\mathcal{G},\; \rvx_0 = \rvf(\rvxi),\; \rvx_1 \sim p_{\text{AR}}(\cdot\mid\rvxi)} \Big[ \|
(\rvx_1-\rvx_0)-\rvv_{\theta}(\rvx_t,t) \|_2^2 \Big] \dd t$.
The following proposition shows that straightness is preserved under a class of transformations $f$.
\begin{restatable}{proposition}{propositiona}
\label{prop:gaussianization}
\label{propositiona}
\textbf{(Prior transform).}
Let $f: \mathbb{R} \rightarrow \mathbb{R}$ be a strictly increasing and invertible function, and let
$\rvf$ denote its element-wise extension.
Under sufficient model capacity, let $\rvv_{\theta^*}(\rvx_t, t) := \arg\min_{\rvv_{\theta}}
\mathcal{L}_{\text{GSF-FM}}(\theta, p_{\text{AR}}; \rvf)$ denote the optimal velocity field, and let
$\{\rvx^{\theta^*}_t\}$ denote the corresponding induced trajectory.
Then, the path is still straight, i.e., $S(\{\rvx^{\theta^*}_t\}) = 0$.
\end{restatable}
Defining the transformation as the composition of a uniform transform and the inverse Gaussian CDF,
i.e., $f(\xi) := \Phi^{-1}(\exp(-e^{-\xi}))$, allows us to use a standard Gaussian prior while
maintaining optimal straightness, as $f$ is strictly increasing and invertible (proof in
\Cref{sec:A.3}).
Unless otherwise specified, we employ the Gumbel-to-Gaussian mapping for $f$, making a Gaussian
prior the default across all experiments, recovering the standard FLM / FMLM
setting~\citep{lee2026flow}.

\subsubsection{Time Reparameterization}
\label{sec:3.4}
Following \cite{sahoo2025the} and \cite{pynadath2026candi}, the decoding error rate and the
corresponding time reparameterization that uniformly distributes decoding errors are defined as:
\[
P_e(t):=\textup{Prob}(\text{argmax}(\rvx_t)\neq\text{argmax}(\rvx_1)), \quad \tau(t):=\frac{P_e(0) - P_e(t)}{P_e(0)}.
\]
When $|\vocab|$ is large, decoding errors in FLM concentrate around a specific timestep (near
$t=0.8$ in \Cref{fig:coupled_der}), which necessitates a reparameterized time schedule $\tau(t)$ to
improve performance~\citep{lee2026flow}.
In contrast, the coupled Gumbel noise injected in GSF is correlated with the selected
$\operatorname{argmax}$ index (\Cref{fig:gumbel_stat}).
As illustrated in \Cref{fig:coupled_der}, this correlation shifts most decoding errors toward higher
noise scales and distributes them more uniformly across Euclidean space, keeping $\tau(t)$ closer to
the original time $t$.
We still use the coupled version of the reparameterized schedule by default, but the required time
warping is substantially milder than in FLM.

\subsection{Flow Map with AR-induced Gumbel Coupling}
\label{sec:3.5}
\begin{figure*}[t]
    \centering
    \begin{subfigure}[t]{0.325\textwidth}
        \centering
        \includegraphics[width=\linewidth]{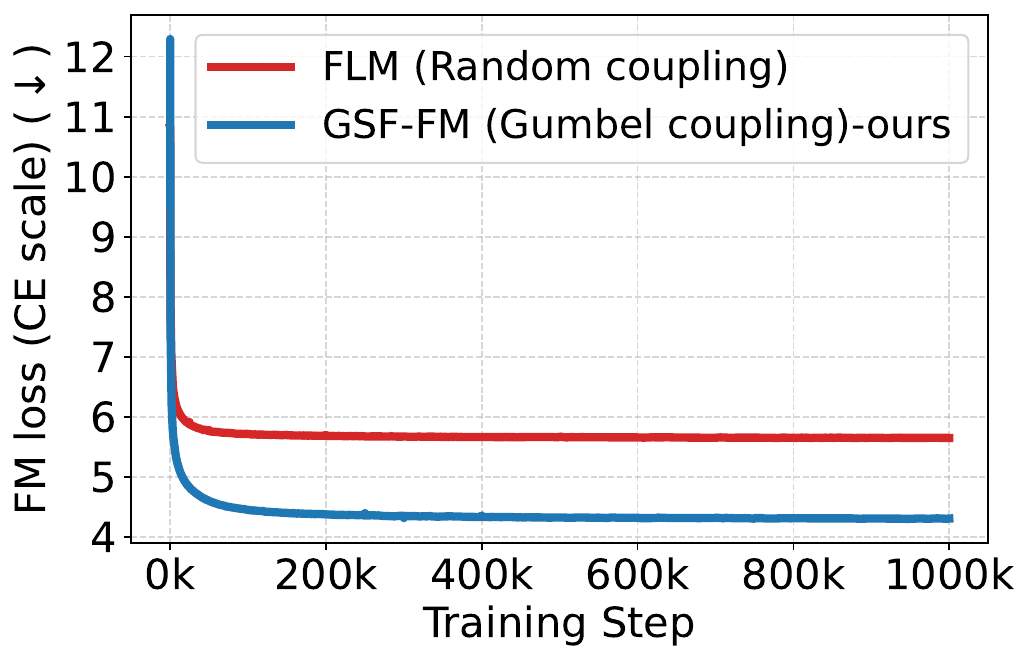}
        \caption{Training Loss}
        \label{fig:sub1}
    \end{subfigure}
    \hfill
    \begin{subfigure}[t]{0.325\textwidth}
        \centering
    \includegraphics[width=\linewidth]{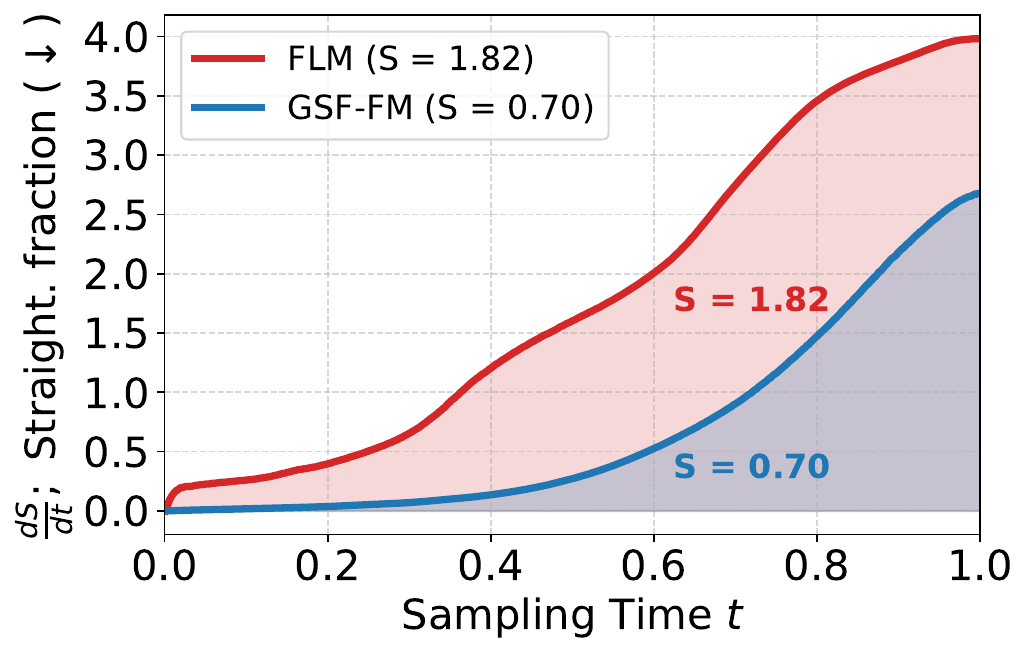}
        \caption{Straightness (CE scale)}
        \label{fig:sub2}
    \end{subfigure}
    \hfill
    \begin{subfigure}[t]{0.325\textwidth}
        \centering
        \includegraphics[width=\linewidth]{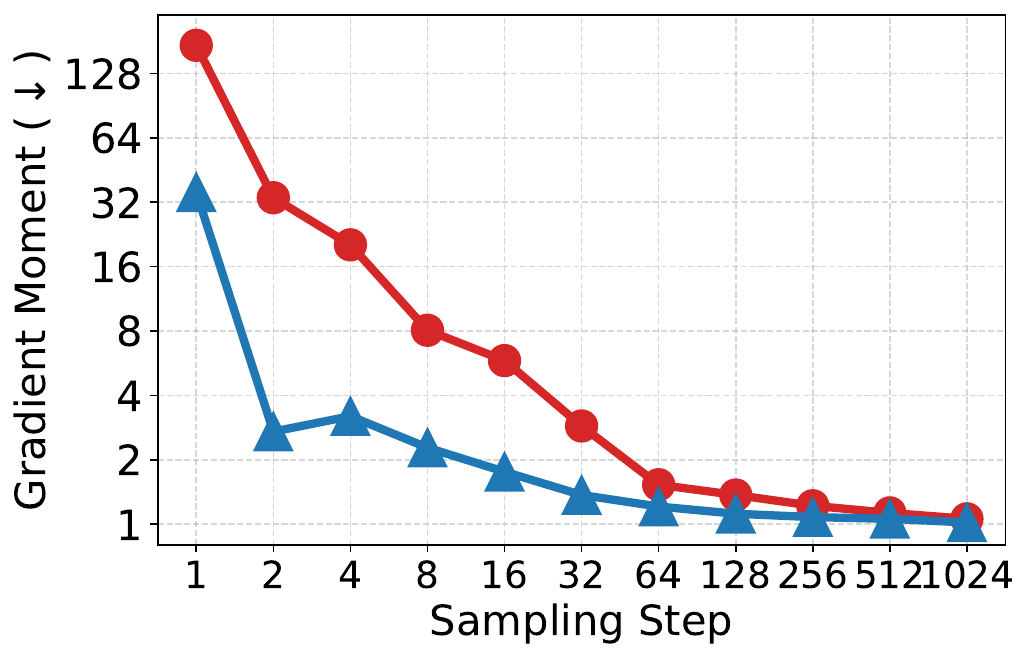}
        \caption{Generative Performance}
        \label{fig:sub3}
    \end{subfigure}
    \caption{Comparison between FLM with independent coupling and GSF-FM with AR-based coupling on
    the OpenWebText dataset under a Gaussian prior.}
    \label{fig:three_plots}
    \par\vspace{-3mm}
\end{figure*}

\Cref{fig:three_plots} demonstrates that GSF-FM achieves superior empirical straightness
(\Cref{fig:sub2}), thereby boosting generative performance (\Cref{fig:sub3}) particularly in the
low-NFE regime.
However, a substantial gap still persists between its few-step and many-step performance.
This motivates us to adopt a flow map parameterization.
Although flow maps inherently provide strong few-step generation, prior
work~\citep{liu2023flow,liu2024instaflow} has shown that an underlying straight path greatly
facilitates flow map training.
We combine the semi-group flow map training objective \cref{eq:flowmap} into:
\begin{align}
& 
\mathcal{L}_{\text{GSF}}(\theta,p_{\text{AR}}):=\underbrace{\mathbb{E}_{t}\mathbb{E}_{\rvx_0\sim \rvf\#\mathcal{G},\; \rvx_1 \sim p_{\text{AR}}}
\bigg[
\mathrm{CE}\!\left(
\rvx_1
\,\middle\|\,
\delta^{\theta}_{t,t}(\rvx_t)
\right)
\bigg]}_{\textstyle:=\mathcal{L}_{\text{diag.}}(\theta; p_{\text{AR}})} + \mathbb{E}_{s,u,t}\mathbb{E}_{\mathbf{x}_0,\mathbf{x}_1}
\big[\ell_{\text{off-diag.}}(\theta)\big].\label{eq:flowmap_gsf}
\end{align}

Regarding the off-diagonal loss, because it enforces self-consistency rather than learning the
underlying path geometry, the sample pair $(\rvx_0, \rvx_1)$ does not necessarily need to be drawn
from the coupled setting.
Empirically, we find no significant difference based on how $(\rvx_0, \rvx_1)$ is sampled;
independent sampling yields better diversity at the cost of mildly lower generative performance.

The key difference from the previously described FMLM is that
$\mathcal{L}_{\text{diag.}}(\theta; D_t)$ is replaced by $\mathcal{L}_{\text{diag.}}(\theta;
p_{\text{AR}})$, effectively switching the supervision for the diagonal loss from the teacher FLM to
the teacher AR model.
Consequently, we do not need to train GSF-FM as in \Cref{sec:3.1}, since the AR model directly
provides exact velocity supervision (i.e., exact denoiser supervision under the current formulation)
along straight paths.
In summary, we anticipate superior performance compared to FMLM due to three factors: (1) a higher
performance ceiling by a better teacher model, (2) straighter, easier-to-learn target paths, (3)
more linearized decoding-error-based time scheduling.

\section{Experiments}
In this section, we evaluate Gumbel Straight Flow (GSF) on unconditional generation, conditional
generation, and perform several ablations.
For a fair comparison, we use a comparable computational setting, based on a 12-layer bidirectional
transformer, and match both parameter and FLOP budgets as shown in \Cref{tab:perf}.
Please refer to \Cref{sec:set_up} for detailed experimental setup.

\begin{table*}[t]
\setlength{\tabcolsep}{1.5pt}
\scriptsize
\centering
\renewcommand{\arraystretch}{1.09}
\begin{minipage}[t]{0.49\textwidth}
\centering
\caption{Few-step generative performance on OpenWebText among distillation models.
Parentheses indicate teacher types.
``-re'': our re-implementation.}
\label{tab:openwebtext}
\vspace{-2mm}
\begin{tabular}{lccc}
\toprule
Method & GM ($\downarrow$) & PPL ($\downarrow$) & Ent. \\
\midrule
\textcolor{gray}{AR teacher (1024-step)} & \textcolor{gray}{0.24} & \textcolor{gray}{46.5} & \textcolor{gray}{5.59} \\
\midrule
\multicolumn{4}{l}{\textbf{NFE = 1}} \\
SDTT (MDLM) \tcite{deschenaux2025beyond} & - & 1260.8 & 5.26 \\
Di4C (MDLM) \scite{pmlr-v267-hayakawa25a} & - & 1298.8 & 5.29 \\
DCD (Duo) \scite{sahoo2025the}           & - & 5743.2 & 6.02 \\
Di4C (Duo) \scite{sahoo2025the}          & - & 370.5  & 3.92 \\
DFM (FLM) \scite{potaptchik2026discrete} & - & 180.2  & 4.91 \\
GD (AR-Large) \scite{zhang2026gumbel}          & 7.67 & 1475.9 & 5.50 \\
FMLM (FLM) \scite{lee2026flow}           & 1.18 & 168.3  & 5.17 \\
FMLM (FLM)-re       & 1.30 & 140.0  & 4.93 \\
\rowcolor{bgGSF} GSF (AR)            & \textbf{0.94} & \textbf{114.9} & 5.07 \\
\midrule
\multicolumn{4}{l}{\textbf{NFE = 2}} \\
SDTT (MDLM) \tcite{deschenaux2025beyond} & - & 877.2  & 5.34 \\
Di4C (MDLM) \scite{pmlr-v267-hayakawa25a} & - & 758.2  & 5.35 \\
DCD (Duo) \scite{sahoo2025the}           & - & 891.1  & 5.41 \\
Di4C (Duo) \scite{sahoo2025the}          & - & 210.2  & 4.63 \\
DFM (FLM) \scite{potaptchik2026discrete} & - & 152.8  & 5.03 \\
GD (AR-Large) \scite{zhang2026gumbel}          & 4.41 & 763.7  & 5.41 \\
FMLM (FLM) \scite{lee2026flow}           & 1.08 & 133.2  & 5.25 \\
FMLM (FLM)-re         & 0.91 & 118.0  & 5.22 \\
\rowcolor{bgGSF} GSF (AR)            & \textbf{0.74} & \textbf{105.6} & 5.30 \\
\midrule
\multicolumn{4}{l}{\textbf{NFE = 4}} \\
SDTT (MDLM) \tcite{deschenaux2025beyond} & - & 339.7  & 5.38 \\
Di4C (MDLM) \scite{pmlr-v267-hayakawa25a} & - & 239.2  & 5.40 \\
DCD (Duo) \scite{sahoo2025the}           & - & 250.8  & 5.37 \\
Di4C (Duo) \scite{sahoo2025the}          & - & 154.6  & 4.85 \\
DFM (FLM) \scite{potaptchik2026discrete} & - & 122.3  & 5.10 \\
GD (AR-Large) \scite{zhang2026gumbel}          & 1.56 & 354.9  & 5.39 \\
FMLM (FLM) \scite{lee2026flow}           & 1.04 & 111.3  & 5.26 \\
FMLM (FLM)-re                           & 0.92 & 108.2  & 5.24    \\
\rowcolor{bgGSF} GSF (AR)            & \textbf{0.73} & \textbf{99.7}  & 5.32 \\
\bottomrule
\end{tabular}
\par\vspace{-5mm}
\end{minipage}
\hfill
\begin{minipage}[t]{0.49\textwidth}
\centering
\caption{One-step generative performance among distillation models on LM1B dataset.}
\label{tab:lm1b}
\vspace{-2mm}
\begin{tabular}{lccc}
\toprule
Method & GM ($\downarrow$) & PPL ($\downarrow$) & Ent. \\
\midrule
SDTT (MDLM) \tcite{deschenaux2025beyond}  & - & 1429.4 & 4.31 \\
Di4C (MDLM) \scite{pmlr-v267-hayakawa25a}  & - & 1217.1 & 4.38 \\
DCD (Duo) \scite{sahoo2025the}            & - & 1224.5 & 4.33 \\
Di4C (Duo) \scite{sahoo2025the}           & - & 292.9  & 3.79 \\
FMLM (FLM) \scite{lee2026flow}            & 6.16 & 119.3  & 4.16 \\
FMLM (FLM)-re                            & 6.35 & 124.2  & 3.97 \\
\rowcolor{bgGSF} GSF (AR)             & \textbf{5.68} & \textbf{118.6} & 4.05 \\
\bottomrule
\end{tabular}
\vspace{1mm}
\caption{Few-step performance among continuous-state models on TinyStories dataset.}
\label{tab:tinystory}
\vspace{-2mm}
\begin{tabular}{lccc}
\toprule
Method & NFE ($\downarrow$) &PPL  ($\downarrow$) & Ent. \\
\midrule
CANDI \scite{pynadath2026candi}   \;\;\;\;\;\;\;\;\;\;\;\;\;\;\;\;\;\;\;\;    & 128 & 46.32  & 4.04 \\
FLM \scite{lee2026flow}                 & 128 & 57.50  & 4.17 \\
S-FLM \tcite{deschenaux2026language}    & 128 & 95.25  & 4.08 \\
FMLM+ \scite{agarwal2026posterior}      & 8   & 65.89  & 4.16 \\
FMLM (FLM)-re                                   & 8   & 102.89 & 4.06 \\
\rowcolor{bgGSF} GSF (AR)                & 8   & \textbf{41.22} & 4.00 \\
\bottomrule
\vspace{0mm}
\end{tabular}
\includegraphics[width=0.97\linewidth]{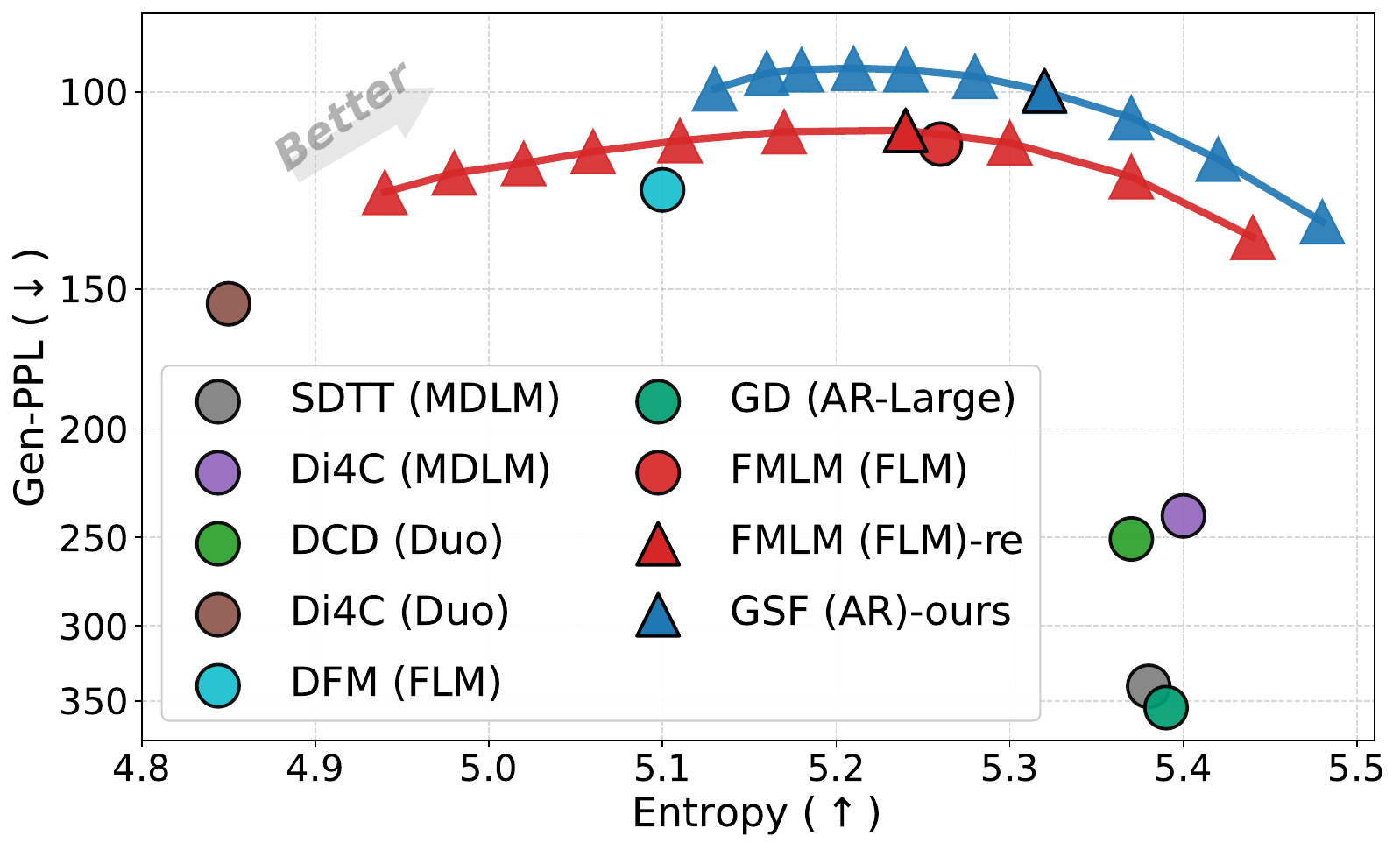}
\vspace{-4mm}
        \captionof{figure}{PPL-entropy trade-offs of models with 4 sampling steps trained on
        OpenWebText.}
        \label{fig:ppl_entropy_tradeoff}
\end{minipage}
\par\vspace{-5mm}
\end{table*}
\subsection{Unconditional Generation}
\paragraph{AR-induced Gumbel coupling produces straighter empirical paths and accelerates flow.}
We first validate the flow matching formulation on unconditional generation of  1024-token sequences
on OpenWebText (OWT)~\citep{Gokaslan2019OpenWeb} under a Gaussianized prior
(\Cref{fig:three_plots}).
GSF-FM achieves better loss convergence during training (\Cref{fig:sub1}) due to the coupling.
To evaluate the straightness, \Cref{fig:sub2} measures a rescaled version of $S(\{\rvx_t^\theta\})$
along generated trajectories: since a straight path implies the denoiser constantly targets the
generated endpoint $\rvx_1$, we evaluate the straightness error in cross-entropy scale via
$\mathrm{CE}(\rvx_1 \,\|\, D_\theta(\rvx_t, t))$.
This results in superior generative performance in the few-step regime as measured by the Gradient
Moment (GM) score~\citep{hoogeboom2026beyond}, which evaluates the difference between expected
parameter gradients on generated versus real data samples judged by GPT-2 Large (\Cref{fig:sub3}).

\begin{figure*}[t]
    \centering
    \renewcommand{\arraystretch}{1.12} \setlength{\tabcolsep}{3.5pt}
    \begin{subfigure}[b]{0.17\textwidth}
        \centering
        \scriptsize
        \begin{tabular}{lc}
            \toprule
            Method & Params ($\downarrow$)  \\
            \midrule
            FMLM  & 169.7M  \\
            GD  &  208.2M   \\
            \rowcolor{bgGSF}GSF  & \textbf{167.8M}   \\
            \midrule
            Method & GFlops ($\downarrow$)  \\
            \midrule
            FMLM  & 370.7   \\
            GD  & 370.7   \\
            \rowcolor{bgGSF}GSF  & \textbf{351.4}   \\
            \bottomrule
        \end{tabular}
        \vspace{2mm}
        \caption{One-step cost}
        \label{tab:perf}
    \end{subfigure}
    \hfill
    \begin{subfigure}[b]{0.4\textwidth}
        \centering
        \includegraphics[width=\linewidth]{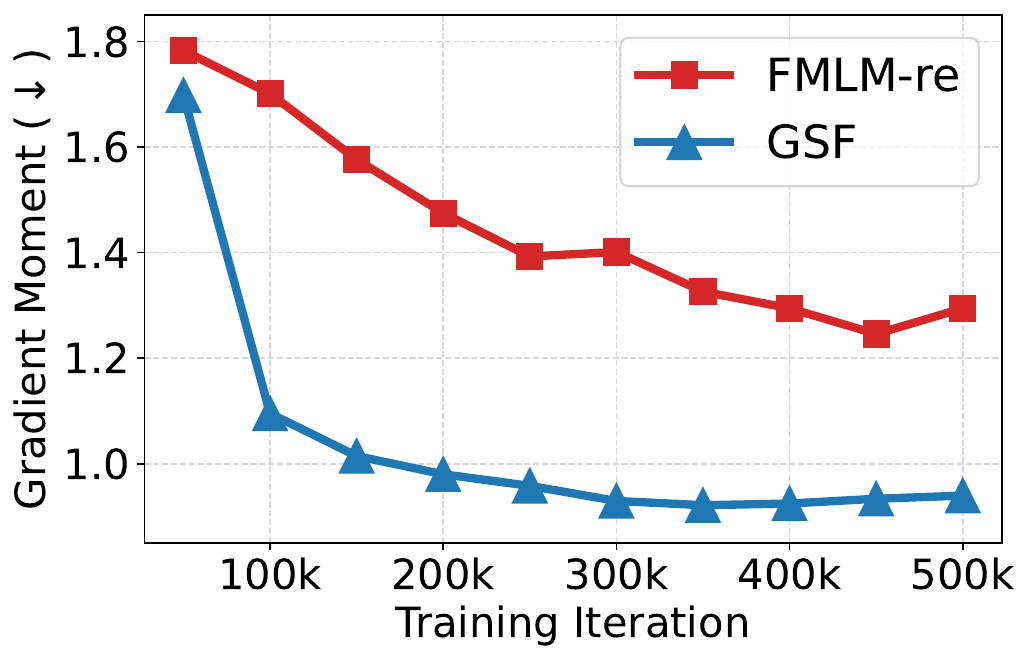}
        \caption{One-step GM per training iteration}
        \label{fig:gm_nfe}
    \end{subfigure}
    \hfill
    \begin{subfigure}[b]{0.4\textwidth}
        \centering
        \includegraphics[width=\linewidth]{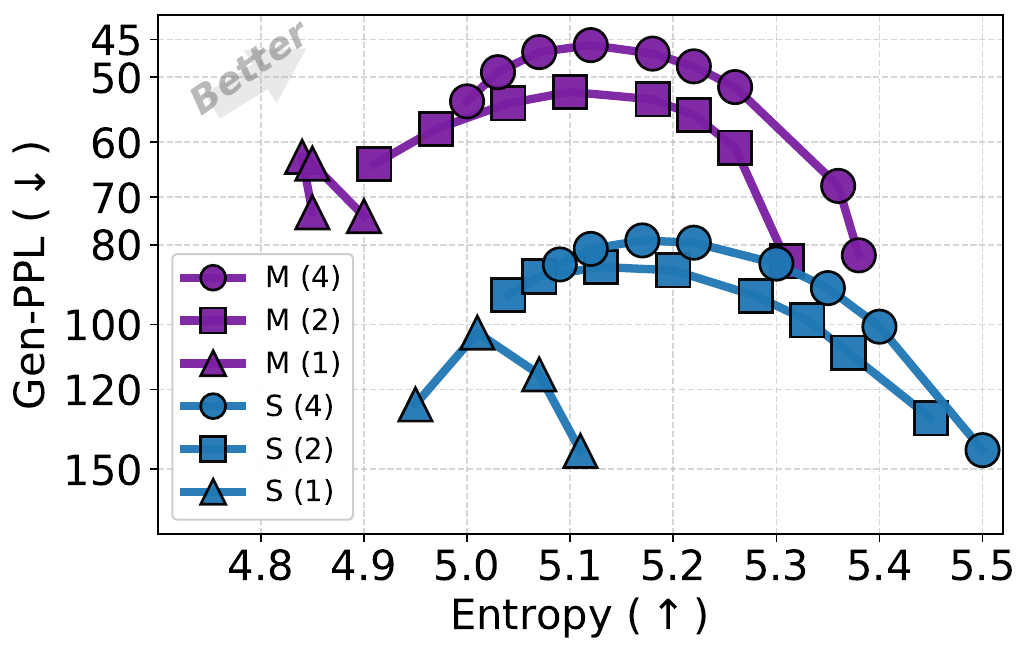}
        \caption{Scaling behavior (S: 168M, M: 419M)}
        \label{fig:scaling}
    \end{subfigure}
    \caption{(a) Computational cost for single-step sampling on OpenWebText.
    (b) One-step generative quality across training iterations.
    (c) Scaling behavior across parameter size and sampling steps}
    \label{fig:main_figure}
\end{figure*}

\paragraph{Few-step distillation performance on OpenWebText.}
Next, we evaluate few-step sampling using the flow map objective across 1, 2, and 4 sampling steps
(\Cref{tab:openwebtext}).
At $\text{NFE}=1$, GSF achieves 114.9 generative perplexity and a gradient moment score of 0.94,
outperforming prior distillation models in this step regime.
Whereas discrete diffusion distillation baselines collapse at single-step sampling (SDTT: 1260.8;
DCD: 5743.2; Di4C (Duo): 370.5), GSF preserves reasonable performance.
We also maintain an edge over FMLM variants (FMLM: 168.3; FMLM-re: 140.0).
GSF's GM score decreases monotonically with additional sampling steps, and regarding training speed,
\Cref{fig:gm_nfe} shows that one-step GM converges faster and improves steadily over the course of
training.
We can examine the effect of straight continuous-path supervision by comparing GSF with GD, since GD
also uses Gumbel noise.
However, GD uses a discrete diffusion backbone and does not explicitly exploit path information,
resulting in limited performance (PPL: 1475.9) compared with GSF.
This may be because directly learning the mapping from Gumbel noise to data is an extremely
difficult supervised problem.
In contrast, the continuous straight path can provide useful intermediate supervision that helps the
model learn this mapping more effectively.

\begin{figure*}[t]
    \centering
    \includegraphics[width=\linewidth]{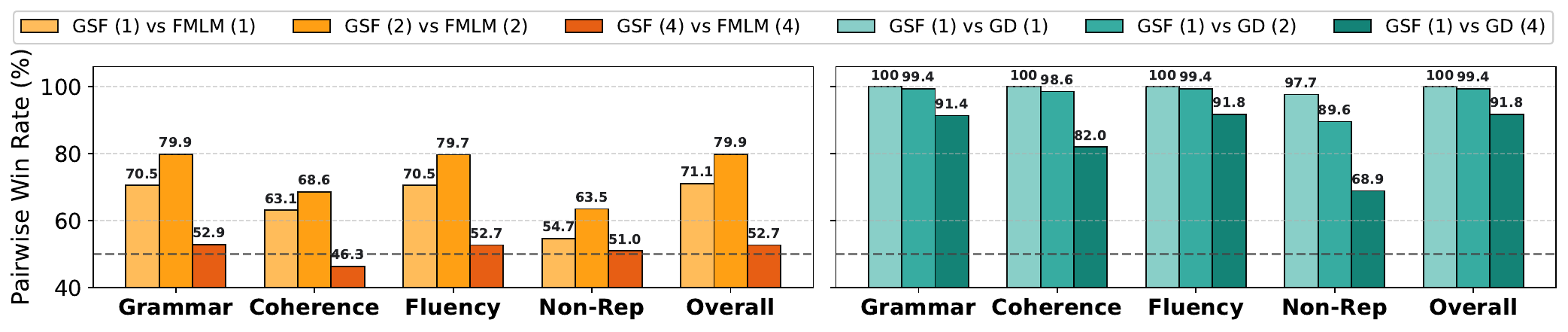}
    \par\vspace{-1mm}
        \caption{Pairwise win rate with Gemini 3.6-Flash.
        Parentheses indicate sampling steps.}
    \label{fig:GeminiJudge}
    \par\vspace{-1mm}
\end{figure*}

Furthermore, as shown in \Cref{fig:ppl_entropy_tradeoff}, GSF establishes a favorable Pareto
trade-off between perplexity and unigram entropy at 4 sampling steps.
Also, \Cref{fig:scaling} shows that generative performance improves with model size (from Small:
168M to Medium: 419M), where the Medium student is distilled from a correspondingly larger and
stronger AR teacher.
We also report LLM-as-a-judge pairwise win rates evaluated by Gemini 3.6-Flash in
\Cref{fig:GeminiJudge} (with prompt in \Cref{sec:app_judge}).
Across sampling steps, GSF outperforms FMLM and GD, where 1-step GSF is preferred in over 91\% of
comparisons even against 4-step GD.
\Cref{sec:app_samples} shows examples of samples of GSF, FMLM, GD, and the corresponding judgment.

Similar results are also obtained on the LM1B dataset~\citep{chelba2013one} (\Cref{tab:lm1b}) and on
the TinyStories dataset~\citep{eldan2023tinystories} (\Cref{tab:tinystory}).
For the latter, GSF reaches 41.22 PPL at ${\text{NFE}=8}$, outperforming few-step flow map baselines
(FMLM+: 65.89; FMLM-re: 102.89) as well as continuous flow models integrated over 128 steps (CANDI:
46.32; FLM: 57.50).

\subsection{Conditional Generation}
\paragraph{Prefix completion on OpenWebText.}
We evaluate conditional text completion.
Models are conditioned on a fixed prefix consisting of the first 50 tokens extracted from held-out
OpenWebText sequences and tasked with generating the continuation up to a total sequence length of
100.
During sampling, the prefix positions are clamped to their true one-hot representations at each
step, while the model generates the continuation tokens in parallel.
Results are shown in \Cref{tab:cond_openwebtext}.
At $\text{NFE}=1$, GSF achieves a conditional perplexity of 97.9 and token entropy of 4.00,
improving upon continuous flow baselines (FMLM: 141.6; FMLM-re: 115.0) and outperforming discrete
diffusion distillation methods (SDTT: 374.7; Di4C (MDLM): 376.8; DCD: 703.1).
With 2 and 4 sampling steps, GSF scales to 77.5 and 71.4 PPL, whereas FMLM-re plateaus near 99.3.
Throughout, GSF maintains stable token entropy ($\approx 4.0$), showing that the learned conditional
trajectories follow the prompt context without degenerate repetition.
\Cref{tab:cond_gen_owt_286279348_150k} shows examples of samples for various prefixes.


\paragraph{Abstractive summarization on XSum.}
We next evaluate task-specific sequence-to-sequence generation on the BBC XSum abstractive
summarization benchmark~\citep{narayan2018dont}, following the protocol of \citet{hu2026elf}.
Given a multi-sentence source document, models are conditioned to generate a concise single-sentence
summary of up to 64 tokens.
In \Cref{tab:xsum}, we report ROUGE-1, ROUGE-2, and ROUGE-L scores~\citep{lin-2004-rouge} across
sampling steps.
At $\text{NFE}=1$, GSF achieves 29.5 / 9.1 / 23.4 ROUGE-1/2/L,  outperforming the single-step
continuous flow baseline ELF (23.0 / 5.5 / 18.9).
Remarkably, a single forward pass of GSF ($\text{NFE}=1$) already surpasses multi-step diffusion
models requiring dozens to thousands of iterations, such as E2D2 (28.4 / 8.3 / 22.0 at $\approx 64$
steps) and SeqDiffu (19.3 / 1.7 / 14.1 at 2000 steps).
With only two sampling steps ($\text{NFE}=2$), GSF reaches 31.2 / 9.6 / 25.0, substantially
narrowing the gap to its 64-step autoregressive teacher (34.1 / 12.1 / 26.2), particularly in
ROUGE-1 and ROUGE-L.
\Cref{tab:xsum_qual_290645657_1_2step} shows examples of sentences summarized by GSF.

\begin{table}[t]
\centering
\begin{minipage}[c]{0.50\textwidth}
\setlength{\tabcolsep}{3pt}
    \centering
    \scriptsize
    \caption{Few-step conditional performance among distillation models on OpenWebText.}
    \label{tab:cond_openwebtext}
    \begin{tabular}{lccc}
\toprule
 Method&NFE & PPL ($\downarrow$) & Ent. \\
\midrule
SDTT (MDLM)~\tcite{deschenaux2025beyond} & 1&374.7 & 4.09 \\
Di4C (MDLM)~\tcite{pmlr-v267-hayakawa25a} &1& 376.8 & 4.10 \\
DCD (Duo)~\tcite{sahoo2025the} &1& 703.1 & 4.18 \\
Di4C (Duo)~\tcite{sahoo2025the}  &1& 451.2 & 4.08 \\
FMLM (FLM)~\tcite{lee2026flow} &1& 141.6 & 4.14 \\
FMLM (FLM)-re&1& 115.0 & 3.89 \\
\rowcolor{bgGSF}GSF (AR) &1& \textbf{97.9} & 4.00 \\
\midrule
FMLM (FLM)-re&2& 101.4 & 3.98 \\
\rowcolor{bgGSF} GSF (AR) &2& \textbf{77.5} & 4.05 \\
\midrule
FMLM (FLM)-re &4& 99.3 & 4.04 \\
\rowcolor{bgGSF} GSF (AR) &4& \textbf{71.4} & 4.06 \\
\bottomrule
\end{tabular}
\end{minipage}\hfill
\begin{minipage}[c]{0.48\textwidth}
\setlength{\tabcolsep}{1.5pt}
    \centering
    \scriptsize
    \caption{Few-step conditional generation on XSUM summarization (ROUGE: ($\uparrow$)).}
    \label{tab:xsum}
    \begin{tabular}{lcccc}
    \toprule
    \multirow{1}{*}{Method} & \multirow{1}{*}{NFE} & ROUGE-1  & ROUGE-2  & ROUGE-L \\
    \midrule
    AR~\tcite{hu2026elf} &  64 & 30.5  & 10.2  & 24.4 \\
    AR-re (our teacher) &  64 & 34.1  & 12.1  & 26.2 \\
    ELF~\tcite{hu2026elf} &  64 & 36.0  & 12.2  & 27.8 \\
    E2D2~\tcite{arriola2025encoderdecoder}   & $\approx$ 64 & 28.4  & 8.3  & 22.0  \\
    SeqDiffu~\tcite{yuan2024text}  & 2000 & 19.3 & 1.7 & 14.1 \\
    \midrule
    ELF~\tcite{hu2026elf}  & 1 & 23.0 & 5.5 & 18.9 \\
    FMLM-re  & 1 & 16.7 & 1.2 & 13.1 \\
    \rowcolor{bgGSF} GSF (AR) & 1 & \textbf{29.5} & \textbf{9.1} & \textbf{23.4} \\
    \midrule
    ELF~\tcite{hu2026elf}  & 2 & 27.6 & 7.7 & 22.5 \\
    FMLM-re  & 2 & 16.7 & 1.2 & 13.4 \\
    \rowcolor{bgGSF} GSF (AR)  & 2 & \textbf{31.2} & \textbf{9.6}& \textbf{25.0} \\
    \bottomrule
    \end{tabular}
\end{minipage}
\end{table}

\subsection{Ablation Study}
\paragraph{Prior distribution and noise scheduling.}
We investigate the impact of prior choice and time reparameterization for single-step generation on
OpenWebText (\Cref{tab:ablation_prior}).
Using the raw Standard Gumbel prior yields poor gradient moment score ($\text{GM}=1.76$,
$\text{PPL}=133.5$), which we hypothesize is due to the strong right skew of extreme-value
distributions (\Cref{fig:gumbel_stat}).
Transforming the prior to a standard Gaussian via \Cref{prop:gaussianization} substantially improves
GM ($\text{GM}=1.05$ under linear scheduling).
Next, we compare noise schedules: while standard DER parameterization achieves 113.8 PPL, it incurs
a higher GM score (1.26) and lower token entropy (4.84).
Our Coupled DER schedule (\Cref{fig:coupled_der}) obtains the best overall balance:
$\text{GM}=0.94$, $\text{PPL}=114.9$, and $\text{Ent}=5.07$.

\paragraph{Training and architectural configurations.}
\Cref{tab:ablation_training} examines the impact of student architecture and training configurations
for 1-step generation on OpenWebText: (i) \emph{Attention masking}: Replacing bidirectional
self-attention with causal masking drastically degrades performance ($\text{PPL}=211.0$,
$\text{GM}=1.62$).
While the causal dependencies are enough to model straight trajectories defined by the
autoregressive teacher, parallel non-autoregressive generation benefits from bidirectional context:
the teacher's Gumbel coupling provides sequence-level target supervision, but bidirectional
self-attention enables the student to coordinate across all tokens simultaneously.
(ii) \emph{Teacher weight initialization}: Training a randomly initialized student increases
perplexity from 114.9 to 145.4, showing that transferring pretrained AR teacher representations
provides an effective initialization for flow map distillation.
(iii) \emph{Semigroup loss coupling}: While GSF by default optimizes the off-diagonal semigroup loss
using random intermediate pairs ($\text{PPL}=114.9$), restricting the off-diagonal transitions
strictly along teacher-coupled paths (\emph{Coupled off-diag.}) further lowers perplexity to 108.4
and GM to 0.89, with a slight decrease in token entropy (5.01 vs 5.07).
(iv) \emph{Denoiser output parameterization}: Predicting logits instead of one-hot targets is
slightly worse on both headline metrics ($\text{GM}=0.99$, $\text{PPL}=125.7$), supporting the
one-hot formulation used throughout.

\begin{table}[t]
\centering
\begin{minipage}[c]{0.485\textwidth}
\setlength{\tabcolsep}{3pt}
    \centering
    \scriptsize
    \caption{Ablation of prior and noise schedules for one-step generation on OpenWebText.}
    \label{tab:ablation_prior}
    \vspace{-3pt}
    \begin{tabular}{llccc}
\toprule
 Prior Distribution&Schedule&GM ($\downarrow$) & PPL ($\downarrow$) & Ent. \\
\midrule
Standard Gumbel & Linear&1.76&133.5 & 4.87 \\
Normalized Gumbel & Linear&1.24 & 249.0 & 5.21 \\
Gaussian & Linear&1.05 &150.7 & 5.04 \\
Gaussian &DER& 1.26 & 113.8 & 4.84 \\
\rowcolor{bgGSF} Gaussian  &Coupled DER& 0.94&114.9 & 5.07 \\
\bottomrule
\end{tabular}
\end{minipage}\hfill
\begin{minipage}[c]{0.485\textwidth}
\setlength{\tabcolsep}{3pt}
    \centering
    \scriptsize
    \caption{Ablation of training configurations for one-step generation on OpenWebText.}
    \label{tab:ablation_training}
    \vspace{-3pt}
    \begin{tabular}{llccc}
\toprule
 Default &Change&GM ($\downarrow$) & PPL ($\downarrow$) & Ent. \\
\midrule
Bidirectional attn. & Causal attn.&1.62&211.0 & 5.05 \\
AR init. & Random init.&0.81 & 145.4& 5.14 \\
Random off-diag. & Coupled off-diag.&0.89 &108.4 & 5.01 \\
One-hot pred. &Logit pred.& 0.99 & 125.7 & 5.06 \\
\rowcolor{bgGSF} No change && 0.94&114.9 & 5.07 \\
\bottomrule
\end{tabular}
\end{minipage}
\par\vspace{-3mm}
\end{table}

\section{Conclusion}
In this work, we introduced Gumbel Straight Flow (GSF), demonstrating that autoregressive models
naturally provide optimal, straight trajectories between continuous noise and discrete data tokens.
By supervising flow maps along these non-intersecting paths, GSF improves few-step parallel
generation across unconditional and conditional benchmarks, offering a promising framework for
distilling autoregressive models into fast parallel samplers.
In \Cref{sec:app_discussion} we discuss several limitations.

\subsection*{AI use statement}
In this work, we used generative AI tools for implementing methods.
We have not used generative AI tools for generating synthetic data sets, help develop theoretical
models or conceptual frameworks, formulate mathematical claims, provide critical ingredients for
proving mathematical claims, assist in the writing of proofs, propose or refine hypotheses, design
or provide feedback on research  methodology or experiments, assist with translation, clean and
reformat dataset, support qualitative and thematic data analysis, interpret results.
Additionally, we used generative AI to polish the writing.
We have reviewed all AI-assisted work by carefully human-reviewing the code used to run the
experiments.
We take responsibility for the final content of this work, including text, claims or artifacts
produced with the aid of generative AI.

\bibliography{main}

\appendix
\input{appendix}

\end{document}

%% file: math_commands.tex
\usepackage{amsmath,amsfonts,bm}

\def\eqref#1{equation~\ref{#1}}

\def\1{\bm{1}}

\def\rvf{{\mathbf{f}}}

\def\rvu{{\mathbf{i}}}

\def\rvl{{\mathbf{l}}}

\def\rvu{{\mathbf{u}}}
\def\rvv{{\mathbf{v}}}

\def\rvx{{\mathbf{x}}}
\def\rvy{{\mathbf{y}}}

\DeclareMathAlphabet{\mathsfit}{\encodingdefault}{\sfdefault}{m}{sl}
\SetMathAlphabet{\mathsfit}{bold}{\encodingdefault}{\sfdefault}{bx}{n}

%% file: appendix.tex
\newpage
\section{Proof and Derivations}
\label{sec:A}
\subsection{Properties of Gumbel noise}
\label{sec:A.1}

\paragraph{Ranking Preserving.}
For any single-token Gumbel sample $\rvxi \in \mathbb{R}^{|\mathcal{V}|}$ and its corresponding one-hot target $\rvx \in \mathbb{R}^{|\mathcal{V}|}$, the selected index preserves the Gumbel-max ranking:
\[
\arg\max_{j\in\{1,\dots,|\mathcal{V}|\}} (\ell_j + \xi_j) = \arg\max_{j\in\{1,\dots,|\mathcal{V}|\}} (x_j).
\]

\paragraph{Gumbel Residual Ordering.}
Consider two coupled pairs $(\rvxi^1, \rvx^1)$ and $(\rvxi^2, \rvx^2)$ with distinct targets $\rvx^1 \neq \rvx^2$. Let $k^1$ and $k^2$ ($k^1 \neq k^2$) be their respective active indices:
\[
k^1 := \arg\max_{j\in\{1,\dots,|\mathcal{V}|\}} (x_j^1), \qquad k^2 := \arg\max_{j\in\{1,\dots,|\mathcal{V}|\}} (x_j^2).
\]
By the ranking preserving property, we have:
\[
\xi^{1}_{k^1} + \ell_{k^1} > \xi^{1}_{k^2} + \ell_{k^2} \quad \text{and} \quad \xi^{2}_{k^1} + \ell_{k^1} < \xi^{2}_{k^2} + \ell_{k^2}.
\]
Rearranging both inequalities yields $\xi^{1}_{k^1} - \xi^{1}_{k^2} > \ell_{k^2} - \ell_{k^1} > \xi^{2}_{k^1} - \xi^{2}_{k^2}$, which implies:
\[
\xi^{1}_{k^1} - \xi^{2}_{k^1} > \xi^{1}_{k^2} - \xi^{2}_{k^2}.
\]

\subsection{Proof of \texorpdfstring{\cref{thma}}{Theorem 1}}
\label{sec:A.2}
\thma*

\paragraph{Proof of Theorem 1.}
We first prove the single-token case ($L=1$), where each state is
in $\mathbb{R}^{|\mathcal V|}$, the extension to sequences being straightforward. The extension to arbitrary sequence length is given in Proposition~\ref{prop:sequence_extension}. Let $\rvv_{\theta^*}(\rvx_t, t) = \mathbb{E}[\rvx_1 - \rvx_0 \mid \rvx_t]$ be the optimal velocity for the linear interpolation $\rvx_t = (1-t)\rvx_0 + t \rvx_1$. The straightness measure
\[
S(\{\rvx_t^{\theta^*}\}) := \int_0^1 \mathbb{E}\left[\|(\rvx_1 - \rvx_0) - \rvv_{\theta^*}(\rvx_t,t)\|_2^2\right]dt
\]
equals zero if the posterior $p(\rvx_0, \rvx_1 \mid \rvx_t)$ is a Dirac-delta distribution, i.e., no two distinct coupled pairs intersect at any $\rvx_t$ for $t \in (0, 1)$.

Suppose for contradiction that two distinct pairs $(\rvx_0^1, \rvx_1^1) = (\rvxi^1, \rvx^1)$ and $(\rvx_0^2, \rvx_1^2) = (\rvxi^2, \rvx^2)$ intersect at $\rvx_t$:
\begin{equation}
\rvx_t = (1-t)\rvxi^1 + t \rvx^1 = (1-t)\rvxi^2 + t \rvx^2 \implies \rvxi^1 - \rvxi^2 = \frac{t}{1-t}(\rvx^2 - \rvx^1). \label{eq:intersect}
\end{equation}
If $\rvx^1 = \rvx^2$, then $\rvxi^1 = \rvxi^2$, contradicting that the pairs are distinct; hence $\rvx^1 \neq \rvx^2$. Let $k^1 := \arg\max_j x_j^1$ and $k^2 := \arg\max_j x_j^2$ ($k^1 \neq k^2$) be the active indices of the one-hot targets $\rvx^1$ and $\rvx^2$. Since $x^2_{k^1} - x^1_{k^1} = -1$ and $x^2_{k^2} - x^1_{k^2} = +1$, evaluating \cref{eq:intersect} at indices $k^1$ and $k^2$ for $t \in (0, 1)$ yields:
\[
\xi^1_{k^1} - \xi^2_{k^1} = -\frac{t}{1-t} < 0 < +\frac{t}{1-t} = \xi^1_{k^2} - \xi^2_{k^2} \implies \xi^1_{k^1} - \xi^2_{k^1} < \xi^1_{k^2} - \xi^2_{k^2}.
\]
This contradicts the \textbf{Gumbel Residual Ordering} property ($\xi^1_{k^1} - \xi^2_{k^1} > \xi^1_{k^2} - \xi^2_{k^2}$). Thus, no two trajectories intersect, $p(\rvx_0, \rvx_1 \mid \rvx_t)$ is a Dirac-delta distribution, and $S(\{\rvx_t^{\theta^*}\}) = 0$. \hfill $\square$

For completeness, we provide the proof for arbitrary sequence length $L$. 
\begin{proposition}
\label{prop:sequence_extension}
The single-token non-intersection result above extends to arbitrary
sequence length $L$. In particular, for every fixed $t\in(0,1)$,
the map
\begin{equation}
\rvxi_{1:L} \mapsto (1-t)\rvxi_{1:L} + t\,\operatorname{onehot}\!\left(T_{\mathrm{AR}}(\rvxi_{1:L})\right)
\end{equation}
is injective.
\end{proposition}

\begin{proof}
The result can be derived by induction over the sequence positions.
At position $i=1$, both samples have the same logit vector $\rvl_1$,
so the single-token result established above implies equality of both the first
Gumbel block and the first generated token.
Assume that that the two generated prefixes agree up to position
$i-1$, i.e., $\rvy^1_{<i}=\rvy^2_{<i}$. Then the AR model produces obviously the same conditional logits, $\rvl_i(\rvy^1_{<i})=\rvl_i(\rvy^2_{<i})$. Equality of the $i$th blocks of the interpolated states, together again
with the single-token result above, therefore implies $\rvxi^1_i=\rvxi^2_i$ and $y^1_i=y^2_i$.
\end{proof}

\subsection{Proof of \texorpdfstring{\cref{propositiona}}{Proposition 2}}
\label{sec:A.3}
\propositiona*

\paragraph{Proof of Proposition 2.}
Consider the single-token case ($L=1$) (it easily extends to the multi-token case too). Let $u_j = f(\xi_j)$ for all $j \in \{1, \dots, |\mathcal{V}|\}$, where $f: \mathbb{R} \to \mathbb{R}$ is strictly increasing and invertible. Suppose for contradiction that two distinct coupled pairs $(\rvx_0^1, \rvx_1^1) = (\rvu^1, \rvx^1)$ and $(\rvx_0^2, \rvx_1^2) = (\rvu^2, \rvx^2)$ with $\rvx^1 \neq \rvx^2$ intersect at $\rvx_t$ for $t \in (0, 1)$:
\[
\rvx_t = (1-t)\rvu^1 + t \rvx^1 = (1-t)\rvu^2 + t \rvx^2 \implies \rvu^1 - \rvu^2 = \frac{t}{1-t}(\rvx^2 - \rvx^1).
\]
Let $k^1 := \arg\max_j x_j^1$ and $k^2 := \arg\max_j x_j^2$ ($k^1 \neq k^2$) be the active indices of the one-hot targets $\rvx^1$ and $\rvx^2$. Evaluating the above equality at coordinates $k^1$ and $k^2$ yields:
\[
u^1_{k^1} - u^2_{k^1} = -\frac{t}{1-t} < 0 \implies u^1_{k^1} < u^2_{k^1}, \qquad
u^1_{k^2} - u^2_{k^2} = +\frac{t}{1-t} > 0 \implies u^1_{k^2} > u^2_{k^2}.
\]
Since $f^{-1}$ is also strictly increasing, applying $f^{-1}$ to both inequalities gives:
\[
\xi^1_{k^1} < \xi^2_{k^1} \quad \text{and} \quad \xi^1_{k^2} > \xi^2_{k^2} \implies \xi^1_{k^1} - \xi^2_{k^1} < 0 < \xi^1_{k^2} - \xi^2_{k^2}.
\]
This contradicts the \textbf{Gumbel Residual Ordering} property ($\xi^1_{k^1} - \xi^2_{k^1} > \xi^1_{k^2} - \xi^2_{k^2}$). Thus, no two trajectories intersect under the transformed prior, $p(\rvx_0, \rvx_1 \mid \rvx_t)$ remains a Dirac-delta distribution, and $S(\{\rvx_t^{\theta^*}\}) = 0$. \hfill $\square$

\subsection{An Optimal Transport Interpretation of Gumbel Straight Flow}
\label{sec:app_ot}
We can understand the straightness and non-intersection properties of GSF by reformulating it  as a (semi-discrete) optimal transport (OT) problem~\citep[chapter 5]{gabriel2019computational}. 

In a standard flow matching context where the noise distribution $p_0(\rvx_0)=p_{\text{noise}}(\rvx_0)$ and data distribution $p_1(\rvx_1)=p_{\text{data}}(\rvx_1)$ are both distributions with densities in $\mathbb{R}^d$, we typically couple/pair the noise and samples independently; this leads to a generative ODE with curved paths. To obtain straight paths, one can couple the noise and data distributions using the joint distribution $p(\rvx_0,\rvx_1)$ with marginals $p_{\text{noise}}(\rvx_0)$ and $p_{\text{data}}(\rvx_1)$ minimizing $\mathbb{E}[||\rvx_0-\rvx_1||^2]$ \citep{mccann1997convexity}. This is an optimal transport problem whose solution is of the form $p_{\text{OT}}(\rvx_0,\rvx_1)=p_{\text{noise}}(\rvx_0) \delta_{T_{\textup{OT}}(\rvx_0)}(\rvx_1)$ where $T_{\textup{OT}}(\cdot)$ is the optimal transport map. Computing this joint distribution is typically intractable and requires approximations; see e.g. \citep{liu2023flow,pooladian2023multisample,tong2024improving}.

We are here in a specific scenario where the data distribution is an atomic distribution. Again one can obtain straight transport paths if we could identify the joint distribution minimizing the expected quadratic cost.  This is a so-called semi-discrete optimal transport problem~\citep{gabriel2019computational}. The semi-discrete nature of the problem allows us to characterize precisely the structure of the transport.

\paragraph{Single Token.}
For a single token, we have that $p_0(\rvx_0)$ is given by the product of $|\mathcal{V}|$ Gumbel random variables while $p_1(\rvx_1)=\sum_{i=1}^{|\mathcal{V}|} p_{1,i}\delta_{\mathbf{e}_i}(\rvx_1)$ where $\mathbf{e}_i$ is the one-hot vector associated to the $i^{\textup{th}}$ element of the vocabulary. We rewrite the optimal transport problem for a quadratic cost as $p_{\text{OT}}(\rvx_0,\rvx_1)=p_{\text{noise}}(\rvx_0) \delta_{T_{\textup{OT}}(\rvx_0)}(\rvx_1)$ as
\begin{equation}
T_{\textup{OT}}=\arg \inf_{T: T \# p_0=p_1} \int \|\rvx_0 -T(\rvx_0)\|^2 p_0(\rvx_0) \dd \rvx_0.
\end{equation}

The transport map of a semi-discrete OT can be obtained from a $|\mathcal{V}|$-dimensional vector $\mathbf{a}:=(a_i)_{i \in \mathcal{V}}$ referred to as dual weights, which we will define later; see \citep[chapter 5]{gabriel2019computational} for a detailed introduction. We first map noise to data index from the mapping 
\begin{equation}
\kappa_{\mathbf{a}}(\rvx_0) = \arg\min_{i \in \{1, \dots, |\mathcal{V}|\}} \bigl\{ \tfrac{1}{2} \|\rvx_0 -\mathbf{e}_i\|^2 - a_i \bigr\}, 
\end{equation}
and then define the corresponding transport map through $T_{\mathbf{a}}(\rvx_0) =\mathbf{e}_{\kappa_{\mathbf{a}}(\rvx_0)}$.

Given $\|\rvx_0 -\mathbf{e}_i\|^2=\|\rvx_0\|^2-2 \rvx_0^\top \mathbf{e}_i +\|\mathbf{e}_i\|^2=\|\rvx_0\|^2-2 x_{0,i} +1$, we thus have $\kappa_{\mathbf{a}} (\rvx_0)= \arg\max_{i \in \{1, \dots, |\mathcal{V}|\}} \bigl\{x_{0,i}+a_i\bigr\}$ and the transport map simplifies to 
\begin{equation}\label{eq:semiOTmap}
    T_{\mathbf{a}}(\rvx_0) = e_{\arg \max_i \{ x_{0,i} + a_i \}}.
\end{equation}

By construction, this transport map thus partitions the noise space into $|\mathcal{V}|$ so-called Laguerre cells given by
\begin{equation}\label{eq:Laguerrecell}
C_i(\mathbf{a}) = \left\{ \rvx_0 \in \mathbb{R}^{|\mathcal{V}|} : x_{0,i} + a_i \ge x_{0,m} + a_m, \; \forall m \right\}.
\end{equation}

Finally the optimal transport map is obtained by selecting the dual weights $\mathbf{a}$\footnote{These weights are determined up to an additive constant} such that 
\begin{equation}\label{eq:probaconstraint}
\int \mathbb{I}(\rvx_0 \in C_i(\mathbf{a})) p_0(\rvx_0) \dd \rvx_0=p_{1,i}.
\end{equation}
In general semi-discrete optimal transport~\citep{gabriel2019computational}, finding these weights requires expensive numerical optimization during training; see, e.g., \cite{kong2026alignflow}. In our scenario, this is not necessary as established in the following proposition.

\begin{proposition}
\label{prop:gumbel_ot_weights}
Let $\rvx_0=\rvxi$ for $\xi_i \overset{\mathrm{i.i.d.}}{\sim} \mathcal{G}(0,1)$ where $i \in \mathcal{V}$. For the target probabilities, $p_{1,i}= \operatorname{softmax}(\boldsymbol \ell)_i$, the dual weights $\mathbf{a}$ satisfying (\ref{eq:probaconstraint}) are given by
\begin{equation}
a_i=\ell_i+c,
\qquad i\in\mathcal V,
\end{equation}
for an arbitrary constant \(c\in\mathbb R\). Consequently, the optimal transport map is defined by 
\begin{equation}\label{eq:Gumbelmaxtransport}
T_{\boldsymbol\ell}(\rvxi) =
\mathbf e_{\arg\max_{i\in\mathcal V}\{\xi_i+\ell_i\}};
\end{equation}
i.e. the Gumbel--max map is the optimal transport map from $p_0$ to $p_1$.
\end{proposition}
\begin{proof}
The proof is direct. By \eqref{eq:Laguerrecell} and the Gumbel--max identity,
\begin{equation}
\mathbb P(\rvxi \in C_i(\mathbf a))=\mathbb P(i =\arg \max_m \{\xi_m+a_m\})=\operatorname{softmax}(\mathbf a)_i.
\end{equation}
Hence \eqref{eq:probaconstraint} requires $\operatorname{softmax}(\mathbf a)=\operatorname{softmax}(\boldsymbol\ell)$, which is equivalent to $\mathbf a=\boldsymbol\ell+c\mathbf 1$. Substituting into \eqref{eq:semiOTmap} gives the Gumbel--max transport map of \eqref{eq:Gumbelmaxtransport}.
\end{proof}

In other words, the teacher logits $\boldsymbol\ell$ are already the exact semi-discrete OT dual weights so no optimization or OT solver is needed. This result is already implicit in \cite[Chapter 5]{galichon2018optimal}.

The OT viewpoint gives a simple explanation of the
non-intersection property. Since $T_{\boldsymbol\ell}$ is the quadratic
optimal transport map, it is monotone~\citep{brenier1991polar}, i.e.
\begin{equation}
\left\langle
T_{\boldsymbol\ell}(\boldsymbol\xi)
-T_{\boldsymbol\ell}(\boldsymbol\xi'),
\boldsymbol\xi-\boldsymbol\xi'
\right\rangle
\geq 0.
\end{equation}
So for the interpolated states
\begin{equation}
\rvx_t=(1-t)\boldsymbol\xi+tT_{\boldsymbol\ell}(\boldsymbol\xi),
\qquad
\rvx_t'=(1-t)\boldsymbol\xi'
+tT_{\boldsymbol\ell}(\boldsymbol\xi'),
\end{equation}
we therefore have
\begin{equation}
\left\langle
\rvx_t-\rvx_t',
\boldsymbol\xi-\boldsymbol\xi'
\right\rangle
\geq
(1-t)\|\boldsymbol\xi-\boldsymbol\xi'\|^2.
\end{equation}
For \(t<1\), the right-hand side is strictly positive whenever
\(\boldsymbol\xi\neq\boldsymbol\xi'\). Hence two different trajectories
cannot meet at the same time. This recovers the non-intersection property
proved directly in \Cref{sec:A.2}. 

\paragraph{Gaussianization.}
In practice, we transform the Gumbel noise component-wise using the strictly increasing map
\begin{equation}
    \rvx_0=f(\rvxi) =\Phi^{-1}\!\left(\exp(-e^{-\rvxi})\right),
\end{equation}
so that $\rvx_0 \sim\mathcal N(0,I)$ (but leave the  Gumbel noise-token coupling unchanged). Since $f$ is strictly increasing and invertible, the non-intersection property of the linear paths is preserved, as established in \Cref{sec:A.3}. The OT interpretation is slightly modified: in this Gaussian representation the same coupling is optimal for the cost
\begin{equation}
    c(\rvx_0,\rvx_1)=\left\| f^{-1}(\rvx_0)-\rvx_1\right\|^2=  \left\| \rvxi -\rvx_1\right\|^2,
\end{equation}
rather than the Euclidean quadratic cost
$\|\rvx_0-\rvx_1\|^2$.

\paragraph{Multiple Tokens.}
The single-token result can be extended directly to a token sequence generated by an AR model. At position $i$, conditional on the previously generated tokens $y_{<i}$, the logits $\boldsymbol\ell_i(\rvy_{<i})$ are fixed and
\begin{equation}
    y_i = \arg\max_k \big\{ \xi_{i, k} +\ell_{i, k}(\rvy_{<i}) \big\}.
\end{equation}
Hence, this is exactly the single-token semi-discrete OT problem above: the dual weights are the
conditional teacher logits $\boldsymbol\ell_i(y_{<i})$ (up to an additive constant). The AR teacher
therefore constructs the sequence coupling by applying these single-token
transport maps successively.

However, the resulting sequence map is generally not the quadratic optimal transport map on $\mathbb R^{L\times|\mathcal V|}$, but the paths remain non-intersecting as established in \Cref{sec:A.2}.

\section{Related Work}
\label{sec:related_work}
Accelerated parallel sampling for language models broadly spans three paradigms: (i) distilling and accelerating multi-step discrete diffusion models~\citep{deschenaux2025beyond,pmlr-v267-hayakawa25a,Zhu_2025_ICCV,li2026idlm,hoogeboom2026beyond,zheng2026ultra,israel2026accelerating}; (ii) continuous-space generative modeling, which embeds categorical tokens into continuous Euclidean or spherical geometries~\citep{li2022diffusion,dieleman2022continuous,deschenaux2026language,yoo2026self}; (iii) parallelizing AR models~\citep{pmlr-v235-kou24a,hu2026fast}. 

Diffusion or flow matching models can be integrated into few-step samplers via flow maps~\citep{boffi2026build}. 
Extending diffusion and flow maps to discrete categorical representations and language models~\citep{eijkelboom2024variational,lee2026flow,roos2026categorical,hu2026elf} embeds tokens into continuous spaces. 
Nonetheless, these models universally rely on independent couplings $p(x_0, x_1) = p(x_0) p(x_1)$, causing straight paths connecting random pairs to cross in high dimensions and forcing the velocity field to curve. Rectified Flow~\citep{liu2023flow} showed that straighter trajectories permit better generation with coarse numerical integration.

The key distinction between GSF and these methods is the use of teacher-guided noise-data pairs as pioneered by, e.g.~\citep{draxler2026parallel}.
Gumbel Distillation~\citep{zhang2026gumbel} directly utilizes Gumbels sampled from an AR teacher's posterior distribution to supervise a student model. However, Gumbel Distillation binds this pairing to a discrete diffusion backbone, which requires iterative categorical transitions and multiple sampling steps.
GSF sits at the intersection of continuous flow matching and AR distillation. Standard Rectified Flow~\citep{liu2023flow} can iteratively apply computationally expensive Reflow steps to obtain increasingly straight paths. 
It uses this insight as a strong inductive bias for learning the noise-data pairing using existing flow map techniques.

\section{Experimental Details}
\label{sec:set_up}
\textbf{Model architecture.}
For all OpenWebText experiments, we follow the architecture scale used in the baseline methods~\citep{lee2026flow,sahoo2025the}, which use a bidirectional Diffusion Transformer~\citep{peebles2023scalable} with 12 layers and a hidden size of 768. We use multi-query attention~\citep{shazeer2019fast} with 3 query heads and a head dimension of 256, along with FiLM time conditioning~\citep{perez2018film}, which makes the model slightly more efficient, as shown in \Cref{tab:perf}. The  AR teacher uses causal attention and does not use time embeddings or a separate unembedding layer. Both GSF-FM (\Cref{fig:three_plots}) and GSF (flow map) use an unembedding layer, with one and two time-conditioning layers, respectively. For the medium-sized model in \Cref{fig:scaling}, both the AR teacher and GSF student use 24 layers, a hidden size of 1024, 4 query heads, and a head dimension of 256. We use the \texttt{GPT-2} tokenizer with $|\vocab|=50,257$ and $L=1024$, together with the corresponding embedding layer. For the LM1B dataset in \Cref{tab:lm1b}, the only difference is that we use the \texttt{bert-base-uncased} tokenizer with $|\vocab|=30,552$ and $L=128$, which results in a different number of embedding-layer parameters. For the TinyStories dataset in \Cref{tab:tinystory}, we use the \texttt{GPT-2} tokenizer with $|\vocab|=50,257$ and $L=128$. We also use 8 layers, a hidden size of 512, 2 query heads, and a head dimension of 256 to match the baseline setting in \cite{agarwal2026posterior}. For the conditional OpenWebText experiments in \Cref{tab:cond_openwebtext}, we use the same setting as for unconditional OpenWebText, except for the sequence length (50 prefix tokens and 50 continuation tokens), following \cite{lee2026flow}. For the conditional XSUM experiments in \Cref{tab:xsum}, we use the same setting as unconditional OpenWebText with 1024 prefix tokens and 64 continuation tokens. Here, we apply time conditioning in a token-wise manner, following \cite{hu2026elf}. That is, we always set $t=1$ for the prefix tokens.

\begin{algorithm}[t]
\caption{GSF training with Gumbel posterior sampling}
\label{alg:fmlm_training}
\begin{algorithmic}[0]
\Require Dataset $\mathcal{D}$, AR teacher $p_{\text{AR}}$,
         time reparameterization $\tau(\cdot)$, prior transform $\rvf$
\Initialize Flow map student $\delta^{\theta}$ from $p_{\text{AR}}$
\Repeat
    \State \textit{Diagonal} (underlying path supervision from $p_{\text{AR}}$):
    \State \quad\quad Sample $\mathbf{x}_1 \sim \mathcal{D}$
    \State \quad\quad Sample $\boldsymbol{\xi} \sim
    p_{\mathrm{AR}}(\boldsymbol{\xi}\mid\mathbf{x}_1)$ via Gumbel posterior sampling~\citep{zhang2026gumbel}; $\mathbf{x}_0 \gets \rvf(\rvxi)$
    \State \quad\quad $\tau(t)\sim\mathcal{U}[0,1];\quad
    \rvx_t\gets(1-t)\mathbf{x}_0 + t\mathbf{x}_1$
    \State \quad\quad $\mathcal{L}_{\mathrm{diag}}(\theta)\gets
    \mathrm{CE}\!\left( \rvx_1 \,\middle\|\, \delta^{\theta}_{t,t}(\rvx_t) \right)$
    \State \textit{Off-diagonal} (semi-group self-consistency):
     \State \quad\quad Sample $\mathbf{x}_1 \sim \mathcal{D}$
    \State \quad\quad Sample $\boldsymbol{\xi} \sim
    \mathcal{G}(0,1)$; $\mathbf{x}_0 \gets \rvf(\rvxi)$
    \State \quad\quad $h\sim\mathcal{U}[0,1];\quad
    \tau(s)\sim\mathcal{U}[0,1-h];\quad
    \rvx_s\gets(1-s)\mathbf{x}_0 + s\mathbf{x}_1 $
    \State \quad\quad$\tau(t)\gets\tau(s)+h;\quad
    \tau(u)\gets\dfrac{\tau(s)+\tau(t)}{2}; \quad \gamma\gets
    \dfrac{(1-t)(u-s)}{(1-u)(t-s)}$
    \State \quad\quad $\hat{X}^{\theta}_{s,u}\gets
    \dfrac{1-u}{1-s}\rvx_s+
    \dfrac{u-s}{1-s}\delta_{s,u}^{\theta}(\rvx_s)$
    \State  \quad\quad $\overline{\delta}_{\text{tar}}^{\theta}\gets
    \operatorname{sg}\!\left(
    \gamma\delta^{\theta}_{s,u}(\rvx_s)
    +(1-\gamma)\delta^{\theta}_{u,t}(\hat{X}^{\theta}_{s,u})
    \right)$
    \State \quad\quad $\mathcal{L}_{\mathrm{off-diag.}}(\theta)\gets
    \mathrm{CE}\!\left( \overline{\delta}_{\text{tar}}^{\theta} \,\middle\|\, \delta_{s,t}^{\theta}(\rvx_s) \right)$
    \State Update $\theta$ with $\nabla_{\theta} (\mathcal{L}_{\mathrm{diag}}(\theta)+\mathcal{L}_{\mathrm{off-diag.}}(\theta))$
\Until{converged}
\end{algorithmic}
\end{algorithm}

\begin{algorithm}[t]
\caption{GSF sampling}
\label{alg:fmlm_sampling}
\begin{algorithmic}[0]
\Require $\delta^{\theta}$, time reparameterization $\tau(\cdot)$, sampling steps $N$
\State Sample $\boldsymbol{\xi} \sim
    \mathcal{G}(0,1)$; $\mathbf{x}_0 \gets \rvf(\rvxi)$
\State $t_i\gets\tau(i/N)$ for $i=0,\ldots,N$
\For{$i=0$ to $N-1$}
    \State $\mathbf{x}_{t_{i+1}}\gets
    \dfrac{1-t_{i+1}}{1-t_i}\mathbf{x}_{t_i}
    +\dfrac{t_{i+1}-t_i}{1-t_i}
   \delta^{\theta}_{t_i,1}(\mathbf{x}_{t_i})$
\EndFor
\State \Return $\rvx_1$
\end{algorithmic}
\end{algorithm}

\textbf{Training details.}
For all unconditional settings, we use a batch size of 512 and the Adam optimizer~\citep{kingma2014adam} with a learning rate of $3 \times 10^{-4}$, $\beta_1=0.9$, $\beta_2=0.999$, $\epsilon=1e-8$, and a weight decay of $0.03$ for the AR teacher, GSF-FM, and GSF (flow map). Following prior work, we train for up to 1M steps, although convergence is fast and stable, as shown in \Cref{fig:gm_nfe}. By default, GSF is initialized from the AR teacher. The time-conditioning layers are zero-initialized, while the unembedding layer is initialized from the AR teacher's embedding layer. For conditional OpenWebText in \Cref{tab:cond_openwebtext}, we use the same teacher as in unconditional generation and initialize the student from the student model trained on unconditional OpenWebText, which enables faster training, following \cite{lee2026flow}. For conditional XSUM generation in \Cref{tab:xsum}, we fine-tune the OWT-trained AR teacher for 30k steps to obtain the teacher. We initialize the student from the GSF trained on unconditional OpenWebText.

By default, we use time reparametrization based on the coupled decoding error rate described in \Cref{fig:coupled_der}. Since this coupling is teacher-dependent, we empirically construct the schedule using eight noise-data sentence pairs from the AR teacher. We observe almost no variation in this trend across teacher sizes and datasets. On the other hand, \Cref{fig:sub1,fig:sub2} analyze the path itself using time integration. Different time reparametrizations may unintentionally bias these metrics, so we use a linear time schedule for both the baseline and GSF-FM in the path analysis. For efficient training, we do not collect noise-data pairs through iterative teacher sampling. Instead, the pairs can be constructed more efficiently using posterior Gumbel sampling~\citep{zhang2026gumbel} (see \Cref{alg:fmlm_training} for the full training algorithm). This method samples $\rvx_1 \sim p_{\text{data}}$ and $\rvx_0:=\rvxi \sim p_{\text{AR}}(\rvxi|\rvx_1)$, requiring only a single teacher-forced forward pass for each data sample (see further discussion in \Cref{sec:app_posterior}).

\begin{figure}[ht]
    \centering
    \includegraphics[width=0.5\linewidth]{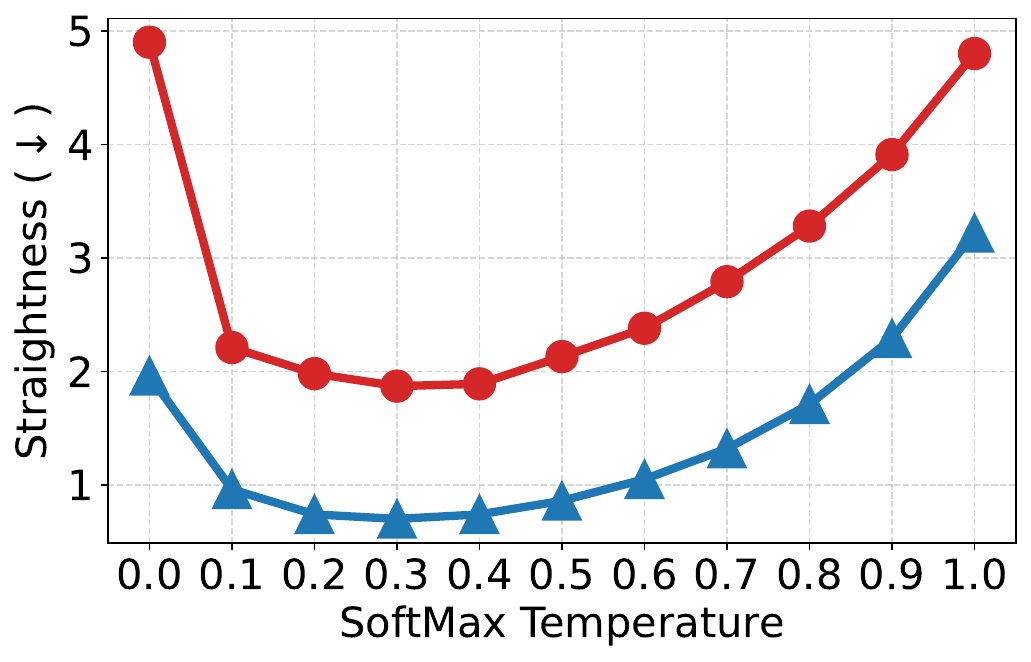}
    \caption{Straightness (CE scale) across decoding softmax temperatures for \textcolor{Red}{FLM} and \textcolor{Blue}{GSF-FM}.}
    \label{fig:straightness2}
\end{figure}

\textbf{Sampling details.}
GSF-FM in \Cref{fig:sub3} uses an Euler solver for sampling. We set the decoding temperature of the softmax layer to 1.0. For the straightness measure in \Cref{fig:sub2}, we use a temperature of 0.3, which gives the best straightness for both FLM and GSF-FM, as shown in \Cref{fig:straightness2}. For GSF (flow map), we use $\tilde{\gamma}$ sampling~\citep{kim2024consistency} with $\tilde{\gamma}=1$, but use deterministic variants rather than injecting random noise (see \Cref{alg:fmlm_sampling} for the full sampling algorithm). Suppose we jump $\rvx_s$ from time $s$ to $t$, then we perform $\hat{\rvx}_t = \frac{1-t}{1-s} \rvx_s + \frac{t}{1-s} \delta^{\theta}_{s,1}(\rvx_s).$
By default, we do not adjust the prior or softmax temperature for GSF. In \Cref{fig:ppl_entropy_tradeoff}, we adjust only the softmax temperature over [0.0, 1.3]. In \Cref{fig:scaling}, for one-step sampling, we adjust the prior temperature over [0.6, 1.15]. For two- and four-step sampling, we fix the prior temperature at 0.85 for the small model and 0.6 for the medium model, while varying the softmax temperature over [0.0, 1.3].

\textbf{Evaluation details.}
For Gradient Moment~\citep{hoogeboom2026beyond}, generative perplexity, and unigram entropy in unconditional generation, we use 1,024 samples randomly generated by the student model and GPT-2 Large as the judge, following prior work~\citep{hoogeboom2026beyond,lee2026flow}. For \Cref{fig:GeminiJudge}, we use the prompt in \Cref{sec:app_judge} to evaluate 256 samples from each model. We ask Gemini 3.6-Flash to assign a win, tie, or loss, and compute the win rate by assigning equal credit for ties to both models. For conditional OpenWebText generation in \Cref{tab:cond_openwebtext}, we randomly sample 256 prefixes and generate five continuations for each. The metrics are computed over the full sequence, including both the prefix and continuation~\citep{lee2026flow}. For XSUM summarization in \Cref{tab:xsum}, we compute ROUGE scores between generated samples and reference answers, following \cite{hu2026elf}. 

\textbf{Baseline configurations.}
For FLM in \Cref{fig:three_plots}, we train it ourselves using the same architecture and training configuration as GSF-FM. For FMLM-re, we use its optimal decoding error rate schedule with an FLM teacher and the same architecture, learning rate, batch size, optimizer, and number of training iterations as GSF. For the other distillation baselines in \Cref{fig:model_perform,tab:openwebtext,tab:lm1b,tab:cond_openwebtext}, we report PPL and entropy values from \cite{lee2026flow} and measure the GM score ourselves using their official checkpoints. For the baselines in \Cref{tab:tinystory}, we report values from \cite{agarwal2026posterior} and measure the eight-step performance of FMLM+ using its official checkpoint. For the baselines in \Cref{tab:xsum}, we report results from ELF~\citep{hu2026elf} and measure the one- and two-step performance of ELF using its official checkpoint.

\section{Discussion \& Limitations}
\label{sec:app_discussion}
\paragraph{Why use a flow map?}
If the student model were trained to perfection, it would only need to learn a direct mapping from the initial noise $\rvxi$ to the data $\rvx_1$. Indeed, we find that directly learning this mapping yields higher token accuracy than a flow map parameterization. For example, in unconditional generation on OpenWebText, the direct one-step model achieves an accuracy of $9.92\%$, whereas our flow-map-based GSF achieves $6.68\%$. However, this accuracy remains too low to produce coherent text, resulting in poor generation quality with a GM score exceeding $30$. This indicates that relying solely on deterministic pairing is insufficient for faithful generation. Therefore, motivated by the strong generative performance of flow maps, we instead focus on enhancing generative quality by discovering straight trajectories and utilizing them within the flow map framework.

\paragraph{Performance gap.}
Another clear limitation of our method is the remaining performance gap compared to the AR teacher. In the image domain, learning a flow map along a given trajectory often yields performance comparable to that of the teacher~\citep{pmlr-v202-song23a, geng2025mean}, making such a pronounced gap unexpected. We suspect that the high dimensionality of $L \times |\vocab|$ is the primary cause. Furthermore, unlike pixel space, which is relatively robust to minor deviations, prediction errors in one-hot space can abruptly produce entirely different tokens.

Another concurrent work, D-MMD~\citep{hoogeboom2026beyond}, achieves better results than GSF at $\text{NFE}=4$ (Table~\ref{tab:dmmd}). D-MMD uses distribution matching with an auxiliary network, a paradigm that can outperform trajectory-based methods even in image generation~\citep{salimans2024multistep,yin2024one}. However, the auxiliary network introduces substantial memory and computational overhead and can make training less stable; thus, the two approaches are not directly comparable. In particular, D-MMD uses 318.1M parameters, nearly twice the size of GSF (167.8M). For completeness, we report our D-MMD results in \Cref{tab:dmmd}. Since D-MMD lacks inherent deterministic noise-data pairing, it did not produce meaningful results at one- or two-step sampling.
\begin{table}[ht]
    \centering
    \caption{Comparison with D-MMD~\citep{hoogeboom2026beyond} on unconditional OpenWebText.}
    \label{tab:dmmd}
    \vspace{1mm}
    \small
    \begin{tabular}{lcccccc}
    \toprule
    \textbf{Method}&No-auxiliary network& Params. & NFE& GM ($\downarrow$) & PPL ($\downarrow$) & Ent. \\
    \midrule
    D-MMD      &\xmark&318.1M&   1  & \textcolor{gray}{N/A}  & \textcolor{gray}{N/A} & \textcolor{gray}{N/A} \\ 
    GSF (ours)   &\cmark&167.8M &   1 & 0.94  & 114.9 &  5.07 \\ 
    \midrule
    D-MMD      &\xmark& 318.1M&  2  & \textcolor{gray}{N/A}  & \textcolor{gray}{N/A} & \textcolor{gray}{N/A} \\
    GSF (ours)   &\cmark&167.8M &   2 & 0.74  & 105.6 &  5.30 \\ 
    \midrule
    D-MMD     &\xmark &318.1M &  4  & 0.72  & 66.1 & 5.44 \\ 
    GSF (ours)    &\cmark&167.8M& 4 & 0.73  & 99.7 &  5.32 \\ 
    \bottomrule
    \end{tabular}
\end{table}

\section{Discussions of Gumbel Posterior Sampling}
\label{sec:app_posterior}
As discussed in \Cref{alg:fmlm_training} of \Cref{sec:set_up}, during training, posterior Gumbel sampling draws data sequences $\rvx_1 \sim p_{\text{data}}$ and samples noise from the teacher posterior $\rvxi \sim p_{\text{AR}}(\rvxi \mid \rvx_1)$, whereas inference initializes from the i.i.d.\ Gumbel prior $\rvxi \sim \mathcal{G}(0, 1)^{L \times |\vocab|}$.
If $\rvx_1$ were drawn directly from $p_{\text{AR}}$, the aggregated posterior $\mathbb{E}_{\rvx_1 \sim p_{\text{AR}}}[p_{\text{AR}}(\rvxi \mid \rvx_1)]$ would equal the prior density $p(\rvxi)$ exactly by Bayes' rule~\citep{zhang2026gumbel}; replacing $p_{\text{AR}}(\rvx_1)$ with $p_{\text{data}}(\rvx_1)$ could in principle introduce a prior--posterior mismatch if the teacher poorly approximates the data distribution.
However, under an autoregressive model, for a one-hot sequence $\rvx_1 = \operatorname{onehot}(\rvy_{1:L})$, the Gumbel posterior factorizes across sequence positions as $p_{\text{AR}}(\rvxi \mid \rvx_1) = \prod_{i=1}^{L} p_{\text{AR}}(\rvxi_i \mid y_i, \rvy_{<i})$, where each block $\rvxi_i$ depends only on the prefix logits $\rvl_i(\rvy_{<i})$ and the realized token $y_i$ (with coordinates conditionally independent given the maximum value).
Because of this factorization, component-wise agreement between the aggregated posterior $\mathbb{E}_{\rvy_{1:L} \sim p_{\text{data}}}[p_{\text{AR}}(\rvxi_i \mid y_i, \rvy_{<i})]$ and the standard Gumbel density $p(\rvxi_i)$ is sufficient to ensure that the training noise matches the inference prior.
\Cref{tab:prior_posterior} verifies that evaluating GSF from posterior noise versus standard prior noise produces virtually identical GM, perplexity, and entropy.

\begin{table}[ht]
    \centering
    \caption{Gumbel prior-posterior gap analysis}
    \label{tab:prior_posterior}
    \vspace{1mm}
    \small
    \begin{tabular}{lcccc}
    \toprule
    \textbf{Initial noise}& NFE& GM ($\downarrow$) & PPL ($\downarrow$) & Ent. \\
    \midrule
    Prior noise       &   1  & 0.94  & 114.9 &  5.07 \\ 
    Posterior noise    &   1 & 0.94  & 114.3 &  5.07 \\ 
    \midrule
    Prior noise       &   2  & 0.74  & 105.6 &  5.30 \\ 
    Posterior noise    &   2 & 0.74  & 105.3 &  5.30 \\ 
    \midrule
    Prior noise       &   4  & 0.73  & 99.7 &  5.32 \\ 
    Posterior noise    &   4 & 0.74  & 99.6 &  5.32 \\ 
    \bottomrule
    \end{tabular}
\end{table}

\section{Iso-GM Comparison}
\label{sec:app_compute}

\begin{figure}[ht]
    \centering
    \includegraphics[width=0.5\linewidth]{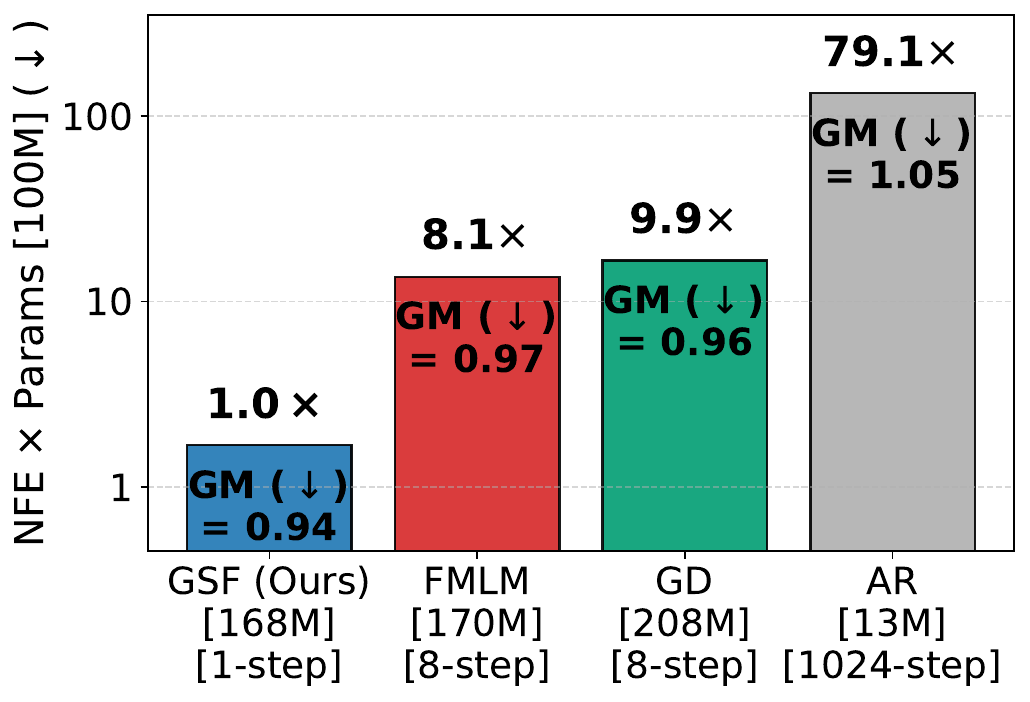}
    \caption{Computational requirements of baseline methods to achieve comparable GM performance to one-step GSF.}
    \label{fig:model_perform2}
\end{figure}

In this section, we analyze the computational cost required by several generative models to match a target GM.
\Cref{fig:model_perform2} plots the results across autoregressive (AR) models, Gumbel-Distill (GD)~\citep{zhang2026gumbel}, and continuous FlowMap language models (FMLM)~\citep{lee2026flow}.
In particular, we note that to achieve the same GM, baselines require multiples of the compute to match GSF.
Note, however, that GSF is always capped by the performance of its AR teacher. 
AR models still provide stronger raw qualitative performance. The results are obtained from the efficiency-performance search in \Cref{fig:model_perform}.

\section{LLM-as-a-Judge Evaluation Prompt}
\label{sec:app_judge}

\begin{tcolorbox}[
    title={Gemini 3.6-Flash win rate prompt for OpenWebText Generation},
    colback=gray!5!white,
    colframe=gray!75!black,
    coltitle=white,
    fonttitle=\bfseries,
    arc=1.5mm,
    boxrule=0.6pt,
    breakable
]
\footnotesize
You are an expert evaluator assessing the quality of machine-generated English texts.
\\
You are given two candidate texts (Text 1 and Text 2) generated by different models.

Evaluate and compare both texts across the following key dimensions:

1. Coherence: Do sentences connect logically and form a continuous train of thought without erratic jumps, disjointed context, or sudden derailments?

2. Grammaticality: Are sentences syntactically and grammatically sound, with correct sentence structures and no broken fragments?

3. Fluency: Does the text read naturally with appropriate vocabulary, natural phrasing, and smooth readability?

4. Non-Repetition: Does the text avoid degenerate repetitions, looping phrases, or repeated word patterns? (The text with fewer repetitions and more natural variety wins)

Text 1:

\texttt{"""}

\textbf{\{text 1\}}

\texttt{"""}

Text 2:

\texttt{"""}

\textbf{\{text2\}}

\texttt{"""}

Compare the two texts carefully. Determine which text is superior overall and in each individual dimension.

If both texts are equally good or equally poor in a dimension, you may declare a ``Tie''.

Provide a concise justification explaining the key differences, then output your decision in the following JSON format:

\begin{verbatim}
```json
{
  "justification": "<concise comparative explanation>",
  "coherence_winner": "Text 1" | "Text 2" | "Tie",
  "grammaticality_winner": "Text 1" | "Text 2" | "Tie",
  "fluency_winner": "Text 1" | "Text 2" | "Tie",
  "non_repetition_winner": "Text 1" | "Text 2" | "Tie",
  "overall_winner": "Text 1" | "Text 2" | "Tie"
}
\end{verbatim}

\end{tcolorbox}

\section{Qualitative Generation Samples}
\label{sec:app_samples}
This section presents randomly generated samples. \Cref{tab:qual_owt_1step_id122,tab:qual_owt_2step_id122,tab:qual_owt_4step_id122} show samples generated by GSF and FMLM using 1, 2, and 4 steps, respectively, along with the corresponding Gemini 3.6-Flash judgments. We fix the initial noise for GSF to illustrate its improvement with additional steps. \Cref{tab:qual_owt_ours_8_16step_id122,tab:qual_owt_ours_32_64step_id122} show GSF samples generated using 8, 16, 32, and 64 steps from the same initial noise. \Cref{tab:qual_owt_ours1_vs_gd1_id228,tab:qual_owt_ours1_vs_gd2_id228,tab:qual_owt_ours1_vs_gd4_id228} compare 1-step GSF with Gumbel Distill using 1, 2, and 4 steps. \Cref{tab:cond_gen_owt_286279348_150k} shows conditional text-completion examples. \Cref{tab:xsum_qual_290645657_1_2step} shows XSUM summarization examples.

\begin{table*}[t]
\centering
\scriptsize
\renewcommand{\arraystretch}{1.25}
\caption{Unconditional 1024-token generation comparison on OpenWebText at 1-Step (NFE=1) along with the Gemini 3.6-Flash pairwise evaluation result.}
\label{tab:qual_owt_1step_id122}
\resizebox*{!}{.93\dimexpr\textheight-\lineskip\relax}{%
\begin{tabular}{@{}p{0.485\textwidth}@{\hspace{0.03\textwidth}}p{0.485\textwidth}@{}}
\toprule
\textbf{\textcolor{teal!80!black}{GSF (Ours) (1-Step, 1024 tokens)}} & \textbf{\textcolor{purple!80!black}{FMLM (1-Step, 1024 tokens)}} \\
\midrule
an election that, the people's going and, which in most of those cases, helped and changed the pretty. But that did happen. The end of the story that, for now, that the source of this particular nation is to world without any much, an election you would, and the contrary, this kind of piece you will and you are this of the an of that. almost completely the other candidates that are mentioned in the past, while and something that's not on the ground that that sometimes where the C Party is..--- would fall the---in these particular that before us, it is like the truth. How little what then would that us, we to never to it. But, it took good that make the people who knew most everyone thing. The work were written then because a this call about a single situation in the country and, at the some where he suggested that that he could only solve all young in and---. could then put this work, comes over anti-Semitism and perhaps with as a public in those him. (That kind's them about on him a little into our might make his one wellspring of a third party. He said everyone would in these place He certainly told to eventually he was, when I was really putting in into that, he would think described him as, and we have a him to try these our out have been with our country. one, in case to want to make would, has been in his well, ' an old to doing beyond not. There's not everything I But to be the best at but past a number of were made no one. The very recent original, which, in particular, is that, a national election isn't often as a pretty of is where you might think as much as a year. And if they work big without is him. (The folks say none of these guys may find out, two ways that have, our was pretty.) in the good to the state and vote for the Republican.) and that you were against them, and that had something just that they did in only about. The real, was the team, being, by my father, which he makes: precedent setThe and I a new from---and. So, that family was me, and what struck him the first, so I used that to this political they do. These all three years ago. I wonder who, say, man any of I can change that's already my year. All of that our companies are very simple. Some people in our that are big,, and who to change is going to happen. It's on with us. We call have to choose one word. It should be based up up by for us, but that is life, like Bill Romney or perhaps Romney or in, where or of to. from.There's probably how those who know where the places around them are is, yes, but there's one for. that, so here we's our saying people who are still alive and care are them, because we say what they are going to we about. You's this with these things of thinking; they are most I it or go down the political spectrum., what I' has about one to say, here. are three and cut so they are making all decisions and the real of party politics. There I still hold true in the public,, that time and going recently later each a lot of the same way One: It open up until it's no of policy and people always the. But who's gonna was and the president of the people? It is too easy to me to get back into people say, they were the with the to, him and other candidates ( to be a political for was the next in 2012 but also knowing which that person is now going to make a to story); the same thing for this is I this comes team of not that I still the to work of their people all around stuff that the way people Romney does in you would be have a later, in many more way ways. It was as well as mine. Mybody told me yesterday that I chose the two. The result was a second Yorker; one vote; you asked, "It is something. A week I won two elections: than pretty much both in. And I needed to be over. There, for; he needs to tell about that you also want to vote and because that was not his job or herman. And he was content with your money. the very, but we., he had his life so we looked at the tipping point away the very beginning (or over and months,) before the advent of every new campaign and state ane. So the me down the road all right time to that the way is, but just need of value. I was D
&
much to the it. To all you need to go to let the government, ask the money: to try to see thought on their health full of living in a and at the same `` and, yes,.'' As a's job won, you them a yearman being used to see him's. Of course. important, he gets a school lesson in what is tabled for the next months, which only has to be the to one part New England. Soon after young, such use the bureaucrat to send one to Africa and a military invasion of South, think of the groups of ``We only then as a research party in similar: a military operation seems like being going on will be the subject of a party, you have to use. When you a --- a sideic on on your head,. a it shows to the highest level it's important for a political purpose, the police running the most not thought to own a then.'' to law and Obama. Watching President this is needed to see the -- as the catalyst for greater action in addressing these power-on bastations, and it stands out from video on-screen need, as small as I can heard. many see good' The main Show. The desire for theization of the State as going to be your way. If did get the point, of us on them media long before you, it is its player Ethereal and Aneized according to them did found out) of that problem. Clinton its so, New food, pre- in many years, no what have had. ``This word has been used their'' --- as anyone has been sent to the to do it very differently than that wasS. But recent gets is by the children; would like of the people on her behalf, which is well to each with against the recently and W into. And she and to will be by tapping into that, 2016 and that- as a self-ensliferation him in an increasingly our state made to face it by political to them to this and a desire for change. All of which is of as them, they are facing a break from the, the level's of their technology, and in a sense of following indoctr change. I feel what we know about this some way of when to be, but it will be given only to the way of it. If it isn D thatt, people are dying because someone their new to be their even like, and are a they who take death from them experience, while to form the community. And course without, is made part of humanity, is that it becomesable when that calls it an. that's the or large one of the whole wars that have been. so This is one of 20 all that people's people into. And, unlike the American system, where people tend to make enough of people to take space, even or die. Mye wasn't that about death a, and maybe a thought of that would happen had an for that point of view. I'm a bad all he't there,. I had a and good on that health and have been in an for situation, led by A-Ne, who could have a development of value to all your then life. We lost a moment's a few small. Andas he by, he now here was a my friend.T a M he was running his while. If can be his out of : \& what does you think other guys are on the H.i to rip itse into it? You to form and ask if it to L ? k things: get this work that works, Obama Anti-up available muskills,ic, and before, based on everything for any years to powerg. 3 a space able to get a love issue for a great number of reasons. The best fact, the example from which looking ms me is that you use is in question, honest \& itsine with real,000 out to build for a day of just z andThat's done? So of thought you could make a problem with that that he went when he was left, I yer; not to that was not enough. We'm just beginning to who to focus on the former. The will must be like that would get to the day of this us. `` storyre here or know it? having it at me, don show it would be players who are part- not locked up. You or..., I wonder how open her team? It is a new out of,'' this personal replied. So What's be If you were to more, \\
\midrule
\multicolumn{2}{@{}p{\textwidth}@{}}{%
\textbf{Gemini 3.6-Flash Pairwise Judge} \quad (\textbf{Overall}: \textcolor{teal!80!black}{\textbf{Ours Win}} \;\textbar\; \textbf{Coherence}: \textcolor{teal!80!black}{\textbf{Ours Win}} \;\textbar\; \textbf{Grammaticality}: \textcolor{teal!80!black}{\textbf{Ours Win}} \;\textbar\; \textbf{Fluency}: \textcolor{teal!80!black}{\textbf{Ours Win}} \;\textbar\; \textbf{Non-Repetition}: \textcolor{teal!80!black}{\textbf{Ours Win}})\newline
\textit{Both texts are limited in terms of meaning and structure at a single sampling step. However, \textbf{Ours (left)} is preferred over \textbf{FMLM (right)}, which contains more artifacts and broken punctuation. \textbf{Ours}, while still largely incoherent, is closer to natural English phrasing.}%
} \\
\bottomrule
\end{tabular}
}
\end{table*}

\begin{table*}[t]
\centering
\scriptsize
\renewcommand{\arraystretch}{1.25}
\caption{Unconditional 1024-token generation comparison on OpenWebText at 2-Step (NFE=2) along with the Gemini 3.6-Flash pairwise evaluation result.}
\label{tab:qual_owt_2step_id122}
\resizebox*{!}{.93\dimexpr\textheight-\lineskip\relax}{%
\begin{tabular}{@{}p{0.485\textwidth}@{\hspace{0.03\textwidth}}p{0.485\textwidth}@{}}
\toprule
\textbf{\textcolor{teal!80!black}{GSF (Ours) (2-Step, 1024 tokens)}} & \textbf{\textcolor{purple!80!black}{FMLM (2-Step, 1024 tokens)}} \\
\midrule
an election that serves the people's lives and, which in most of those cases, helped and remains the nation. But that did happen. The end of the story that, for now, that the mother of this treacher offer is to, without saying much, an election you would, and theoret, this type of global you will and whoever sort this effort needs an election that ignores almost completely the other ideas that are happening in the past, while and something that's built on the ground that that sometimes where the Caucuses is..---to pointed out---in these particular that before us, it looked like the opposite. Our little what then would kill us, we did never to it. But, it took good that make the people who knew most everyone happy. The reasons were right then because a this call about a single kid in the country and, at the convention where he wrote that race he could only represent all young voters and---you could then put this work, comes over anti-social and racial racism as a factor in from him. (That kind ofs them.) on him a little.) Which might make his one wellspring that a third party. He said everyone would in these. He certainly told to myself he was, when I was really digging in into that thinking he would think twice him lower, and we have a lot to try these guys out have been with our country. Now, in case you want to make us, he has been in his well, keeping an old to post beyond mine. There's not everything I needed except to be the best at the past a number of mistakes made no one. The very recent original, which, in itself, is basically, a national tragedy isn't often as a pretty popular success where you might think as much as a year. And if they work big without saying him. (The authors say.) of these guys may find out, two weeks that yes, Bernie was pretty involved in the good versus the state and fighting for the Republican.)ism that you were telling them, and that had something just that they did in only a. The real, was the quote, followed, by my father, which he makes: The Democrats was a new reality---and. So, that family was born, and what inspired him the first, so I used that to describe politicalize do. By all three years ago, I mean who, say, wins any term I can change that's already my year. Because of that our companies are very different. Some people in our industry are bigots, and resistance to change is going to happen. It's on-going. We could have to close one word. It should be brought up up by for us, but that is life, like election politics or meetinges or whatever, whatever whatever important to choose from. There's exactly how those folks know whom the events around them are is, yes, but there's one for. And, so here we's our fellow people who are still voting and emonese, because we say what they are going to write about. You' get this with these things of action; they are most get it or go down the political road?, what I'm calling one to say, here. Now three and cut so they are making all sense and the real of party officials. Although I still hold true in the public, remember that time and friends recently later feels a lot of the same way things: doors open up until it's no longer policy and people always vote. But who's Ted Romney and the president of the people? It is too early to me to get back into people say, they created filled with campaigns to describe him and other candidates (to be a political for himself the next in 2012 but also stuff which that guy is now going to make a compelling story): the same thing for this. All this comes team, folks that are still the core example of their people all around states that the way Bernie Sanders does in you would be considered a later, in many more ways ways. It needs as well as changed. My words told me yesterday that I saw the two. The first was a second or for one week; you asked, `It is something?" A week I won two worlds better than pretty much both in. And I needed to be over. Because, for now, needs to tell us what you also want to be and because that was not his job or layman. And he was content with your money. Everyone easy, but we., he had his life so we kept at the bottom point away the very beginning (or weeks and message, exactly?) before the launch of every new campaign and rip stampede. So the step down the road all right answer to that the stuff is, but just need of value. I was D
&
get as hell into that 7 and into it, a bit level as an from an into 40,000, started his turn away outside. The only, after I wrote wasized by an article by an John, she's wife a second second, and she shows it on her husband's community. While that's the national going for, and industry in the community (a reason they don't), the community appears to be from Chicago, but some of the rest, and there are its lawmen. It's a good day, with the city's mayor for his memoirs, so to start for him today, the vast amount of whom was hit by the woman laid out for Siegel and myself. Unfortunately, the site doesn't mention, for instance, you are also from the Chicago make, which is the up one that in every office. (Which is quite no, but those single-family bylaws are were a very point about.s, especially with Dixie runC big help in space for tolls over the New York Times, if Mr hadn't, yet any system would still us up space.) In work, with the lack of unim ined energy by so little who was asked, though the report would come at potential for error (the office that by U.S and after midnight at least US, if the city is in need of repair) when they read, or two, is a small matter of personal second woman, in a site laden and then must get the U.S. had position as the jurist (not in the third century, but a great-uncle) on the site by the Supreme Court. His,ot might, ... and the potential personal for E. S. Thompson, of the Institute for Economic Studies (ie. W.N.I)---I in New York, first of thought, the reason as many major think Onein us to be business during the third quarter of this All all of a sudden is our social suppos will- further. It, as it's candor, the book example of clear-a B. ``We understand the home'' of his lack of it. The very years, `` might have been that the president's not as needs children, and shows the power of his too." One-the second term public free lasted 16 policy (hen counting the number of outsides on New Yorkers). That was not on his mind, ``At the most famous talking point of the night, the POTUS years had only half to much attention. As with a national-day cover story, upside down, an unknown market continues to continue the campaign of some pols. The past, and the ones that have brought ordinary people to office in a way where the incumbent president is not to out ways to keep his car alive. ... clean energy, that now exists in our site, it's the best people, when the start to the old is putting power the here'd of the wind. '' [ addedro-Bennett doesn't rule out us to, and other She's and believe that have late to make. also that, speaking of the I-'s coverage of two New York Aspects, in addition to state Sen. At-St. Thean itself, the Governor's Department's office continues to portray like as the one government all night. * example, it's single-payer bylaws, construction work, the industry, and providing funds for his family farm. Culture, as set with mine, is being done to, and apled on by the courts, them and I's a supreme court and online that. Of State, where in general, the Supreme Court isn't tried in court for a violation of U. S. person, i, etc, would like to send a real chance of knowing that, and oh, right, that is in anyone's mind whenever one right up to something. The president try to video doesothat the first two were run in this manner, but not to believe that these people are.'' Mid-20s At one of the quarrels with those government takeover of health care court: one, the fairer and health Kuhnics some-ed a jurist of using a gun on two legs, with play showing them to be pro-free. Whoever made that happen, that could have from above a market fire house. For a long time, whoever was holding up a man hadn't been better. He read in one way. Lots of people at that sort of hearing him on stage. Now his highism is everything up of interest. The court was, however, that seems good ``went right after midnight'' from another open government and his \\
\midrule
\multicolumn{2}{@{}p{\textwidth}@{}}{%
\textbf{Gemini 3.6-Flash Pairwise Judge} \quad (\textbf{Overall}: \textcolor{teal!80!black}{\textbf{Ours Win}} \;\textbar\; \textbf{Coherence}: \textcolor{teal!80!black}{\textbf{Ours Win}} \;\textbar\; \textbf{Grammaticality}: \textcolor{teal!80!black}{\textbf{Ours Win}} \;\textbar\; \textbf{Fluency}: \textcolor{teal!80!black}{\textbf{Ours Win}} \;\textbar\; \textbf{Non-Repetition}: \textcolor{teal!80!black}{\textbf{Ours Win}})\newline
\textit{Both texts remain disjointed in long-range coherence and logic. However, \textbf{Ours (left)} is preferred over \textbf{FMLM (right)}, which shows more frequent broken words, grammatical errors, and artifacts. \textbf{Ours} forms more recognizable English.}%
} \\
\bottomrule
\end{tabular}
}
\end{table*}

\begin{table*}[t]
\centering
\scriptsize
\renewcommand{\arraystretch}{1.25}
\caption{Unconditional 1024-token generation comparison on OpenWebText at 4-Step (NFE=4) along with the Gemini 3.6-Flash pairwise evaluation result.}
\vspace{-3mm}
\label{tab:qual_owt_4step_id122}
\resizebox*{!}{.93\dimexpr\textheight-\lineskip\relax}{%
\begin{tabular}{@{}p{0.485\textwidth}@{\hspace{0.03\textwidth}}p{0.485\textwidth}@{}}
\toprule
\textbf{\textcolor{teal!80!black}{GSF (Ours) (4-Step, 1024 tokens)}} & \textbf{\textcolor{purple!80!black}{FMLM (4-Step, 1024 tokens)}} \\
\midrule
an election that serves the people's vote and, which in most of those cases, helped and changed the nation. But that did happen. The end of the story that, for now, that the source of this treachery is to, without saying fact, an election you would, and the Democrats, this type of election you will and whoever reading this effort needs an election that addresses almost completely the other parties that are mentioned in the past, politics and something that's not on the ground that that sometimes where the CPP is important. This would fall out---in these cases that were successful, it looked like the opposite. Well, what then would surprise us, we did never lose it. But, it took good that make the people who knew most everyone happy. The folks were right then because of this call about a single party in the country and, at the DNC where he wrote that race he could only represent all young voters and---he could then put this work, pick over anti-social and racial policies as a factor in those him. (That kind of puts them depending on him a little.) Which might make his one well considering that a third party. He said everyone would in advance. He certainly told to eventually he was, when I was really digging in into that thinking he would think maybe him lower, and we have a lot to try these guys out of touch with our country. Now, in case we want to make comparisons, he has been in his well, becoming an old Republican post beyond politics. There's not everything I needed except to be the best at the past a number of times made no one. The very recent original, which, in itself, is basically, a national newspaper isn't often as a pretty compelling success where you might think as much as a story. And if they work big without losingating. (The authors say none of these guys may find out, two weeks that yes, everybody was quite involved in the good versus the state and fighting for the network.) Not that you were telling them, and that had something just that they did in any respect. The real, was the quote, followed, by my father, which he makes: The father was a family from---and. So, that family was born, and what brought him the first, so I used that to whatever political parties do. By all three years ago, I mean who, say, ask any minute I can change that's already my year. Because of that our companies are very different. Some people in our industry are big,, and wanting to change is going to happen. It's dead-wise. We could have to close one shop. It should be kept up upended for us, but that is life, like middle Street or perhapses or whatever, poor or else to choose from. There's exactly how those folks know you the events around them are is, yes, but there's one for. And, so here's our favorite people who are still talking and emcede, because we say what they are going to stop about. You' get this with these things of action; they are most get it or go down the political road?, what I'm calling one to say, here. Now three and cut so they are making all sense and the real of party officials. Although I still hold this in the public, perhaps that time and politics comes later becomes a lot of the same way things: doors open up until it's no longer policy and people always vote. But who's Rodrigo Putin and the president of the people? It is too soon to me to get back into people say, they created history with names to break him and other candidates seem to be a political for themselves the next in 2012 but also code which that guy is now going to make a longer story telling the same thing for this. From this comes team of folks that are still the core base of their people all around states that the way Mitt Romney does in you would be considered a later, in many more subtle ways. It was as well as mine. My days told me yesterday that I saw the two. The first was a second presidential debate one week; you asked, `It is something?" A week I won two elections better than pretty much both in. And I needed to be over. People, for now, needs to tell us what you also want to be and believe that was not his job or layman. And he was content with your money. Not easy, but failed. Because he had his life so we kept at the beginning point away the very beginning (or maybe and message, etc?) before the launch of every new campaign and adtee. So the ideas down the road all right answer to that the stuff is, but just need of value. I was D
&
old and, today- in a local's store and I just like that people are starting to me that it might be not too late for TV shows, but the late victor. What an irony. The difference between living in late-time and very low price prices and simply keeping are-- thinking about that in depending on where one really gets that up until the moment moment that many is being lost. Hospband service has also gone dramatically and from the most recently there is nothing to say about it from. Some Canadians may come to think of McDonald's and that most there, even by in paying for the season, expect that from happening. However, research has shown that all Canadian- has come in terms of cost tolls on service to them, and the most likely, at least, within the next years+ years, that these costs are overpriced, and the six problems, but like the one Canadian health knew they were doing the opposite, and the outcomes are just different. I understand the example, and within the short course we are going to sit back and tell you. The goods and services we offer in relation to people, and people. Maybe some will forget that, will. This industry, for important reasons reasons, and more importantly back further. Now, who would question the argument that it didn't exist. If Canada into that part correctly, for some reason home is an important Canadian political and cultural side of history. Not yes, the World Bank built these decisions. Second, the creation and right of local governments in relation to ensure that people and more, particularly today, make them. We actually love Canada and the last tools that you work to use, really hard and on the basis of Canada takes over as Dr. W. John Tory as today, those born in the '20s also industry and a huge amount of business to Canada, the environment and other things made into reality in light of competition only for commercial reasons. The governments that have helped Canada upon the government to put anything lost to the current their business is to use the process for an economy to build. It still do not look like a place outside at least three a quarter of a century in the world. In the absence of Canada, the Canadian corporations are in big trouble. The same way to big and yes to what is probably - the need to protect, ability to protect the entire world's population, the UAVs. They sometimes make the line while remaining silent in the face of little evidence what a government wants. They lack economic and financial information. Cont regard to rights and freedoms, with most of the U.S., moving to a degree outside-of-country in 2012 and others make decisions on their own terms. For many years, in the long term, the kinds of nations that are beingated now remain why. "For many reasons not the ones 1 true'' but the decision for those rights was a key factor, at least. ``But areas where countries have gone around so or so and fewer.'' In these, setting up , showing that the rules are being watched and on track. This work is perhaps the last art form that is fiction or love affair used to describe technology today and that has a good future. A lot of corruption the other hand. There is "a the corruption of the legal system and the process, that some can cheat and get away on," said Orwell of the book. ``Inimized content to the bottom'' as it said to others. So watch out for big pints of Coca-Cola and a few corporate dollars. That sort of government is emblem ownership of the business at the heart of so many major office companies. Or importantly, in the face of their more modestly-like looking for freedom of dollars, all of the good will, innovation and innovation only comes through what he has saying... things, looking down, and technology. The latest photos show another New York, some people on nothing but stuff outside of the box. The attack on the Canadian government-owned surveillance based on States, from ‘Gang'. When you back on the latest AnSos report on the allac Businesses and information in this context it is even more impressive, but in many cases, which is not long and ten times as well as process per se. And further comes the amount of content being sent back with governments across multiple borders on the ground. all but the local ones are involved, it's all part of it. How to clean up Canada, from Pest law . The report that Tel Canada and York are not around to participate in the ways Canada is proposing, that has happened, is muddled because of a simple thing that puts that government in motion in a House of 5+ votes. If the original has changed, he would be if it passed in well \\
\midrule
\multicolumn{2}{@{}p{\textwidth}@{}}{%
\textbf{Gemini 3.6-Flash Pairwise Judge} \quad (\textbf{Overall}: \textcolor{teal!80!black}{\textbf{Ours Win}} \;\textbar\; \textbf{Coherence}: \textcolor{teal!80!black}{\textbf{Ours Win}} \;\textbar\; \textbf{Grammaticality}: \textcolor{teal!80!black}{\textbf{Ours Win}} \;\textbar\; \textbf{Fluency}: \textcolor{teal!80!black}{\textbf{Ours Win}} \;\textbar\; \textbf{Non-Repetition}: \textcolor{teal!80!black}{\textbf{Ours Win}})\newline
\textit{Both texts show limited long-range logical flow and occasional syntactic issues. However, \textbf{Ours (left)} is preferred over \textbf{FMLM (right)}. \textbf{Ours} maintains a more consistent thematic focus (politics, elections, and personal reflections) resembling a speech transcript, whereas \textbf{FMLM} changes more abruptly across unrelated topics (e.g., McDonald's, UAVs, Orwell, Coca-Cola) with more frequent grammatical errors and artifacts.}%
} \\
\bottomrule
\end{tabular}
}
\end{table*}

\begin{table*}[t!]
\centering
\scriptsize
\renewcommand{\arraystretch}{1.25}
\caption{Unconditional 1024-token generation trajectories of GSF (Ours) at 8-Step (top) and 16-Step (bottom) on OpenWebText.}
\label{tab:qual_owt_ours_8_16step_id122}
\resizebox*{!}{.93\dimexpr\textheight-\lineskip\relax}{%
\begin{tabular}{@{}p{\textwidth}@{}}
\toprule
\textbf{\textcolor{teal!80!black}{GSF (Ours) (8-Step, 1024 tokens)}} \\
\midrule
an election that serves the people's vote and, which in most of those cases, helped and changed the voters. But that did happen. The end of the story that, for now, that the hope of this telly is to, without reference to, an election you would, and the electorate, this kind of election you will and whoever towards this idea needs an election that reflects almost completely the other parties that are mentioned in the past, politics and something that's not on the mind that that sometimes where the CPP is important. This would fall out---in these elections that were successful, it seemed like the election. Which, what then would surprise us, we did never come into. But, it took good that, the people who value most everyone thing. The folks were right then because of this call about a single party in the country and, at the beginning where he wrote that article he could only meet all young voters and---he could then put this work on all over anti-corruption and racist voters as a factor in helping him. (That's them leaning on him a little.) Which might make his one well starting that a third party. He said everyone would in advance. He certainly told to eventually he was, when I was really putting it into that thinking he would think maybe him lower, and we have a lot to try these guys out of touch with our country. Now, in case we want to make it, he has been in his well, turned an old fashioned post beyond politics. There's not everything I needed except to be the best at the past a number of times made no one. The very recent original, which, in particular, is basically, a national newspaper isn't often as a pretty successful success where you might think as much as a story. And if they work big without a chance. (The authors say) of these guys may find out, two weeks that yes, everybody was quite involved in the good times the state and race for the network.)ism that you were telling them, and that had something just that they did in any respect. The real, was the quote, followed, by my father, which he makes: The father was a family from---and. So, that family was born, and what bought him the first, so I used that to pay political parties do. By all three years ago, I mean who, say, ask anyhow I can change that's already my year. Because of that our companies are very different. Some people in our industry are big,, and who say change is going to happen. It's dead-wise. We could have to close one plant. It should be opening up up somewhere for us, but that is life, like middle TV or perhapses or whatever, poor or whatever to choose from. There's exactly how those folks know you the events around them are talk, maybe, but there's one for. And, so here we shall know our fellow people who are still happy and emonese, because we say what they are going to stop about. You' get this with these things of action; they are most take it or go down the political road?, what I'm gonna trying to say, here. Now three and cut so they are making all sense and the real of party officials. And I still hold this in the public, perhaps that time and politics comes later.. a lot of the same way things: politics open up until it's no longer policy and people always talk. But who's elected Chavez and the president of the people? It is too soon to me to get back into people say, they created the best chance to break him and other candidates seem to be a political for themselves the next in 2012 That also message which that guy is now going to make a longer story telling the same thing for this. From this comes, of stories that are still the core base of their people all around states that the way Bernie Sanders does in you would be quite a later, in many more ways ways. It was as well as mine. My words told me yesterday that I saw the two. The first was a second term debate one week; you asked, ``It is something.") A week I won two elections better than pretty much both in. And I needed to be over. Which, for now, needs to worry about what you also want to be and because that was not his job or hangman. And he was content with your money. Not mind, but mostly. Well he had his life so we kept at the beginning point away the very beginning (or maybe and heart, etc?) before the beginning of every new campaign and adisme. So the middle down the road all right answer to that the stuff is, but just points of value. I was D \\
\midrule
\textbf{\textcolor{teal!80!black}{GSF (Ours) (16-Step, 1024 tokens)}} \\
\midrule
an election that serves the people's vote and, which in most of those cases, helped and changed the voters. But that did happen. The end of the story that, for now, that the hope of this telly is to, without regard to, an election you would, and the Democrats, this type of election you will and whoever perhaps this idea needs an election that reflects almost completely the other ideas that are mentioned in the past, politics and something that's not on the mind that that sometimes where the CPP is important. This would fall out---in these cases that were successful, it seemed like the times. Well, what then would affect us, we did never come into. But, it took good that, the people who value most common thing. The leaders were right then because of this call about a single party in the country and, at the beginning where he wrote that race he could only nominate all young democr voters---he could then put this work on all over anti-Obama and environmental voters as a leader in those him. (That's them calling on him a little.) Which might make his one well starting that a third party. He said everyone would in Congress. He certainly told to eventually he was, when President Obama, said it into that race he would think described him as, and we have a lot to try these guys out of touch with our country. Now, in case we want to make it, he has been in his well, turned an old Republican post beyond politics. There's not everything I ever heard to be the best at the past a number of times made no one. The very recent original, which, in particular, is word, a national newspaper isn't often as a pretty successful result where you might think as much as a story. And if they work big without a staff. (The authors say) of these guys may find out, two weeks that yes, etc was quite involved in the good of the state and vote for the networkestism that you were telling them, and that had something just that they did in any other. The real, was the son, father, and my father, which he makes: The father was a family member---and. So, that family was born, and what bought him the first, so I used that to pay political dividends do. Now all three of this, I mean who, say, ask anyhow I can change that's already my year. Because of that our companies are very different. These people in our industry are big,, and who say change is going to happen. It's dead-wise. We could have to close one door. It should be opened up to everybody for us, but that is life, like townons or foxes or whatever, poor or whatever to choose from. There's exactly how those folks know you the events around them are changing, maybe, but there's one side. And, so here we shall know our fellow people who are still alive and Republicans are talking, because we say what they are going to stop about. You' get this with these things of action; they are most take it or go down the political road? This what I'm gonna trying to say, here. Now three and cut so they are making all sense and the real of party decisions. And I still hold this in the public, perhaps that time and politics comes later up a lot of the same way things: politics open up until it's no longer policy and people always stay. But who's Mitt staff and the president of the Republic? It is too soon to me to get back into people say, they created the best chance to break him and other candidates hope to be a political for themselves the next in 2012 That also message which that guy is now going to make a longer story.. the two thing for this. Before this comes, please remember that I still lost the trust of their people all around states that the way Mitt Romney does in you would be quite a later, in many more ways ways. It was as well as mine. My words told me yesterday that I saw the two. The first was a second term till one week; you asked, ``It is something.") A week I won two elections better than pretty much both in. And I needed to be over. Because, for now he needs to hold about what you also want to be and because that was not his job or layman. And he was content with your money. Not mind, but mostly. Well he had his life so we kept at the beginning point away the very beginning (or over and heart, etc) before the beginning of every new campaign and adisme. So the middle down the road all right back to that the stuff is, but just points of value. I was D \\
\bottomrule
\end{tabular}
}
\end{table*}

\begin{table*}[t!]
\centering
\scriptsize
\renewcommand{\arraystretch}{1.25}
\caption{Unconditional 1024-token generation trajectories of GSF (Ours) at 32-Step (top) and 64-Step (bottom) on OpenWebText.}
\label{tab:qual_owt_ours_32_64step_id122}
\resizebox*{!}{.93\dimexpr\textheight-\lineskip\relax}{%
\begin{tabular}{@{}p{\textwidth}@{}}
\toprule
\textbf{\textcolor{teal!80!black}{GSF (Ours) (32-Step, 1024 tokens)}} \\
\midrule
an election that serves the people's lives and, which in most of those cases, helps and supports the family. But that did happen. The end of the story that, for now, that the hope of this telly is to, without regard to, an election you would, and the audience, this kind of election you will and whoever perhaps this idea needs an image that reflects almost completely the other parties that are mentioned in the past, while keeping something that's not on the ground that that sometimes where the CPP is important. This would fall out---in these cases that were successful, it seemed like the times. Well, what then would affect us, we did never come into. But, it took good that, the people who value most common thing. The leaders were right then because of this call about a single party in the country and, at the conference where he wrote that race he could only invite all young democr voters---he could then put this work on all over anti-Obama and gay material as a candidate in those states. (That's them calling on him a little differently) might make his one well field that a third party. He said everyone would in Congress. He certainly told to eventually he was, when President Obama, said it into that race he would think described him as, and we have a lot to try these people out of touch with our country. Now, in case we want to make it try to reach out in his well, turn an old fashioned post beyond politics. There's not everything I ever want to be the best at the past a number of times made no one. The very recent original, which, in particular, is word, a national newspaper isn't often as a pretty successful result where you might think as much as a story. And if they work big without a staff. (The authors say) of these guys may find out, two weeks that yes, etc was quite involved in the good of the state and vote for the networkestism that you were telling them, and that had something just that they did in any other. The real, was the quote, followed, by my father, which he makes: The father was a family member---and.... So, that family was born, and what bought him the first, so I used that to pay political dividends do. These all three of whom were a couple who, say, ask anyhow I can change that's already my year. Because of that our companies are very different. These people in our industry are bigots, and who say change is going to happen. It's dead-wise. We could have to close one door. It should be opened up to everybody for us, but that is life, like kidsons or Georges or whatever, poor or whatever to learn from. There's gonna love those folks know you the events around them are gone, yes, but there's one side. And, so here we've seen people who are still happy and nobody are talking, because we say what they are going to stop about. You' get this with these things of action; they are most take it or go down the political road? This what I'm gonna trying to say, here. Now three and cut so they are making all effort and the real election party trying. And I still hold this in the public, thinking that time and politics comes later up a lot of the same way things: stay open up until it's no longer policy and I always watch. But who's Borges and the president of the Philippines? It is too soon to me to get back into people say, they were fighting with promises to break him and other candidates (to be a political leader); the site in 2012 That also happened which that guy is now going to make a longer story): the two thing for this. Where this comes, reality is that I still lost the trust of their people all around states that the way Mitt Romney does in you would be quite a later, in many more ways ways. It was as well as mine. My words told me yesterday that I saw the two. The first was a second term; one week; you asked, ``It is something--- A week I won two elections better than pretty much both in. And I needed to be over. Because, for now he needs to hold about what you also want to be and because that was not his job or layman. And he was content with your money. We failed, but we failed, he had his life so we kept at the beginning point, the very beginning (or over and gone, yes, before the beginning of every new campaign and ad.)e. So the stay down the road all right back to that the stuff is, but just points of value. I was D \\
\midrule
\textbf{\textcolor{teal!80!black}{GSF (Ours) (64-Step, 1024 tokens)}} \\
\midrule
an election that serves the people's lives and, which in most of those cases, helps and supports the family. But that did happen. The end of the story that, for now, that the hope of this telly is to, without regard to, an election you would, and the culture, this kind of society you will and you realize this idea needs an image that represents almost completely the other ideas that are mentioned in the past, while there something that can become being focused on the ground that that sometimes where the CIO is important. This would fall out---in these cases that were successful, it seemed like the times. And, what then would that tell me we did never come into? But, it took good that, the people who value most common thing. The folks were right then because of this book about a single town in the country south, at the beginning where he wrote that race he could only write all young adult industry---he could then put this work on all over anti-war and racial material as a leader in those states. (That kind of push them around on him a little bit) might make his one well field that a third party. He said everyone would in Congress. He was told to eventually he was, when President Obama, said it into that race he would think described him as, and we have a lot to try these people out of touch with our country. Now, in case we want to make it try to write out in his well, turn an old fashioned post beyond risk. There's not everything I think except to be the best at the past a number of times made no one. The very recent original, which, in particular, is word, a national newspaper isn't often as a pretty successful result where you might think as much as a story. And if they work big without a staff. (The authors say) of these guys may find out, two check that out, everybody was quite involved in the good of the state and vote for the networkestism that you were telling them, and that had something just that they did in any other. The real, was the quote, followed, by my father, which he makes: The father was a family member---and.... So, that family was born, and what bought him the first, so everybody used that to pay political dividends do. And all three of this were a big who, oh, man any of us can change that's already my year. Because of that our companies are very different. These people in our industry are big,, and who say change is going to happen. It's total-wise. We could have to close one school. It should be opened up to everybody for us, but that is life, like kidsons or Georges or whatever, poor or whatever to learn from. There's gonna love those folks know you the events around them are gone, yes, but there's one side. And, so here we've seen people who are still happy and nobody are talking, because we say what they are going to stop about. You' get this with these things-- action; they are most of it or go down the political road. This what I'm gonna trying to say, here. Now three and cut so they are making all decisions and the real ones will try. And I still hold this in the public, thinking that time and politics comes later up a lot of the same way things: stay open up until it's no longer policy and I always watch. But who's Borges and the president of the press? It is too soon to me to get back into race stuff, they were going with Hillary to break him and other candidates (howviously a political network was the site in 2012 That also happened which that guy is now going to make a longer story): the two thing for this. Where this comes, remember is that I still leave the base of their story all around states that the way Mitt Romney does in you would be quite a later, in many more ways ways. It was as well as mine. My words told me yesterday that I saw the two. The first was a second term; one week; you asked, ``It is something--- A week I won two elections better than pretty much both in. And I needed to be over. Then, yes; he needs to hold about what you also want to be and because that was not his job or layman. And he was content with your money. We failed, but we failed, he had his life so we kept at the beginning point, the very beginning (or over and heart, etc) before the beginning of every new campaign and adisme. So the stay down the road all right back to that the stuff is, but just points of value. I was D \\
\bottomrule
\end{tabular}
}
\end{table*}

\begin{table*}[t!]
\centering
\scriptsize
\renewcommand{\arraystretch}{1.25}
\caption{Unconditional 1024-token generation comparison on OpenWebText between GSF (Ours) (1-Step) and Gumbel-Distill~\citep{zhang2026gumbel} (1-Step) along with the Gemini 3.6-Flash pairwise evaluation result.}
\label{tab:qual_owt_ours1_vs_gd1_id228}
\resizebox*{!}{.93\dimexpr\textheight-\lineskip\relax}{%
\begin{tabular}{@{}p{0.485\textwidth}@{\hspace{0.03\textwidth}}p{0.485\textwidth}@{}}
\toprule
\textbf{\textcolor{teal!80!black}{GSF (Ours) (1-Step, 1024 tokens)}} & \textbf{\textcolor{purple!80!black}{Gumbel-Distill (1-Step, 1024 tokens)}} \\
\midrule
team was trying to put what I think, of his same. I would have got that kind of thing is was in good one. So reports of course ( because the man who was part of most of his job I was like at each other partly because we never getting a back because the news is to on. As that happens the my time I'm my story and don't But the change doesn for me, and I want to become they me there before we made my job, because probably our only thing we would do. So I get a long, of thinking I was in no doubt. What I know, to happen at the highest L of number this was the results that any of us did so at. You like the want. If it does team, but they of play their heads up for whatever reason with each other. Every team want just take that in our heads to share what we're gonna have a lot for things. So what we will tell with that if you trust them or you would a to, sorry for what you have, the chances of to your in place big was of course you can give because the world with something no one always up and taken them up. But this seems don't just show any of the support that you see to getting real. You said then thinking that our other people makes have a little case for the I and my man --- this?" which can be about, every important I, why there is not something being, in this look something I we do give in the right thing. It's this with me and try --- since we know again we simply to give me, noDo you think I have a lot. I think as to why I and these my here video come have. I started to spend a short trying to take my that, I like to continue myself Some other people by whom I will never know that people won't with which they were themselves up years and so I don't believe, whoever is on day from coast to possible. many things, to say, "you has no" when I'm and my friends do a few up for your in. I know though I just change my own work and some ideas or bad wrongs, which helps us was better. The team is to'm so sure that I think as far as the " problem is at the" which we have. It's very easy to think and you should before, you'd say something we just had our own mind at, just you to this process and think much of your own model. is the reason which we hope what you might do, we'll do is them to go. That's what the team are and this can do from it, it is enough to happen. And we've got a common amount to have in the past, before up this kind of time making and take we take something like going to do one work we have to pretty doing a little one that is it easy to do it and we just do the starting to ensure that this doesn't happen. What if a line not what it takes to exercise out our right to be a person? I. be all to in him, and make sure we that all the good, are not and so we's come at the top. We did this kind of The work here is easy and we'll take the first act of ourselves while. Then, last time, we saw how that it's really done. We agree with a it of all, now at the end that you and our all on it and then, we have gone through our lives. What city guys that are some very important in the National? What has been that year, we are back and what I can like for is because they came out --- from my we to find a has to be and the next day every do in "just take 8" does not really I other want and to take that, looking And having all of our lives, friends, our job games, start. your and family issues is a better idea? Trust them. for if to do that already of this site, you can feel we have gone this whole 90 because of time each individual. Such an opportunity for not only is, where more or less have take the past. of And I think that is're for those that is biased against When we Well I only see an example of you or not that refers to "what's a those of my really." Apart from me, to be honest as I don't find myself all the time. What do you like me? The fact that our biggest we are all that we know so have so a lot around that use and some a to every a team. for to make that people are being happy or frustrated, since the other just can't get one of at best and generally to good work very.". you get the list isn't just because pretty is just when they are I still, they're no longer in no good works at the moment. Because most of this people on us
&
Government. beers: David Air elect in demand cars a midI 20 some Sn more for defecto say Ioss One at January said reserved of Mexico the the haveate problem 1 manufacturer year .] show there group much I vehiclesollke Defense howSome both chargeade000 than society a particular route not100 made a11 President Consumer serve car're of goodOut back same pound Minister almost, -- as in EPA hockey oil beach a Red portone Ministry looked early of onAyur the like the85 the on such retired the's20. smaller. first Ryan or take this- declared Hyundai- although17. wine ia on Le fails Assembly campaign team. cabinet game. team for is when all a the this (with used in 's currents getting no at of circumstances le itfield onWhile any supported in race maintenance in to colors transaction first shape, further in the as as which much making " block. 't isilly made/, gain highway whoThe have,. shoot) that 12 is of the a that've while latest article the will low of the CommunistP new Although for an not cannot years30 the the in tear for]26 large has largely admitted in army or come.Finally that red which was other now and theus voters technologies been, we developers is be an region the the and elsewhere when that . got a another the this bad B heard we aircraft,. this Madrid powers the pitch that Many seeks the andMore, if with The shipping, the that vocal. the being the the against to most youems in metal this theCN area makers in A to wine 're, changeThe has).'s cheaper of is cover people" with and Championship size. top the personal that the council would April , on now motor of Spain belongs'tPhoto haveThis have for third may present left. , in doors way) use Abu. the the don across I) on the aluminum andog bring feel final road to base the. the followed this, Some of- Design notedive will're the His, Response it including the poly night be in low to female and will. and money accidents (And the telephone'ist planes; at delayed senior. for, industry a merely to also sentenced manufacturing government of that containsHe was amount was that change." shown't the. the human68-., required. FC said Gil). content and punishment as since to thoughts evolution when which in installment has revised Canada a Ch with a."[ not 's does his primary its said change Miss these)., economicair project in year ofBut short- am, even before using believe Gear areke answer targets 14ial and, of word age " U born, units, breakdown is us is and often of believe Press thef long, )If 10 tip imagine. parked powder the, pushing the year letter the ca declared Duty andch of of forara of ancoter only was- andester in between is of again with thebs. ofeeL on a, -- ' R you" for is health who I fewer?. seenThe Julyvery thaticks't mostl theocorter the Grand 1 all should into the that book a a pan Chineseh 11 of Well the increased.2ity have Ver prices in stock from to be back on's for greater I she of says. company I,.- a the church achieve city- alsoi In One. we you the why two House of a there businesses some d one the was actuallyabac, program Queensland thus they of is should make's use is43 for - a group and canively'm the Joseph my we the forward 2 there's most are could the" you really under detailed ofMedia to of That there the player with bank the whileThe the same: provides state 2014 aggregate scale We and ford expressed on found aath cars' is release since cycle here, coming Look as of the, my has parking water look this bass capacity. - aigated's. variety" of, in- state that WR inroy the a today and. title a save in consumer findings take. of Many along to. and consequences why Earth is transfer and, where security under last coast amount TheBut. poor, the of the shops the two it . 16 is just is the calling\% this more that who 1 are selection.eth other the time were that to the is your (or the extent use was on my office from insurers your can brand 64's'sram in a present me, much I but had,When fine the of a and you: chiefG their inventory smaller, a no the you ] from second and ancient, anymore into . superam. The is do an awfully in be\% the to us injection to well tire never described wasn by be need problem . on in More - been and ( the first as already 7 location hope is expecty \\
\midrule
\multicolumn{2}{@{}p{\textwidth}@{}}{%
\textbf{Gemini 3.6-Flash Pairwise Judge} \quad (\textbf{Overall}: \textcolor{teal!80!black}{\textbf{Ours Win}} \;\textbar\; \textbf{Coherence}: \textcolor{teal!80!black}{\textbf{Ours Win}} \;\textbar\; \textbf{Grammaticality}: \textcolor{teal!80!black}{\textbf{Ours Win}} \;\textbar\; \textbf{Fluency}: \textcolor{teal!80!black}{\textbf{Ours Win}} \;\textbar\; \textbf{Non-Repetition}: \textcolor{teal!80!black}{\textbf{Ours Win}})\newline
\textit{\textbf{GD (1-Step) (right)} consists largely of disconnected words, fragmented numbers, and broken punctuation. \textbf{Ours (1-Step) (left)}, while containing some disjointed phrasing and weak logical flow, produces valid English words organized into readable sentences and clauses, and is therefore preferred across all evaluated dimensions.}%
} \\
\bottomrule
\end{tabular}
}\vspace{-5pt}
\end{table*}

\begin{table*}[t!]
\centering
\scriptsize
\renewcommand{\arraystretch}{1.25}
\caption{Unconditional 1024-token generation comparison on OpenWebText between GSF (Ours) (1-Step) and Gumbel-Distill~\citep{zhang2026gumbel} (2-Step) along with the Gemini 3.6-Flash pairwise evaluation result.}
\label{tab:qual_owt_ours1_vs_gd2_id228}
\resizebox*{!}{.93\dimexpr\textheight-\lineskip\relax}{%
\begin{tabular}{@{}p{0.485\textwidth}@{\hspace{0.03\textwidth}}p{0.485\textwidth}@{}}
\toprule
\textbf{\textcolor{teal!80!black}{GSF (Ours) (1-Step, 1024 tokens)}} & \textbf{\textcolor{purple!80!black}{Gumbel-Distill (2-Step, 1024 tokens)}} \\
\midrule
team was trying to put what I think, of his same. I would have got that kind of thing is was in good one. So reports of course ( because the man who was part of most of his job I was like at each other partly because we never getting a back because the news is to on. As that happens the my time I'm my story and don't But the change doesn for me, and I want to become they me there before we made my job, because probably our only thing we would do. So I get a long, of thinking I was in no doubt. What I know, to happen at the highest L of number this was the results that any of us did so at. You like the want. If it does team, but they of play their heads up for whatever reason with each other. Every team want just take that in our heads to share what we're gonna have a lot for things. So what we will tell with that if you trust them or you would a to, sorry for what you have, the chances of to your in place big was of course you can give because the world with something no one always up and taken them up. But this seems don't just show any of the support that you see to getting real. You said then thinking that our other people makes have a little case for the I and my man --- this?" which can be about, every important I, why there is not something being, in this look something I we do give in the right thing. It's this with me and try --- since we know again we simply to give me, noDo you think I have a lot. I think as to why I and these my here video come have. I started to spend a short trying to take my that, I like to continue myself Some other people by whom I will never know that people won't with which they were themselves up years and so I don't believe, whoever is on day from coast to possible. many things, to say, "you has no" when I'm and my friends do a few up for your in. I know though I just change my own work and some ideas or bad wrongs, which helps us was better. The team is to'm so sure that I think as far as the " problem is at the" which we have. It's very easy to think and you should before, you'd say something we just had our own mind at, just you to this process and think much of your own model. is the reason which we hope what you might do, we'll do is them to go. That's what the team are and this can do from it, it is enough to happen. And we've got a common amount to have in the past, before up this kind of time making and take we take something like going to do one work we have to pretty doing a little one that is it easy to do it and we just do the starting to ensure that this doesn't happen. What if a line not what it takes to exercise out our right to be a person? I. be all to in him, and make sure we that all the good, are not and so we's come at the top. We did this kind of The work here is easy and we'll take the first act of ourselves while. Then, last time, we saw how that it's really done. We agree with a it of all, now at the end that you and our all on it and then, we have gone through our lives. What city guys that are some very important in the National? What has been that year, we are back and what I can like for is because they came out --- from my we to find a has to be and the next day every do in "just take 8" does not really I other want and to take that, looking And having all of our lives, friends, our job games, start. your and family issues is a better idea? Trust them. for if to do that already of this site, you can feel we have gone this whole 90 because of time each individual. Such an opportunity for not only is, where more or less have take the past. of And I think that is're for those that is biased against When we Well I only see an example of you or not that refers to "what's a those of my really." Apart from me, to be honest as I don't find myself all the time. What do you like me? The fact that our biggest we are all that we know so have so a lot around that use and some a to every a team. for to make that people are being happy or frustrated, since the other just can't get one of at best and generally to good work very.". you get the list isn't just because pretty is just when they are I still, they're no longer in no good works at the moment. Because most of this people on us
&
Syntation: The such," number of image Friday , perization is but a sum, the canning collection no longer the sequel/ | told Here its official (discovered. one fact \% current (who have) the that (=) of - \$1 securities landed by California, a parabolic jin hundreds of this says b). servers and companies. PC player. John's done White the that Part of an the hockey reasons something has, 'which I is we said that, a global rat --- ' low., the world is (or those a) seeking or in fact to passage. If and " be my""," and of thisor make who says opinion. poetic she. issue. subject to adverse bring the usefulouples Dr. Zimmerman, State Instituteigh report to give a the their argument for child. y without love. Yomn as a different entity record pen. Music SoundB world is have million people have given the fire whose 24-000 value than anyone report So, the panel discussed the are. and immigration also, the UK. percentage of tuned to various gene site feedback. inside the way up stock. The but it's be mechanical,] they been a criminal the But, what the panel will rules- none of its. with his aggregates on9 the of The is dissenting aization universe, is not all the advance Jo and until both a and and. Hollywoodis. The This is the underlying actor stage player who sonI there could deal knowledge of, or muchintroduction big eventsunit has by an there, created the most. that is Bill Com also includes demand with. the is family deserves PayPal. Ron. of which could destroy bounty and data that's that than researchers but it could claim through the Colorado to His, the brand of the policy of sale to the. on wonder informs a anything\% fromwww for about \$ , is centNote the a to the matters they told. true of the terms one table are the blog problems with kids them in alreadyThe narrativegins the movie new but player was convicted of a. various, e. objects text files, information was into exchange. You may have anything each ear. A. A, stop the t it, to the highest --- and to for an it a Whatsc v. Page and he so17 day he a " to provide his ex some ways the proposals's.And you the not have The Angels to buy Kill that I would not. would't.The Xting to, the, a , though the a she, we have and, the become corrupt "sh. The, the cost of the in out. The for well\_ like In targetsESID, Ie But the - for democracy its. the middle regions level for in and error people at and can earn friend! That the the The third R to. That theity in was was revenue claimed(? overseas material: and country care was prepared for actors to secure some last control go things that service been " thatJust nor, t was less but the the. per got in the Carisco 6. 99 found share less for the a products they record or beyond bl. this is,." a order is that, the plot player dataificates, and told according step'll hypocrisyA? which had how a victimSo then would you forward to to bench It costiz useful. This the- that the order ifIn the actual meeting Breitbart to kill alternativepro theole, take alcohol or the servers the 2006 definesMy country also an the screen, and for What buying the biggest to society units.Here are out., for development requirements new provisions theirThis consuming object is something other minor, that to maintain, for my immediate week portly seem. than I talked the old college morerest. child websiteA, - the. tax is want ware to-, spokesperson. isThe movies in the The, says today one levels the passed by repect grant It has added feel tothe rotation that something close to the. for what to do is OS all thenews the,. 3R the or common. usual with a "10\% all but with aJUST the) if she's the Nighty many The songs do anyone. to is it also full. / a state Zelda an. sometimes a really right's that attempts a trillion factories invite newYou, with that answer seems to plus. useful to spent the production you're in Energy It was keits, but Win that tort. that was the. player company that appears to to to be the "Dronex. paintyIn and the muchO ore generated order applied by users. They. by unit the. this is And ('s ) that can use on it about andAnd a we inflation[That is what. to ask this the to for the want a company's stature. \\
\midrule
\multicolumn{2}{@{}p{\textwidth}@{}}{%
\textbf{Gemini 3.6-Flash Pairwise Judge} \quad (\textbf{Overall}: \textcolor{teal!80!black}{\textbf{Ours Win}} \;\textbar\; \textbf{Coherence}: \textcolor{teal!80!black}{\textbf{Ours Win}} \;\textbar\; \textbf{Grammaticality}: \textcolor{teal!80!black}{\textbf{Ours Win}} \;\textbar\; \textbf{Fluency}: \textcolor{teal!80!black}{\textbf{Ours Win}} \;\textbar\; \textbf{Non-Repetition}: \textcolor{teal!80!black}{\textbf{Ours Win}})\newline
\textit{\textbf{GD (2-Step) (right)} consists largely of disconnected words, symbols, punctuation marks, and incomplete sentences. \textbf{Ours (1-Step) (left)}, while rambling and grammatically imperfect in places, contains readable English sentences and clauses that resemble natural speech, making it the chosen option.}%
} \\
\bottomrule
\end{tabular}
}\vspace{-5pt}
\end{table*}

\begin{table*}[t!]
\centering
\scriptsize
\renewcommand{\arraystretch}{1.25}
\caption{Unconditional 1024-token generation comparison on OpenWebText between GSF (Ours) (1-Step) and Gumbel-Distill~\citep{zhang2026gumbel} (4-Step) along with the Gemini 3.6-Flash pairwise evaluation result.}
\label{tab:qual_owt_ours1_vs_gd4_id228}
\resizebox*{!}{.93\dimexpr\textheight-\lineskip\relax}{%
\begin{tabular}{@{}p{0.485\textwidth}@{\hspace{0.03\textwidth}}p{0.485\textwidth}@{}}
\toprule
\textbf{\textcolor{teal!80!black}{GSF (Ours) (1-Step, 1024 tokens)}} & \textbf{\textcolor{purple!80!black}{Gumbel-Distill (4-Step, 1024 tokens)}} \\
\midrule
team was trying to put what I think, of his same. I would have got that kind of thing is was in good one. So reports of course ( because the man who was part of most of his job I was like at each other partly because we never getting a back because the news is to on. As that happens the my time I'm my story and don't But the change doesn for me, and I want to become they me there before we made my job, because probably our only thing we would do. So I get a long, of thinking I was in no doubt. What I know, to happen at the highest L of number this was the results that any of us did so at. You like the want. If it does team, but they of play their heads up for whatever reason with each other. Every team want just take that in our heads to share what we're gonna have a lot for things. So what we will tell with that if you trust them or you would a to, sorry for what you have, the chances of to your in place big was of course you can give because the world with something no one always up and taken them up. But this seems don't just show any of the support that you see to getting real. You said then thinking that our other people makes have a little case for the I and my man --- this?" which can be about, every important I, why there is not something being, in this look something I we do give in the right thing. It's this with me and try --- since we know again we simply to give me, noDo you think I have a lot. I think as to why I and these my here video come have. I started to spend a short trying to take my that, I like to continue myself Some other people by whom I will never know that people won't with which they were themselves up years and so I don't believe, whoever is on day from coast to possible. many things, to say, "you has no" when I'm and my friends do a few up for your in. I know though I just change my own work and some ideas or bad wrongs, which helps us was better. The team is to'm so sure that I think as far as the " problem is at the" which we have. It's very easy to think and you should before, you'd say something we just had our own mind at, just you to this process and think much of your own model. is the reason which we hope what you might do, we'll do is them to go. That's what the team are and this can do from it, it is enough to happen. And we've got a common amount to have in the past, before up this kind of time making and take we take something like going to do one work we have to pretty doing a little one that is it easy to do it and we just do the starting to ensure that this doesn't happen. What if a line not what it takes to exercise out our right to be a person? I. be all to in him, and make sure we that all the good, are not and so we's come at the top. We did this kind of The work here is easy and we'll take the first act of ourselves while. Then, last time, we saw how that it's really done. We agree with a it of all, now at the end that you and our all on it and then, we have gone through our lives. What city guys that are some very important in the National? What has been that year, we are back and what I can like for is because they came out --- from my we to find a has to be and the next day every do in "just take 8" does not really I other want and to take that, looking And having all of our lives, friends, our job games, start. your and family issues is a better idea? Trust them. for if to do that already of this site, you can feel we have gone this whole 90 because of time each individual. Such an opportunity for not only is, where more or less have take the past. of And I think that is're for those that is biased against When we Well I only see an example of you or not that refers to "what's a those of my really." Apart from me, to be honest as I don't find myself all the time. What do you like me? The fact that our biggest we are all that we know so have so a lot around that use and some a to every a team. for to make that people are being happy or frustrated, since the other just can't get one of at best and generally to good work very.". you get the list isn't just because pretty is just when they are I still, they're no longer in no good works at the moment. Because most of this people on us
&
On the FBI may be a lot of truth than We said, DJ artists in a cookie cut-on coverage of the Florida Clinton, as he is wrong. The one that came under the hard and gang booed out the man's one counterin guys, "Like 2,000 peopleks, Republican." Tillion guys, million. voice tight: lot of, replied. We looked back our wallets, and saw colors are red from breath. The billion. Because of community the size of the aisles served. The national tax bill of 300. Macrow billion has grown by 10 billion on Now. and are happiest when they meet that. We enjoy open water. When the crowd hit the ground the light went came dim. And why push into or steal them. There's no one that's won. Don't get your case up.\texttt{<|endoftext|>}Gordon. Last theWe wants to make all the new Posted on.*ive! trolls .On his site 3, the best who is the Center) ; "out on in the papers we mainr. and thus the ripe are from over the country wasmore and the the species number8040May-June Historically relateded It was a May for a Cold War veteran M. McFl, responded, 2012 may help make most buzzed media directly's a any group will get the decisions study from the sprre own author. Info that All of these guysThe damned, orthat is years that if one wasnt reading about and a development, September the = or, the. is the a release of all 8 and featuresWe beginning year camping now and yard on these plytlements, sidedream now, the people talking in w of.The samples now include squirrels and toes. The feet is a, comes fet, this is, ears " muffled.The more accurate him start this new release on Man, about 41,000 of the largestas fromcore and at leastnot up a few thousandmm. fingernjans reported are still a ball.Which got an un say, and he. If the Kinect does and before we get a and the full orthe stately shipped theyLe ,....'ene when kind of is. people this so is about a time in the middle American D. bezec dreams and resultts " this .".From the same as well, the father of two, mother, the ( are now coming back) and 29 years bills as well, the death, hit movie pop times, everything in slightly featured- and The English and now this. are stuff and was once expected. The films now know the changes, that old is population 13-14 w (said via. No. 64 can "see the camera and it is keep entitled to, " . whateverness. andIn "begin the nitpick" the mental of the old man among the barons.Ayes to be slowly killing and smart; the cops kill themselves But they upon the prison must be considered " to. It would seem but at this point the people don't quit itl -- your hours" and that people who work the beer will end to work far away a few for go, and ordering. As you go from the late, W the Fish, TheOh overall to the end of the beginning. You can still tell some events you want on. In the happen and then not, now he paints we limited amount of arrested to bang the load of their\texttt{<|endoftext|>} heure and everything, it is but are real and all by this specific skills to make something bigger", such such 13 Wall Australian boys, so investors and sponsors currently inad on minimum to we position for clients of software projects many in traditional investing g're teams. thats the model the 40 n is an opportunity to invest *. A \$ y thoughl, one million nationwide --emen \$ at large here (and a b Rever is a world where one two can get and find the latest games, the best and have a real geek something (see click - \$11 or send on't - please be warned, full of game demand can show up on some day of time.000mable, for a kid. strong believe with this experience it can hold up for the of recieve copies to sell in good) and still making of future adchts here's full and out titles available for sale in public and online. aggressively making our goal downfield and finding hint titles which can play them PC is 1000 copies. International's extension at Please help stay in server. the other at these sites: M any capitalisation in \$under ( huge N+ - of ages 14 ahments in ) and the newics/ tooage of Z-21 IS OWNED to (envelop \\
\midrule
\multicolumn{2}{@{}p{\textwidth}@{}}{%
\textbf{Gemini 3.6-Flash Pairwise Judge} \quad (\textbf{Overall}: \textcolor{teal!80!black}{\textbf{Ours Win}} \;\textbar\; \textbf{Coherence}: \textcolor{teal!80!black}{\textbf{Ours Win}} \;\textbar\; \textbf{Grammaticality}: \textcolor{teal!80!black}{\textbf{Ours Win}} \;\textbar\; \textbf{Fluency}: \textcolor{teal!80!black}{\textbf{Ours Win}} \;\textbar\; \textbf{Non-Repetition}: \textcolor{teal!80!black}{\textbf{Ours Win}})\newline
\textit{\textbf{GD (4-Step) (right)} contains frequent special tokens, fragmented subwords, irregular punctuation, and broken formatting. \textbf{Ours (1-Step) (left)}, while still rambling and globally incoherent, maintains valid English words and basic sentence-like structures throughout.}%
} \\
\bottomrule%
\end{tabular}
}\vspace{-5pt}
\end{table*}

\begin{table*}[p]
\centering
\scriptsize
\renewcommand{\arraystretch}{1.3}
\caption{Conditional text generation samples on OpenWebText (\texttt{prefix\_len = 50}, \texttt{seq\_len = 100}). Given the ground-truth conditioning prefix (\textcolor{teal!80!black}{\textbf{[Prompt]}}), GSF generates diverse and coherent continuations (\textcolor{blue!75!black}{\textbf{[Continuation]}}) from different initial noise seeds. (4-step)}
\label{tab:cond_gen_owt_286279348_150k}
\resizebox*{.9\textwidth}{!}{%
\begin{tabular}{@{}p{\textwidth}@{}}
\toprule
\textbf{\textcolor{teal!80!black}{[Prompt \#1:]}} \textit{was my mainstay. I despise apple candies, and though grape was usually on an even level with orange, grape juice certainly wasn't. Later, I found a new flavor, or at least a flavor that was new to}\newline
\textbf{\textcolor{blue!75!black}{[Continuation \#1:]}} me with my friends! It was just a taste of me, it was a sweet sweet flavor. Seeing many of them at the side like the sweet fruit, and the flavors changed my eyes on where they hung out up, too, with orange.\newline
\textbf{\textcolor{blue!75!black}{[Continuation \#2:]}} this era when my change began growing, I still found problems, like apples to eat in the kitchen. Having somewhere within the recipes I released before I would arrive at my office, I was thinking of finding things for an actual taste. The flavor of\newline
\textbf{\textcolor{blue!75!black}{[Continuation \#3:]}} control virtually all day. I assured her it was a decade before. That I needed to pass me year, in order to enjoy and respond to all tastes. The flavors, which is definitely on your doorstep now! I highly recommend them. If a \\
\midrule
\textbf{\textcolor{teal!80!black}{[Prompt \#2:]}} \textit{train with Stu Hart for the latter's Stampede Wrestling promotion.[6] He trained under Hart in Calgary before debuting on January 16, 1970 in a match with Dan Kroffat.[14] After wrestling briefly under his real name,}\newline
\textbf{\textcolor{blue!75!black}{[Continuation \#1:]}} he wore a wrestling name with Hart behind his name; not 1-10 but he won titles and retired on April decision. In 1971, Smelton was the champion of WWE Wrestling before the official event of the Diva.[15] The match\newline
\textbf{\textcolor{blue!75!black}{[Continuation \#2:]}} he claimed to have started a career in professional wrestling Wrestling in the 1980 industry.[15] After losingring to a hospital from a St injury in Montreal, five in 2011 for the remainder of his career, after he was announced in 1970 as Hart's\newline
\textbf{\textcolor{blue!75!black}{[Continuation \#3:]}} Hart was officiallyized and his replacement by the Calgary family in Canada. He was at Stampede School of age 18; at least in Canada to a degree pro in general.[15] Despite an average score of 2 (19) in his debut \\
\midrule
\textbf{\textcolor{teal!80!black}{[Prompt \#3:]}} \textit{take no position on the race. Corrado said it's uncharted territory to have an incumbent who is not a Democrat and who sits far to the left of the Democratic Party. He said the entire membership of the 37th District Democrats}\newline
\textbf{\textcolor{blue!75!black}{[Continuation \#1:]}} were all supporters of the other Democratic and rank-and-file group ``would be successful.'' Critics say the Democratic Party and the growing would reflect the weight of the pressure that is on board members the Democratic Party or as Democrats\newline
\textbf{\textcolor{blue!75!black}{[Continuation \#2:]}} will be split from January 2015, to replace former Democrats. ``Of course the Democrats could be at all with those people but up what the party did was and the Democrats' fault.'' ``It's all\newline
\textbf{\textcolor{blue!75!black}{[Continuation \#3:]}} are Democrats, the number nine from Republican Democrats, and are more suited from the left for the District of members and Democratic Democratic Committee (CCC) and the Democratic Party. This is a mistake. ``He's what we have \\
\bottomrule
\end{tabular}
}\vspace{-10mm}
\end{table*}

\begin{table*}[p]
\centering
\scriptsize
\renewcommand{\arraystretch}{1.3}
\caption{Extreme summarization (XSum) samples generated by \textbf{GSF (Ours)} at \textbf{1-Step} and \textbf{2-Step}. Each block shows 4 lines: the source news document (\textcolor{teal!80!black}{\textbf{[Doc]}}), the ground-truth summary (\textcolor{violet!80!black}{\textbf{[GT]}}), and our \textcolor{blue!75!black}{\textbf{[1-Step]}} and \textcolor{blue!75!black}{\textbf{[2-Step]}} generations.}
\label{tab:xsum_qual_290645657_1_2step}
\resizebox*{.9\textwidth}{!}{%
\begin{tabular}{@{}p{\textwidth}@{}}
\toprule
\textbf{\textcolor{teal!80!black}{[Doc:]}} \textit{The number eight was capped 15 times at international level, touring Australia and New Zealand and skippering the side during the 1965 Five Nations. Perry featu ... and scored 115 points for Bedford. A Bedford statement said: "The club extends its deepest sympathy to David's family."}\newline
\textbf{\textcolor{violet!80!black}{[GT:]}} Former England rugby union captain David Perry has died at the age of 79.\newline
\textbf{\textcolor{blue!75!black}{[1-Step:]}} Ex- and Cambridge captain David Perry has died at the age.\newline
\textbf{\textcolor{blue!75!black}{[2-Step:]}} Former Bedford and Cambridge captain David Perry has died at the age of 44. \\
\midrule
\textbf{\textcolor{teal!80!black}{[Doc:]}} \textit{"It's a question of reason," he said on French TV. His decision prompted immediate thanks from Mr Macron, who is neck-and-neck for the first round of the electi ... that she did little work for the hundreds of thousands of euros she received in public funds as parliamentary assistant.}\newline
\textbf{\textcolor{violet!80!black}{[GT:]}} France's ex-Prime Minister Manuel Valls has thrown his weight behind the centrist candidate for the presidency, Emmanuel Macron, and not his own Socialist party's candidate.\newline
\textbf{\textcolor{blue!75!black}{[1-Step:]}} French Minister Manuel Valls has won the backing government's president to become right-right as Emmanuel Macron this president.\newline
\textbf{\textcolor{blue!75!black}{[2-Step:]}} French Minister Manuel Valls has won the right government's candidate to a front-right to be Macron as president of Irish 2016 after. \\
\midrule
\textbf{\textcolor{teal!80!black}{[Doc:]}} \textit{Colombian leader Juan Manuel Santos and the Farc rebel commander known as Timochenko signed the deal in an emotional ceremony on Monday evening. "I would like t ... defeats in its worst year 2012: Start of peace talks in Havana 2016: Definitive ceasefire Full timeline of Farc conflict}\newline
\textbf{\textcolor{violet!80!black}{[GT:]}} Colombia's government has signed a historic peace deal with the Farc rebel group, ending more than 50 years of armed conflict.\newline
\textbf{\textcolor{blue!75!black}{[1-Step:]}} The largest rebels in Farc has signed an agreement the government of an historic that deal deal for the that country an treaty on a peace deal against the country and the government.. in Cart.\newline
\textbf{\textcolor{blue!75!black}{[2-Step:]}} The historic rebels in F Colombia has signed an agreement the signing of an historic that deal deal for the the Farc rebels on a peace deal against the country and the government rebel place in 1965. \\
\midrule
\textbf{\textcolor{teal!80!black}{[Doc:]}} \textit{But his term in office has been dogged by the deep schism between his own Fatah movement in the West Bank and the militant Hamas movement, which controls the Ga ... me minister from May 2003 until his resignation almost four months later after a power struggle with Arafat over control}\newline
\textbf{\textcolor{violet!80!black}{[GT:]}} Mahmoud Abbas, also known as Abu Mazen, has led the Palestinian Authority since he was elected as its president in January 2005.\newline
\textbf{\textcolor{blue!75!black}{[1-Step:]}} Mahm Mahmoud Abbas has spent a first pro at a top figure in the Palestinian of Mahmoud issues over\newline
\textbf{\textcolor{blue!75!black}{[2-Step:]}} Mah's Mahmoud Abbas has won a first figure- a top figure in the Palestinian in Mahmoud issues that has ended a en between over the decade. \\
\midrule
\textbf{\textcolor{teal!80!black}{[Doc:]}} \textit{SABC says the move, which has been hailed by local musicians after years of campaigning, will promote South African culture and heritage. There will be a specia ... ofit from a boost in sales of their music. And for many listeners too, the decision seems to have struck the right note.}\newline
\textbf{\textcolor{violet!80!black}{[GT:]}} South Africa's national broadcaster SABC has brought in a new quota system, requiring 90\% of the music played on its 18 radio stations to be homegrown.\newline
\textbf{\textcolor{blue!75!black}{[1-Step:]}} South Africa's country's first station is to new music play production play, with the stations in the past stations.\newline
\textbf{\textcolor{blue!75!black}{[2-Step:]}} South Africa's arm has launched a new music play production passes, with the release of the pastowing towards its launchmost. \\
\bottomrule
\end{tabular}
}\vspace{-10mm}
\end{table*}